\RequirePackage{fix-cm}
\documentclass[smallextended]{svjour3mod}       
\smartqed  

\usepackage{graphicx}
\usepackage{color}
\usepackage{amssymb}
\usepackage{amsmath}
\usepackage{cases}
\usepackage{url}
\usepackage{multirow}
\usepackage{subcaption}
\usepackage{tabularx}
\PassOptionsToPackage{usenames,dvipsnames,svgnames}{xcolor}  
\usepackage{tikz}
\usetikzlibrary{arrows,positioning,automata}
\usepackage{verbatim}

\usepackage{geometry}
\usepackage[ruled,vlined]{algorithm2e}
\SetKwInput{KwData}{Input}
\SetKwInput{KwResult}{Output}

\newcommand{\blue}{\color{black} } 

\journalname{AStA Advances in Statistical Analysis}

\begin{document}

\title{Hierarchical Clustering and Matrix Completion for the Reconstruction of World Input-Output Tables 
}
\subtitle{}

\titlerunning{Reconstruction of world input-output tables}        

\author{Rodolfo Metulini \and Giorgio Gnecco \and Francesco Biancalani \and Massimo Riccaboni}


\institute{R. Metulini \at
	Department of Economics and Statistics (DISES), University of Salerno, Via Giovanni Paolo II, 132 - 84084 Fisciano (SA), Italy \\
	\email{rmetulini@unisa.it - Corresponding author}  \\
	ORCID iD: 0000-0002-9575-5136
	\and 
	G. Gnecco \at
	Laboratory for the Analysis of CompleX Economics Systems (AXES), IMT School for Advanced Studies, Piazza S. Francesco, 19 - 55100 Lucca, Italy 
\\
ORCID iD: 0000-0002-5427-4328
        \and
 F. Biancalani \at
    Laboratory for the Analysis of CompleX Economics Systems (AXES), IMT School for Advanced Studies, Piazza S. Francesco, 19 - 55100 Lucca, Italy \\
ORCID iD: 0000-0002-4955-9871
\and
 M. Riccaboni \at
Laboratory for the Analysis of CompleX Economics Systems (AXES), IMT School for Advanced Studies, Piazza S. Francesco, 19 - 55100 Lucca, Italy   \\
ORCID iD: 0000-0003-4979-8933
}

\date{}

\maketitle

\begin{abstract}



{\blue World Input-Output (I/O) matrices provide the networks of within- and cross-country economic relations. 
In the context of 
I/O analysis, the methodology adopted by national statistical offices in data collection raises the issue of obtaining reliable data in a timely fashion and it makes the reconstruction of (part of) the I/O matrices 
of particular interest.} 
In this work, we propose a method combining hierarchical clustering and Matrix Completion (MC) with a LASSO-like nuclear norm penalty, to impute missing entries of a partially unknown I/O matrix. 
Through simulations {\blue based on synthetic matrices} 
we {\blue study} the effectiveness 
of the proposed method 
to predict missing values from both previous years data and current data related to countries similar to the one for which current data are obscured.
{\blue To show the usefulness of our method, an application based on World Input-Output Database (WIOD) tables -- which are an example of industry-by-industry I/O tables --
is provided. Strong similarities in structure between WIOD and other I/O tables are also found, which make the proposed approach easily generalizable to them.}
   
\keywords{Matrix Completion \and LASSO-like Nuclear Norm Penalty \and Panel Data Analysis \and Hierarchical Clustering \and Input-Output Tables}
\end{abstract}

\section{Introduction}


{\blue The world economy is characterized by the interdependence of all countries with respect to their industrial activities. It can be modeled as a network in which each {\blue  ``country-sector''} pair exchanges goods and services with any other (and even the same) such pair, at a different extent and using different technologies.} The structure of this network may provide information about {\blue  such} interdependencies of national economies and their changes over time.
{\blue The literature usually refers to the study of these interdependencies with the term ``Input-Output (I/O) analysis''. {\blue I/O tables portray flows of what is produced by some economic agents and consumed by other economics agents (either as intermediate or final consumption). They are arranged into two types, either according to relationships between industries (industry-by-industry I/O tables) or according to relationships between products (product-by-product I/O tables)}. Many world I/O tables for selected countries are periodically produced by, e.g., the World Input-Output Database (WIOD, \cite{Timmer2015,Timmer2016}), the EORA Global Supply Chain initiative \cite{Lenzen2012}, EXIObase \cite{Tukker2013}, the Full International and Global Accounts for Research in input-Output analysis (FIGARO,   \cite{Remond2019}) from EUROSTAT, and OECD \cite{OECD2018}\footnote{Other economic applications of network studies 
include, e.g., the cases of ownership networks \cite{Riccaboni2019}, banking networks \cite{IoriDeMasi2008}, international trade networks and foreign direct investment networks \cite{FagioloReyes2008},  \cite{Metulini2017}, \cite{Sgrignoli2015}}}. 
 By analyzing I/O matrices, \cite{McNerneyFath2013,Cerina2015,Zhu2015} studied the network topology of inter-industry flows and \cite{LiangQi2016} {\blue  investigated} its scaling patterns.
Some studies highlighted the presence of significant asymmetry between inflow data (i.e., what a {\blue ``country-sector''} pair uses from other {\blue ``country-sector''} pairs) and outflow data (i.e., what a country-sector sells to other {\blue ``country-sector''} pairs), and the emergence of a clustering pattern among either countries or sectors. This can derive, e.g., from the presence of similar production technologies (i.e., similar dependence of two specific countries or sectors on other {\blue  ``country-sector''} pairs) \cite{Carvalho2009}. This issue was taken into account, e.g., in \cite{ZhuMorrisonPuligaChessaRiccaboni2018}, where an ad-hoc measure was used to detect patterns of similarity and to group together similar countries and similar sectors, and in \cite{OlivaSetolaPanzieri2016}, where a spectral clustering approach was applied to group similar sectors of the Italian economy.

Our work has to do with the reconstruction of I/O matrices. The interest in {\blue this topic} arises if one takes into account some aspects related to the methodology of data collection adopted by national statistical offices. Indeed, based on \cite{Percoco2006}, the work \cite{WenXuWen2014} provides the following motivations to apply a reconstruction method to I/O matrices:
\begin{itemize}
\item Data collection for I/O matrices is typically based on {\blue direct methods, e.g., on} surveys ({\blue made} at different times and in different countries), which leads to classical sampling errors;
\item For large surveys, errors from the inference design can easily arise;
\item The elements of an I/O matrix change over time. However, collecting them in a timely fashion (for instance, once for each country every year) is an almost impossible task, due to limited resources. For this reason, historical data are typically used to approximate the current {\blue I/O} matrix.
\end{itemize}

{\blue Direct compilation  of  I/O tables does not rely only on surveys but also on administrative registers (e.g., firms registries, accountancies of companies in official registers, public budget documents, trade customs information, taxes authorities reports, households registers), censuses (population, firms directories) and, among  others, sectorial reports providing global economic information about specific sectors in an industry or in a country. It follows that, in order to compile an I/O table, many different statistical sources are contrasted (sometimes even contradictory ones). This is a huge time-consuming and resources-consuming operation, and the time gap between the publication of an I/O table and its reference year is one of the main reasons why also indirect methods 
exist (see \cite{Valderas-Jaramilloetal2019} for a review).}
 
In this work, taking advantage from the similarity patterns discussed above, we apply, {\blue as an indirect method}, Matrix Completion (MC, \cite{Hastie2015,Mazumder2010,Negahban2012}) to {\blue I/O submatrices associated with suitable} groups of countries, by judiciously clustering {\blue them} together, {\blue via a proper clustering method}.  
{\blue We recall here that MC is a set of advanced statistical methods that can be used to predict unobserved entries of a matrix in terms of the set of the remaining observed entries (more technical details about MC are reported later in Subsection \ref{sec:clustering_matrix_completion}). In this way, the partially observed matrix is ``completed'' by the predictions produced by MC.}
We show that using a selected group of similar countries permits to increase the effectiveness of MC. This is done by comparing the results obtained by MC when, on the contrary, the selected group is made by highly dissimilar countries. {\blue To the best of the authors' knowledge, this is the first article in which MC is applied to I/O tables in connection with clustering.}

The aforementioned similarity patterns have  important consequences on the structure of the I/O matrices, justifying the {\blue application} of MC. In fact, due the presence of countries that share similar technologies for producing the same goods, an I/O matrix might be low-rank, or, at least, might be well-approximated by a low-rank matrix (in the sense that a few singular values would dominate all the other ones, i.e., {\blue the singular values would decade quickly to $0$}). This low-rank (approximation) property suggests, among other possible statistical or machine learning techniques, the adoption of MC to reconstruct potentially missing entries in an I/O matrix. Moreover, {\blue such} a property {\blue is a necessary condition for obtaining} good MC results {\blue (see Appendix \ref{appendix:1} for a discussion on this issue)}. 
{\blue Nevertheless}, the application of MC to a full I/O matrix is not straightforward,{
\blue since elements in different blocks of that matrix have quite different orders of magnitude. 
Having analyzed different real-world I/O matrices, we found that: (1) within-country values are way larger than cross-country ones; (2) I/O matrices are sparse, because of many cross-country zero entries, and (3) there is a clear separation between large-to-large countries' values and small-to-small countries' ones. 
This suggests performing a pre-processing step, in which some blocks are removed from the full I/O matrices.}
{\blue In particular, we focus our analysis on bilateral trade blocks, which is quite a relevant problem in multi-country and multi-regional I/O tables.} 

{\blue Based on cross-country subsets of real-world I/O matrices}, we {\blue performed} a panel {\blue analysis} using {\blue MC based on} a LASSO-like nuclear norm penalty \cite{Mazumder2010,Negahban2012}, which permits, through {\blue a suitable} choice of the regularization parameter {\blue (based on a validation set)}, to select the number of non-zero singular values to be kept in the reconstructed matrix. {\blue The specific selection of countries was generated by the output of hierarchical clustering, whose application was based on a dissimilarity measure (the Average Absolute Correlation Distance or AACD, see later) highly related to the successive application of MC. Robustness of the results produced by hierarchical clustering was evaluated by considering synthetic counterparts of real-world I/O matrices, presenting a structure that is common to many I/O tables. This was done in order to generalize the approach to all kinds of I/O matrices, not limiting just to only one of them.}

{\blue In summary, this paper proposes a two-step methodological approach where, in the first step, judiciously selected groups of countries -- in terms of either i) the distribution by country-industry from where they buy inputs, or ii) the distribution by country-industry to which they sell outputs -- are retrieved using a hierarchical cluster analysis \cite{Revelle1979}. This choice is preferable to other clustering techniques (e.g., $k$-means clustering \cite{MacQueen1967}) to group countries according to their I/O exchanges where the set of countries themselves is not a-priori partitioned into a certain number of groups, but their complex structure suggests a hierarchy of clusters (see also Subsection \ref{sec:clustering_matrix_completion} for other technical motivations behind this choice). In the second step, an I/O submatrix associated with this selection of countries is analysed based on a LASSO-like algorithm for MC.}
{\blue In particular, our approach is applied to a 5-year panel of (cross-country subsets of) I/O matrices for subsets of countries selected after performing hierarchical} {\blue clustering, where a known part of the matrix associated with a specific year (i.e., the latest one) has been artificially obscured. The validation/testing phase is based on the Root Mean Square Error (RMSE) and on the Symmetric Mean Absolute Percentage Error (SMAPE) between actual and estimated values of the obscured part of the matrix, which is not provided as input to the MC algorithm.}
{\blue Ad-hoc simulations have been also performed in order to i) evaluate the improvement that could be achieved by a suitable pre-processing of raw data (by eliminating domestic blocks); ii) select the proper number of clusters; iii) evaluate the MC performance if the clustering failed in the first step. 
According to the latter point, we apply MC to subsets of either similar (cluster does not fail) or dissimilar (cluster fails) countries.}

Results show the effectiveness 
of the proposed method to predict missing values in the current I/O matrix from both previous years data and current data related to countries similar to the one for which current data are obscured. In contrast, the effectiveness reduces, {\blue as expected}, if similar countries are replaced by ones belonging to quite different clusters.


The rest of the manuscript begins with presenting the proposed methodological approach in Section \ref{sec:method} and {\blue reports} the specific application to WIOD {\blue and simulated matrices} and its results in Section \ref{sec:application}. Section \ref{sec:conclusion} is dedicated to future research directions and conclusions. {\blue To improve the readability of the work, some details (especially related to Section \ref{sec:application}) are deferred to footnotes and to the Appendix.}

\section{Methods} \label{sec:method}

\subsection{The Input-Output model} \label{sec:inoutmodel}
The traditional I/O matrix \cite{Leontief1986} depicts inter-industry relationships within an economy (or country), showing how the output from one sector becomes an input to another sector ({\blue or} itself), or it contributes to the final demand. Row indices represent inputs (in nominal monetary values) from an industrial sector, while column indices represent intermediate outputs to a given sector or needed to produce a final output. This kind of table shows how dependent each sector is on every other sector, both as a customer of outputs from other sectors, and as a supplier of inputs.

Suppose an economy with $n$ sectors and $l$ final outputs is given, the assumption of constant returns to scale can be made, and sectors use inputs in fixed proportions. Fix also a specific year. In that year, each sector $i$ produces a monetary value $x_{i}$ of good $i$. Let $z_{i,j}$ be the value that sector $i$ sells to sector $j$ in that year, and let $f_{i,j}$ be the value that sector $i$ sells to the final user in that year, to produce the final output $j$. 
In matrix notation, if one lets
\begin{equation}\label{eq:xvector}
\bold{x} = \begin{bmatrix} x_1 \\ \cdot \\ \cdot \\ \cdot \\ x_n \end{bmatrix}\,,
\end{equation}
\begin{equation}\bold{Z} = \begin{bmatrix} z_{1,1} & \cdot &  \cdot & \cdot & z_{1,n} \\ \cdot & & & & \cdot \\ \cdot & & & & \cdot \\ \cdot & & & & \cdot \\ z_{n,1} & \cdot & \cdot & \cdot & z_{n,n} \end{bmatrix}\,,
\end{equation}
\begin{equation}\label{eq:f}
\bold{F} = \begin{bmatrix} f_{1,1} & \cdot & \cdot & \cdot & f_{1,l} \\ \cdot & & & & \cdot \\ \cdot & & & & \cdot \\ \cdot & & & & \cdot \\ f_{n,1} & \cdot & \cdot & \cdot & f_{n,l} \end{bmatrix}\,,
\end{equation}
one can write $\bold{x} = \bold{Z}\bold{i}_n +\bold{F} \bold{i}_l$, where $\bold{i}_n \in \mathbb{R}^{n \times 1}$ and $\bold{i}_l \in \mathbb{R}^{l \times 1}$ are column vectors made of all ones.

{\blue Available I/O tables can also report the multi-national structure of intra- and inter-industries (products) exchanges. In this case}, let $m$ be the number of countries considered. Then, analogously as in Equations (\ref{eq:xvector})-(\ref{eq:f}), one sets
\begin{equation}\label{eq:xWIOD}
\bold{x} = \begin{bmatrix} \bold{x}^1 \\  \cdot \\ \cdot \\ \cdot \\ \bold{x}^m \end{bmatrix}\,,
\end{equation}
\begin{equation}\label{eq:ZWIOD}
\bold{Z} = \begin{bmatrix}
\bold{Z}^{1,1} & \cdot & \cdot & \cdot & \bold{Z}^{1,m} \\ \cdot & \cdot & & & \cdot \\ \cdot & & \bold{Z}^{h,k} & & \cdot \\ \cdot & & & \cdot & \cdot \\ \bold{Z}^{m,1} & \cdot & \cdot & \cdot & \bold{Z}^{m,m} \end{bmatrix}\,,
\end{equation}
\begin{equation}\label{eq:FWIOD}
\bold{F} = \begin{bmatrix}
\bold{F}^{1,1} & \cdot & \cdot & \cdot & \bold{F}^{1,m} \\ \cdot & \cdot & & & \cdot \\ \cdot & & \bold{F}^{h,k} & & \cdot \\ \cdot & & & \cdot & \cdot \\ \bold{F}^{m,1} & \cdot & \cdot & \cdot & \bold{F}^{m,m} \end{bmatrix}\,,
\end{equation}
where the generic (column) block $\bold{x}^{m}$ of the vector $\bold{x}$ in Equation (\ref{eq:xWIOD}) can be expressed as
\begin{equation}
\bold{x}^m = \begin{bmatrix} x^m_1 \\ \cdot \\ \cdot \\ \cdot \\ x^m_n \end{bmatrix}\,.
\end{equation}

{\blue In this extended framework, one can analogously write $\bold{x} = \bold{Z}\bold{i}_{nm} +\bold{F} \bold{i}_{lm}$, where $\bold{i}_{nm} \in \mathbb{R}^{nm \times 1}$ and $\bold{i}_{lm} \in \mathbb{R}^{lm \times 1}$ are column vectors made of all ones and $\bold{x}$, $\bold{Z}$ and $\bold{F}$ are those in Equations (\ref{eq:xWIOD})-(\ref{eq:FWIOD})}.

In the above, the generic block $\bold{Z}^{h,k}$ of the matrix $\bold{Z}$ in Equation (\ref{eq:ZWIOD}) represents the I/O subtable where $h$ is the country in input and $k$ is the country in output. Such a block can be expressed as
\begin{equation}
\bold{Z}^{h,k} = \begin{bmatrix}
z^{h,k}_{1,1} & \cdot & \cdot & \cdot & z^{h,k}_{1,n} \\ \cdot & & & & \cdot \\ \cdot & & & & \cdot \\ \cdot & & & & \cdot \\ z^{h,k}_{n,1} & \cdot & \cdot & \cdot & z^{h,k}_{n,n} \end{bmatrix}\,,
\end{equation}
whereas the generic block $\bold{F}^{h,k}$ of the matrix $\bold{F}$ in Equation (\ref{eq:FWIOD}) can be expressed as 
\begin{equation}
\bold{F}^{h,k} = \begin{bmatrix}
f^{h,k}_{1,1} & \cdot & \cdot & \cdot & f^{h,k}_{1,l} \\ \cdot & & & & \cdot \\ \cdot & & & & \cdot \\ \cdot & & & & \cdot \\ f^{h,k}_{n,1} & \cdot & \cdot & \cdot & f^{h,k}_{n,l} \end{bmatrix}\,.
\end{equation}

The ``transition'' matrix $\bold{T}=[\bold{Z} | \bold{F} ]$ is obtained from Equations (\ref{eq:ZWIOD}) and (\ref{eq:FWIOD}). Here, $\bold{T} \in \mathbb{R}^{mn \times m(n+l)}$ is a 
matrix whose row and column indices refer to an ordered {\blue ``country, intermediate/final output''} pair.

{\blue Since hierarchical clustering will be applied -- depending on either i) input or ii) output criteria -- to specific I/O submatrices, it is worth specifying two types of} submatrices of $\bold{T}$ obtained by combining several blocks of the form $\bold{T}^{h,k}=[\bold{Z}^{h,k} | \bold{F}^{h,k}]$. 
As an example, for $h \neq 1,m$, let 
\begin{align}
\bold{T}^{h,.}
= \begin{bmatrix}
\bold{T}^{h,1} |
\cdots |
\bold{T}^{h,h-1} | 
\bold{T}^{h,h+1} |
\cdots |
\bold{T}^{h,m} 
\end{bmatrix}
\end{align}
be the transition submatrix related to what all sectors of country $h$ sell to all sectors/final users of all countries except country $h$; and similarly, for $k \neq 1,m$, let 
\begin{align}
\bold{T}^{.,k}
= \begin{bmatrix}
\bold{T}^{1,k} \\
\cdots  \\
\bold{T}^{k-1,k} \\ 
\bold{T}^{k+1,k} \\
\cdots \\
\bold{T}^{m,k} 
\end{bmatrix}
\end{align}
be the transition submatrix related to what all sectors/final users of country $k$ buy from all sectors of all countries except country $k$. Similar definitions obviously hold for $h,k \in \{1,m\}$.
The submatrices $\bold{T}^{h,.}$ have $n$ rows and $(m-1)(n+l)$ columns, whereas the submatrices $\bold{T}^{.,k}$ have $(m-1)n$ rows and $n+l$ columns.
About the first kind of submatrix ({\blue  criterion i) of clustering)}, for example, the first block $\bold{T}^{h,1}$ expresses what all sectors/final users of country 1 buy from all sectors of country $h$ (say, Italy).
About the second kind of submatrix ({\blue  criterion ii) of clustering)}, for example, the first block $\bold{T}^{1,k}$ expresses what all sectors of country 1 sell to all sectors/final users of country $k$ (say, Italy).

\subsection{Clustering and matrix completion}\label{sec:clustering_matrix_completion}

It is recalled here that clustering is an unsupervised learning technique whose goal consists in partitioning a data set into several subsets (called clusters), by making more similar data points belong (if possible) to the same cluster, while assigning less similar data points (if possible) to distinct clusters \cite{AggarwalReddy2014}\footnote{Clustering techniques were recently applied to {\blue I/O} matrices, e.g., in \cite{OlivaSetolaPanzieri2016,ZhuMorrisonPuligaChessaRiccaboni2018}. In \cite{OlivaSetolaPanzieri2016}, a hierarchical clustering approach was adopted, based on the recursive application of spectral clustering, which is a clustering technique suitable for data organized according to a graph structure.
The analysis was done at the sector level, for the WIOD submatrix obtained by keeping only rows and columns related to Italy, for the period 1995-2011. 
In \cite{ZhuMorrisonPuligaChessaRiccaboni2018}, a network-based measure of similarity (and the resulting unweighted average distance clustering) was proposed, to compare the (upstream and downstream) Global Value Chains (GVCs) between any pair of countries, for each sector and each year available at the time in the WIOD database.}. 
{\blue Among several clustering methods\footnote{\blue It is worth mentioning  that also global and local spatial clustering techniques have been developed in the literature to, respectively, test whether a clustering structure is present in the analysed region and to identify the location of clusters (see, e.g., \cite{Aldstadt2010} for an exhaustive review).
Even if countries of I/O tables present a spatial dimension, we are not going to use any kind of spatial clustering technique, due to the following reasons: i) 
volumes of I/O {\blue exchanges across countries, especially in recent years, are likely not determined by spatial contiguity, but by network relationships, ii) in this work we are not interested in finding the presence of a spatial pattern: the clustering step aims at selecting a subset of similar countries, in terms of network relationships quantified by a suitable dissimilarity measure (AACD), in order to make the successive application of MC easier.}}, we choose a hierarchical method because, in I/O tables, the set of countries is not a-priori partitioned into a certain} {\blue number of groups and because their  complex  structure  suggests  a  hierarchy of clusters. By using a set of pairwise dissimilarities for {\blue $m$} objects (i.e., countries), {\blue hierarchical clustering} first assigns each object to its own cluster, then it 
proceeds iteratively by joining at each stage the two most similar clusters, continuing until there is just a single cluster. At each stage, distances between clusters are recomputed by the Lance--Williams formula \cite{LanceWilliams1967}, based either on the complete or on the Ward linkage criterion \cite{MurtaghLegendre2014}. According to hierarchical clustering, differently from (e.g.) the $k$-means clustering method, one can achieve different partitions of objects depending on the level of resolution one is looking at. 
Moreover, despite $k$-means is less computationally expensive compared to hierarchical clustering, it requires strict assumptions regarding the homoschedasticity and the spherical variance of the variables, and that each cluster has -- a-priori -- roughly an equal number of objects. Hierarchical clustering performs well even when those assumptions are not satisfied.}

To measure how (dis)similar any two objects are, several (dis)similarity measures have been developed in the literature. Among them, the most famous and commonly used are the $l_1$ norm of the difference of data points (also called Manhattan distance), and the $l_2$ norm of their difference (also called Euclidean distance). {\blue In this work, we use the Average Absolute Correlation Distance (AACD) as a dissimilarity measure for clustering. In other words, the absolute value of the correlation between the $i$-th corresponding non-constant columns ${\bold b}^{c_1}_{i}$ and ${\bold b}^{c_2}_{i}$ of blocks\footnote{\blue The specific structure of the blocks considered is reported, e.g., in Table \ref{tab:2temp} for the case of inputs from Italy, and in Table \ref{tab:2temp_input} for the case of outputs from Italy. Each column ${\bold b}^{c}_{i}$ {\blue is composed of all the data related to a specific country $c$ (different from Italy) and the intermediate/final output $i$ in the years 2010-2013. The last year (2014) is not considered for the computation of the correlation because parts of its data refer to the validation/test set related to the successive MC application.}} associated with two}  {\blue  different countries $c_1$ and $c_2$ is evaluated, then it is averaged with respect to the columns, and subtracted from $1$. In formulas, one has
\begin{equation}
    AACD_{c_1,c_2} = 1-\frac{\sum_{i=1}^{n+l}|corr({\bold b}^{c_1}_{i},{\bold b}^{c_2}_{i})|} {n+l}\,.
\end{equation}
This choice of the dissimilarity measure is motivated by the fact that AACD is highly related to the formulation of the optimization problem modeling MC (see the next Equation (\ref{eq:matrix_completion2})). Indeed, in a sense, it quantifies the average linear dependence of corresponding columns of blocks associated with different countries\footnote{\blue In a first version of the manuscript, the $l_2$ norm was adopted as dissimilarity measure. Nevertheless, it was verified numerically that AACD produces better clustering results for what concerns the successive application of MC, since the $l_2$ norm evaluates as dissimilar two highly correlated blocks whose entries have different sizes. The same issue holds for the $l_1$ norm.}. The adoption of this particular dissimilarity measure also i) provides an additional motivation for the application of hierarchical clustering instead} {\blue of a different clustering technique, since such a distance does not satisfy the triangle inequality, which is not required by hierarchical clustering but is required, e.g., by $k$-means; ii) operates as a data pre-processing step because, differently from the $l_1$ and $l_2$ norms, it is not affected by each country's average dimension.} 

In this work, we apply {\blue hierarchical clustering} 
to all countries in the I/O matrix -- except {\blue a specific} country $h$ (respectively, {\blue a specific country} $k$) -- in terms of what they use from (or what they sell to) country $h$ (respectively, $k$). In other words, we consider, {\blue for various years}, the submatrices $\bold{T}^{h,.}$ and $\bold{T}^{.,k}$ defined in Subsection \ref{sec:inoutmodel} to compare {\blue any} two countries {\blue $c_1$ and $c_2$} {\blue (different, respectively, from $h$ and $k$)} in terms of what they use from (or what the sell to) {\blue  the} specific country {\blue $h$ or $k$} (for illustrative purposes, in the simulations reported in the {\blue  application}, Italy is chosen as such a {\blue  specific} country).

Clustering is often used as a preliminary data pre-processing step to a successive supervised learning task. In the present context, clustering is used as a pre-processing step for MC applied to a suitable submatrix of an I/O table\footnote{Other applications of MC to networks include \cite{Nguyenetlal2019}, where this method was used to recover sensor maps in 2-D or 3-D Euclidean spaces from local or in any case partial sets of pair-wise distances. Up to the authors' knowledge, MC was applied in the literature to I/O matrices 
only in \cite{WenXuWen2014,XuLinHeWen2014}. In \cite{WenXuWen2014}, it was used as a pre-processing step for robust linear optimization, for a problem whose coefficient matrix is a partially-observable {\blue I/O} table. In \cite{XuLinHeWen2014}, MC was applied to predict zero entries in an I/O matrix, based on a set of observed entries mainly made by zeros.}. The idea is that MC is expected to perform better if the submatrix refers to countries belonging to the same cluster. This expectation is justified by one of the reasons provided in the literature as a motivation for the effectiveness of MC \cite{Hastie2015}, {\blue  which is summarized as follows.} 
{\blue Given a subset of observed entries of a matrix ${\bf M} \in \mathbb{R}^{m \times n}$, MC works by finding a suitable low-rank approximation (say, with rank $r$) of ${\bf M}$, by assuming the following model:
\begin{equation}\label{eq:matrix_completion1}
{\bf M}= {\bf C} {\bf G}^T + {\bf E}\,,
\end{equation}
where ${\bf C} \in \mathbb{R}^{m \times r}$, ${\bf G} \in \mathbb{R}^{n \times r}$, whereas ${\bf E} \in \mathbb{R}^{m \times n}$ is a matrix of modeling errors.} The rank-$r$ approximating matrix ${\bf C} {\bf G}^T$ is found by solving a suitable optimization problem (see , e.g., Equation (\ref{eq:matrix_completion2}) reported later). 
{\blue Equation (\ref{eq:matrix_completion1}) can be written element-wise as
\begin{equation}
M_{i,j}=\sum_{l=1}^r C_{i,l} G_{j,l} + E_{i,j}\,.
\end{equation}
Often, $C_{i,l}$ is interpreted as the degree of membership of row $i$ of matrix ${\bf M}$ to some ``latent'' cluster $l$ (for a total of $r$ such clusters), and $G_{j,l}$ as the prediction of an element in column $j$ of matrix ${\bf M}$, conditioned on its row $i$ belonging to cluster $l$\footnote{\blue It is worth mentioning that estimates ${\hat{\bf C}}$ and ${\hat{\bf G}}$ of the matrices ${\bf C}$ and ${\bf G}$ can be obtained as a by-product of the singular value decomposition ${\hat{\bf M}}={\bf U}({\hat{\bf M}}) {\bf \Sigma}({\hat{\bf M}}) \left({\bf V}({\hat{\bf M}})\right)^T$ of the matrix ${\hat{\bf M}}$ produced as output by a MC algorithm (e.g., one can set ${\hat{\bf C}}:={\bf U}({\hat{\bf M}})$ and ${\hat{\bf G}}:= {\bf V}({\hat{\bf M}}) \left({\bf \Sigma}({\hat{\bf M}})\right)^T$).}. As an example, in the application of MC to collaborative filtering for movie ratings, which is reported in \cite{Hastie2015},  $i$ in Equation (\ref{eq:matrix_completion1}) denotes a specific person, $j$ a specific movie, whereas $l$ may be interpreted as a specific movie genre. In our application, ${\bf M}$ is composed of several cross-country blocks (coming from I/O tables in different years), whereas $i$ and $j$ refer, respectively, to an input sector of a country and an output sector/final user of another country. Moreover, $l$ may be interpreted as a specific ``latent'' cluster, possibly discovered by the MC algorithm. It is argued in the present paper that such a discovery can be helped by a pre-processing step, in which a suitable clustering technique is used to restrict the attention to a subtable (made of ``similar'' countries either in input or output) of the original I/O table.}

{\blue It is worth observing that}, in order for MC to work properly in the case of {\blue an I/O} table (possibly partially observed for a set of  consecutive years), it can be useful to apply it not necessarily to the whole such matrix, but to its suitable submatrix determined by a pre-preprocessing step of clustering (hoping that the resulting clusters will be not so different from the latent ones). Intuitively, missing blocks of an I/O table that, thanks to historical data in the past years, are expected to be similar to the other observed blocks in the current year, could be reconstructed more effectively than missing blocks in the current year that, again based on historical data in the past years, are expected to be less similar to the other observed blocks in the same year. Another reason to focus the analysis on a submatrix of an I/O table is that solving the MC problem becomes computationally more expensive as the size of the matrix ${\bf M}$ increases.

In the work, we consider the following formulation for the MC optimization problem, which was investigated theoretically in \cite{Mazumder2010}:
\begin{equation}\label{eq:matrix_completion2}
\underset{{\bf \blue \hat{M}} \in \mathbb{R}^{m \times n}}{\rm minimize} \left(\frac{1}{2} \sum_{(i,j) \in \Omega^{\rm tr}} \left(M_{i,j}-{\blue \hat{M}}_{i,j} \right)^2 + \lambda \|{\bf \blue \hat{M}}\|_*\right)\,,
\end{equation}
where $\Omega^{\rm tr}$ (which, using a machine-learning expression, may be called training set) is a subset of pairs of indices $(i,j)$ corresponding to positions of known entries of $\bf M$, ${\bf \blue \hat{M}}$ is the completed matrix (to be optimized), $\lambda \geq 0$ is a regularization constant, and $\|{\bf \blue \hat{M}}\|_*$ is the nuclear norm of the matrix ${\bf \blue \hat{M}}$, i.e., the sum of all its singular values.
The regularization constant $\lambda$ controls the trade-off between fitting the known entries of the matrix ${\bf M}$ and achieving a small nuclear norm. The latter requirement is often related to getting a small rank of the optimal matrix ${\bf \blue \hat{M}}^\circ$, which follows by geometric arguments similar to the ones typically adopted to justify how the classical LASSO (Least Absolute Shrinkage and Selection Operator) penalty term achieves effective feature selection in linear regression \cite{Tibshirani1996}.

The optimization problem (\ref{eq:matrix_completion2}) can be also written as
\begin{equation}\label{eq:matrix_completion3}
\underset{{\bf \blue \hat{M}} \in \mathbb{R}^{m \times n}}{\rm minimize} \left(\frac{1}{2} \|{\bf P}_{\Omega^{\rm tr}}({\bf M})-{\bf P}_{\Omega^{\rm tr}}({\bf \blue \hat{M}})\|_F^2 + \lambda \|{\bf \blue \hat{M}}\|_*\right)\,,
\end{equation}
where, for a matrix ${\bf Y} \in \mathbb{R}^{m \times n}$, 
\begin{equation}
(P_{\Omega^{\rm tr}}({\bf Y}))_{i,j}:=\begin{cases} Y_{i,j} & {\rm if\,\,} (i,j) \in \Omega^{\rm tr}\,, \\
0 & {\rm if\,\,} (i,j) \notin \Omega^{\rm tr}
\end{cases}
\end{equation}
represents the projection of ${\bf Y}$ onto the set of positions of observed entries of the matrix ${\bf M}$, and
$\|{\bf Y}\|_F$ denotes the {\blue Euclidean} norm of ${\bf Y}$ (i.e., the square root of the summation of squares of all its entries).

It is shown in \cite{Mazumder2010} that the optimization problem (\ref{eq:matrix_completion3}) can be solved by applying the following Algorithm \ref{alg:1}, named Soft Impute therein\footnote{Compared to the original version, here we have included a maximal number of iterations $N^{\rm it}$, which can be helpful to reduce the computational effort when one has to run the algorithm multiple times, e.g., for several values of the regularization constant $\lambda$.}. {\blue  This is a state-of-the-art algorithm in the MC field.}

 \begin{algorithm}[H]\label{alg:1}
\SetAlgoLined
\KwData{Partially observed matrix ${\bf P}_{\Omega^{\rm tr}}({\bf M})$ , regularization constant $\lambda \geq 0$, tolerance $\varepsilon \geq 0$, maximal number of iterations $N^{\rm it}$}
\KwResult{Completed matrix ${\bf \blue \hat{M}}_\lambda \in \mathbb{R}^{m \times n}$}
  \begin{enumerate}
  \item Initialize ${\bf \blue \hat{M}}$ as ${\bf \blue \hat{M}}^{\rm old}={\bf 0} \in \mathbb{R}^{m \times n}$
 \item Repeat for at most $N^{\rm it}$ iterations:
 \begin{enumerate}
 \item Set ${\bf \blue \hat{M}}^{\rm new}\leftarrow {\bf S}_\lambda\left({\bf P}_{\Omega^{\rm tr}}({\bf M})+{\bf P}_{\Omega^{\rm tr}}^{\perp}({\bf \blue  \hat{M}}^{\rm old}) \right)$
 \item If $\frac{\|{\bf \blue \hat{M}}^{\rm new}-{\bf \blue \hat{M}}^{\rm old} \|_F^2}{\|{\bf \blue \hat{M}}^{\rm old}\|_F^2} {\blue \leq} \varepsilon$, exit\
 \item Set ${\bf \blue \hat{M}}^{\rm old}\leftarrow{\bf \blue \hat{M}}^{\rm new}$
 \end{enumerate}
 \item Set ${\bf \blue  \hat{M}}_\lambda\leftarrow{\bf \blue  \hat{M}}^{\rm new}$
  \end{enumerate}
 \caption{{\bf Soft Impute} \cite{Mazumder2010}}
\end{algorithm}

In Algorithm \ref{alg:1}, for a matrix ${\bf Y} \in \mathbb{R}^{m \times n}$, ${\bf P}_{\Omega^{\rm tr}}^{\perp}({\bf Y})$ represents the projection of ${\bf Y}$ onto the complement of $\Omega^{\rm tr}$, whereas
\begin{equation}\label{eq:operatorS}
{\bf S}_\lambda({\bf Y}):= {\bf U} {\bf \Sigma}_\lambda {\bf V}^T\,,
\end{equation}
being
\begin{equation}
{\bf Y}={\bf U} {\bf \Sigma} {\bf V}^T
\end{equation}
(with ${\bf \Sigma}={\rm diag} [\sigma_1,\ldots,\sigma_r]$) the singular value decomposition of ${\bf Y}$, and 
\begin{equation}\label{eq:Dlambda}
{\bf \Sigma}_\lambda:={\rm diag} [(\sigma_1-\lambda)_+,\ldots,(\sigma_r-\lambda)_+]\,,
\end{equation}
with $t_+:=\max(t,0)$.

In \cite{LiZhou2017}, a particularly efficient implementation of the operator ${\bf S}_{\lambda}(\cdot)$ defined in Equation (\ref{eq:operatorS}) is proposed (by means of the MATLAB function {\tt svt.m} reported therein), which is based on the determination of only the singular values $\sigma_i$ of ${\bf Y}$ that are larger than $\lambda$, and of their corresponding left-singular vectors ${\bf u}_i$ and right-singular vectors ${\bf v}_i$. Indeed, all the other singular values of ${\bf Y}$ are annihilated in ${\bf \Sigma}_\lambda$ (see Equation (\ref{eq:Dlambda})).

{\blue For this work} we combine the original MATLAB implementation of Soft Impute from  \cite{Mazumder2010} with the MATLAB function {\tt svt.m} from \cite{LiZhou2017}. Moreover, to avoid overfitting, we select the regularization constant $\lambda$ via the following validation method. First, the set of positions of unobserved entries of the matrix ${\bf M}$ is divided randomly into a validation set $\Omega^{\rm val}$ (about $25\%$ of the positions of the unobserved entries) and a test set $\Omega^{\rm test}$ (the positions of the remaining entries). {\blue In the present context of application of MC to I/O subtables, the union of the validation and test sets correspond to a block which is artificially obscured (but which is still available as a ground truth), whereas the training set corresponds to the positions of all the remaining entries of the submatrix considered.} It is worth observing that, by the construction above, there is no overlap among the training, validation, and test sets. Then, the optimization problem (\ref{eq:matrix_completion3}) is solved for several choices $\lambda_k$ for $\lambda$, exponentially distributed as {\blue  $\lambda_k=2^{k/2-10}$}, for $k=1,\ldots,{\blue  40}$. For each $\lambda_k$, the Root Mean Square Error (RMSE) of matrix reconstruction on the validation set is computed as
\begin{equation}
RMSE_{\lambda_k}^{\rm val}:=\sqrt{\frac{1}{|\Omega^{\rm val}|}\sum_{(i,j) \in \Omega^{\rm val}} \left(M_{i,j}-{\blue \hat{M}}_{\lambda_k,i,j} \right)^2}\,,
\end{equation}
then the choice $\lambda_k^\circ$ that minimizes $RMSE_{\lambda_k}^{\rm val}$ for {\blue  $k=1,\ldots,40$} is found. Finally, the RMSE of matrix reconstruction on the test set is computed in correspondence of the optimal value $\lambda_k^\circ$ as
\begin{equation}
RMSE_{\lambda_k^\circ}^{\rm test}:=\sqrt{\frac{1}{|\Omega^{\rm test}|}\sum_{(i,j) \in \Omega^{\rm test}} \left(M_{i,j}-{\blue  \hat{M}}_{\lambda_k^\circ,i,j} \right)^2}\,.
\end{equation}

{\blue A similar expression holds for the RMSE of matrix reconstruction on the training set, in correspondence of the optimal value $\lambda_k^\circ$:
\begin{equation}
RMSE_{\lambda_k^\circ}^{\rm tr}:=\sqrt{\frac{1}{|\Omega^{\rm tr}|}\sum_{(i,j) \in \Omega^{\rm tr}} \left(M_{i,j}-{\blue  \hat{M}}_{\lambda_k^\circ,i,j} \right)^2}\,.
\end{equation}
}

It is worth mentioning that in our application of MC to submatrices of I/O tables, the very few missing/negative 
entries of such submatrices (when present) are replaced by zeros before running Algorithm \ref{alg:1}. The tolerance is chosen as $\varepsilon=10^{-9}$. Moreover, when convergence is not achieved, in order to reduce the computational time, the algorithm is stopped after {\blue $N^{\rm it}=500$} iterations.
An additional post-processing step is included, thresholding to $0$ any negative element (when present) of the completed submatrices\footnote{\blue It is worth noticing that our method is applicable also in case of negative elements. Although real-world I/O tables contain sometimes a very small percentage of negative entries (see, e.g., \cite{UN2018} for some motivations behind the possible presence of such entries), ``ideally'' they should have no negative entries at all (recall the definitions of $z_{i,j}$ and $f_{i,j}$ in Subsection \ref{sec:inoutmodel}). Hence, this a-priori knowledge about non-negativity of an ``ideal'' I/O table is used in the paper as a post-processing step to possibly improve the MC predictions. By the way, we have found in all our numerical results that, indeed, this post-processing step always improved the MC predictions.}. {\color{black} In the following, in order to avoid introducing new notation, the expression ${\bf \hat{M}}_{\lambda_k}$ is actually used to denote each post-processed MC output.}

{\blue Finally, in order to quantify the MC performance, we also use the following second metric, which is known in the literature as Symmetric Mean Absolute Percentage Error (SMAPE)\footnote{\blue As an alternative, one could use a more recent weighted version of the SMAPE, known as SWAPE (see, e.g., \cite{Valderas-Jaramilloetal2019}).}. Differently from the RMSE, it takes into account the relative error of reconstruction. Its definition for the validation set is as follows:
\begin{equation}
SMAPE_{\lambda_k}^{\rm val}:=\frac{100}{|\Omega^{\rm val}|}\sum_{(i,j) \in \Omega^{\rm val}} \frac{\left|M_{i,j}-\hat{M}_{\lambda_k,i,j} \right|}{\left|M_{i,j}\right|+\left|\hat{M}_{\lambda_k,i,j} \right|}
\end{equation}
(the constant $100$ is used\footnote{\blue In some references, it is replaced by the constant $200$.} to make the metric range from $0$ to $100$; when both the numerator and the denominator are equal to $0$, the ratio is assumed to be equal to $0$, too). 
Similar definitions hold for the test set and the training set. Again, the metric is first evaluated on the validation set for different choices of $\lambda_k$, then it is computed on both the training and test sets in correspondence of the value of $\lambda_k$ that minimizes the SMAPE on the validation set. Differently from the RMSE, this metric is not directly related to the optimization problem (\ref{eq:matrix_completion2}) solved by MC, nor to the choice of AACD as the dissimilarity measure used by hierarchical clustering in the present article. Hence, for this metric, differently from the RMSE, one does not necessarily expect an improvement in MC performance when moving from ``dissimilar'' to ``similar'' countries.
}

\section{Application of clustering and matrix completion}\label{sec:application}

{\blue In this section we present an application of the proposed methodological approach to WIOD data.
Before presenting the application to real data, some aspects need to be considered.
In order to generalize our method to any I/O table (whether they are either industry-by-industry or product-by-product ones), in Subsection \ref{sec:structure} we show how the structure of WIOD tables -- described in Subsection \ref{sec:description} -- is similar to those of alternative I/O tables.
Then, in Subsection \ref{sec:preprocessing} some simulation results are reported, based on synthetic I/O matrices generated from the raw WIOD tables, in order to discuss the benefits from applying MC to proper pre-processed data and to determine the optimal number of clusters for the choice of similar and dissimilar countries. Finally, in Subsection \ref{sec:forecasting} we provide full details on how we operatively apply the method to real data.

It is important to clarify why we use WIOD matrices in our application. 
According to \cite{Timmer2015}, WIOD represents a real improvement over other databases (such as EORA, EXIObase, FIGARO, and OECD tables) for several reasons: i) its data are extrapolated by certified national statistical institutions, ii) to determine data from the rest of the world, data from United Nations (UN), International Monetary Funds (IMF) and other international institutions were used, iii) all versions of WIOD are available for free from the website \url{www.wiod.org}. 
Compared with other I/O datasets, we choose WIOD tables for two additional reasons: first, because they are characterized by  a  quite  large  coverage  period;  second,  because  their  size  is quite representative of the ones of the other tables (i.e., both the number of countries considered and the sector disaggregation are neither too small, nor too large with respect to the other tables). The latter issue is also important for  computational  reasons,  because  the  application  of  MC is typically slow for quite large matrices. }



\subsection{Data}\label{sec:description}
The WIOD database was constructed and developed in the seventh framework programme funded by the European Commission in 2009, and is licensed under a Creative Commons Attribution 4.0 International-license. 
From a technical point of view, WIOD tables are built up from public databases coming from different national and international statistics offices. 
Currently, there exist two releases of WIOD: the 2013 and the 2016 release. The latest release covers the period between 2000 and 2014 and 43 among the most relevant countries in the world: EU-28 (including the UK), Australia, Brazil, Canada, Switzerland, China, Indonesia, India, Japan, South Korea, Mexico, Norway, Russia, Turkey, Taiwan, USA\footnote{\blue In WIOD tables, countries are represented by ISO-3166-1 alpha-3 codes.}. The provided yearly tables are splitted {\blue into} 56 different macro-industries, classified according to the International Standard Industrial Classification Revision 4 (ISIC Rev. 4), and their pair-wise combinations\footnote{In contrast, {\blue the} 2013 release covers the period 1995-2011, considering only 40 countries and 38 macro-sectors.}. Moreover, 5 final aggregated outputs {\blue -- still classified according to ISIC Rev. 4 --}  are reported in the tables. Finally,  an estimation for the remaining non-covered part of the world economy (called ``Rest Of the World'', ROW) is reported (see \cite{Timmer2015,Timmer2016} for the details). Thereby, using WIOD can help to perform excellent and detailed input/output analyses (some very recent applications being provided, e.g., in \cite{BhattacharyaBhandariBairagya2020,ChenWuGuoMengLi2019,WangDingGuanZia2020,XuLiang2019}). 

{\blue In the following subsection we put WIOD tables in comparison with OECD and FIGARO ones. The latest release of OECD Inter-Country Input-Output (ICIO) tables dates back to 2018. ICIO tables report yearly data from 2005 to 2015 among 64 countries (including ROW) and 36 industries (products).
FIGARO tables, also known as EU inter-country Supply, Use and Input-Output tables (EU IC-SUIOTs) are available on a yearly basis from 2010 to 2019 and display exchanges among EU economies, the United Kingdom and the United States and among 64 industries (products).}

\subsection{Characterization of the data}\label{sec:structure}

Here we provide a descriptive analysis of WIOD tables 
{\blue in comparison with OECD and FIGARO (whose data are available in the same timespan of WIOD ones)} in terms of their within-country and cross-country value distributions, level of sparsity (i.e., percentage of zeros in each {\blue subtable}), and separation between values of transactions between so-called large-to-large and small-to-small country pairs. 
These {\blue analyses are made on industry-by-industry tables, as the product-by-product ones depart from the former to just a little extent (Pearson correlation of 0.9958 for FIGARO, year: 2010)}, and are reported for the years from 2010 to 2014, which is the time span considered in our panel analysis. 
{\blue After having removed ROW rows and columns, and also rows related to taxes and value added, WIOD tables count for 2408 rows (which is the product of 56 intermediate industries and 43 countries) and 2623 columns (which is the result of 56 intermediate industries plus 5 final outputs, multiplied by 43 countries). The 56 macro-industries (which are produced by the aggregation of various micro-industries) are reported in the following order: primary industry appears in the first $4$ positions, followed by secondary industry ($18$ positions), and finally by tertiary industry ($34$ positions). 
} 
About the sparsity of WIOD I/O matrices, Table \ref{tab:zeros} shows that, consistently over the years, the percentage of zeros is between 17\% and 18\%. {\blue Moreover, results are consistent if compared to those of OECD tables while the number of zeros is slightly larger in FIGARO ones.}
\begin{table}[]
    \centering
    \begin{tabular}{c|ccccc}
    \hline
   Source &   2010   & 2011 & 2012 & 2013 & 2014  \\
      \hline
  WIOD & 17.904\%    &17.924\%  & 17.788\%  & 17.854\%  &17.821\%   \\ 
OECD & 21.480\%    & 21.195\%  &   21.440\% & 21.354\%  & 21.230\% \\ 
FIGARO & 39.293\%    & 38.408\%  & 37.244\%  & 37.563\%  & 37.270\%   \\ 
    \hline
        \end{tabular}
    \caption{\blue Percentage of zero values in the I/O tables, by year.}
    \label{tab:zeros}
\end{table}



\begin{figure}
\begin{subfigure}{0.95\textwidth}
  \centering
 \includegraphics[scale=0.132]{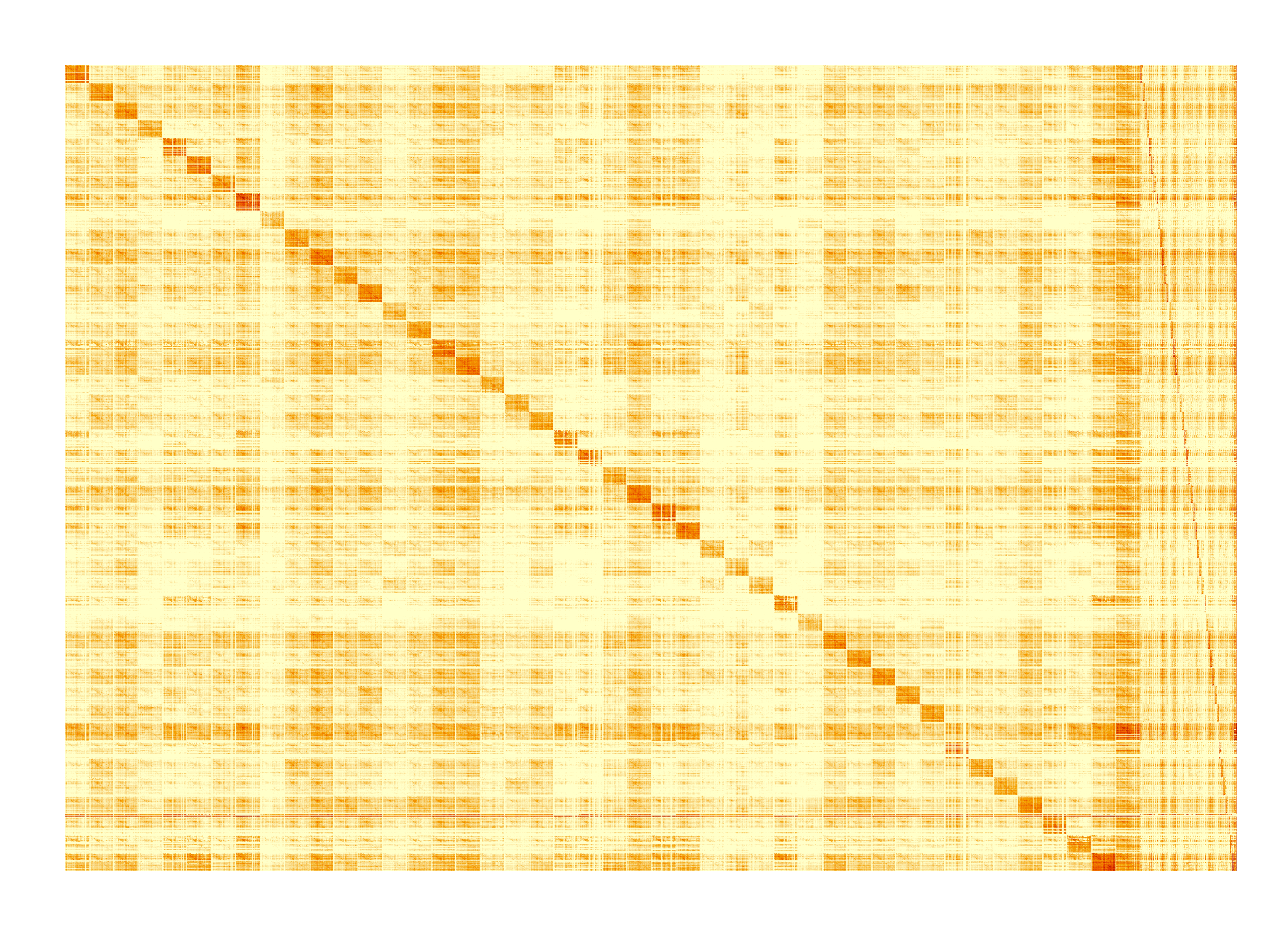}
  \includegraphics[scale=0.132]{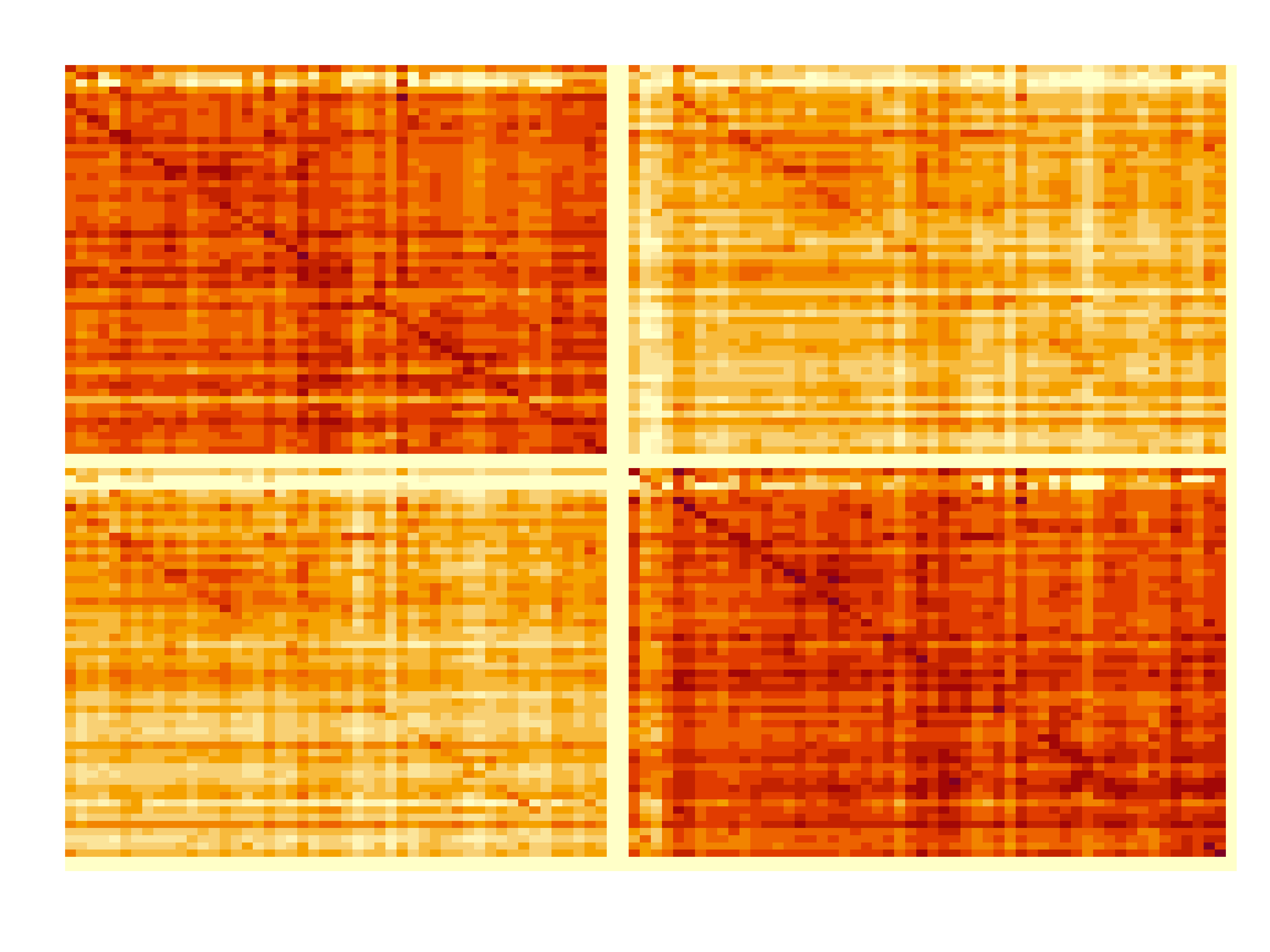}
  \caption{{\blue WIOD (left). Zoom of the WIOD submatrix composed of rows and columns of Italy and Spain (right).}}
  \label{fig:matrix_wiod}
\end{subfigure}
\begin{subfigure}{0.95\textwidth}
  \centering
 \includegraphics[scale=0.132]{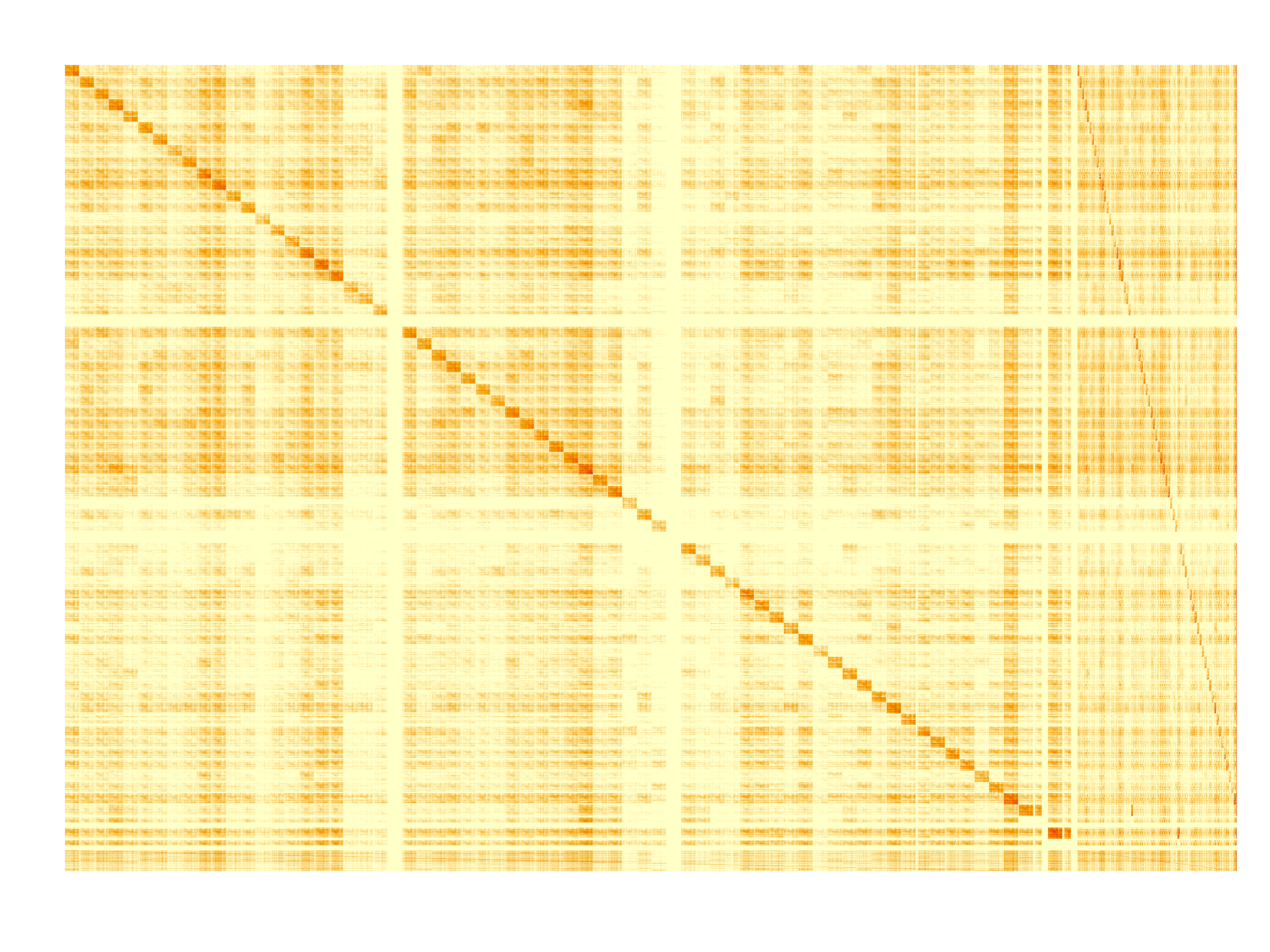}
  \includegraphics[scale=0.132]{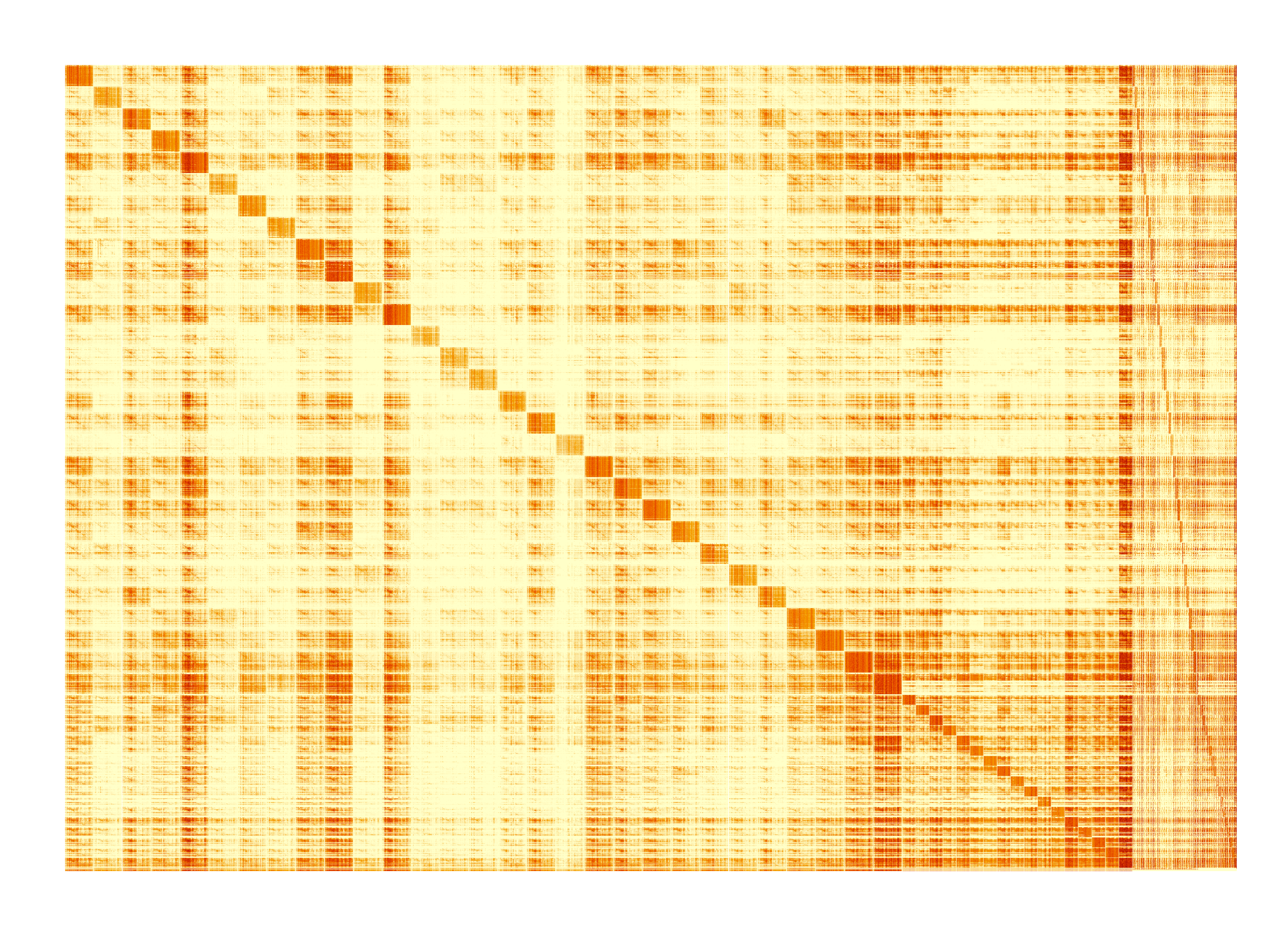}
  \caption{{\blue OECD (left). FIGARO (right).}}
  \label{fig:matrix_oecd}
\end{subfigure}
\caption{\blue Colored visualization of the elements of the complete I/O tables (year: 2013). For a better visualization, a logarithmic scale is used in the figures.}
\label{fig:matrix}
\end{figure}

{\blue Figure \ref{fig:matrix} shows, as an example, a colored visualization of the elements of the 2013 I/O tables where each colored rectangle corresponds to the exchange between country-industry pairs. 
In its subfigures, final consumption is reported on the right extremes. 
The figure sheds lights on the fact that, consistently over the three compared I/O tables, the largest values (depicted in red) are concentrated in the domestic blocks (main diagonal blocks). 
Indeed, industries usually tend to consume (with respect to trade) products coming from their home country, for reasons such as higher proximity and safety (i.e., less uncertainty in terms of price, and more regularity in terms of supplies). Moreover, flows from a specific ``country-industry'' pair to the same ``country-industry'' pair (main diagonal of the tables) are generally much larger than the other flows (this holds especially for the case of secondary macro-sectors, as one can see from the right chart of Figure \ref{fig:matrix_wiod} representing exchanges within and between Italy and Spain for the case of WIOD). 
This issue is partially motivated by outsourcing, but especially by the fact that in some industries (e.g., manufacturing industries) there are several concatenated products in the production line, which in the case of WIOD tables are aggregated in the same macro-sector. 
Moreover, it can be noticed from the figure that secondary industrial products (e.g., ore, iron, oil, metals, technical equipment, manufacturing industry in general) are quite prone to international trade (see the corresponding parts of the main diagonals of the off-diagonal blocks, also called international trade blocks), whereas services are less traded internationally.
Finally, by looking indifferently to one of the I/O tables in Figure \ref{fig:matrix} and by comparing either single country blocks by row (supplying countries) or single country blocks by column (receiving countries), it is possible to notice how some countries are similar to each other according to how they are dependent with respect to other specific countries.

Overall, these evidences show that {\blue  real-world} I/O tables provided by different institutions share very similar characteristics and motivate us to work just on one kind among the analyzed matrices (WIOD tables, in our specific selection) and to generalize the results obtained using it over different I/O tables.
Moreover, as discussed also in Subsection \ref{sec:preprocessing}, all these evidences about the quite different structures and orders of magnitude of the elements of different portions of the I/O matrices motivate us to pre-process them before they undergo the application of MC.}


\subsection{{\blue Clustering} {\blue step and simulations}}\label{sec:preprocessing}

In the case of I/O tables, we argue in the following that a suitable pre-processing step is needed for an effective application of MC. Indeed, as discussed in Subsection \ref{sec:structure}, the main diagonal blocks in the I/O tables ({\blue also called domestic blocks, since} each of {\blue them} refers to trade inside the same country), and especially their entries which refer to exchanges within the same sector and the same country, are characterized by much larger values than the other blocks, which may cause {\blue troubles in the application of MC, as the quite different orders of magnitude could make it difficult for MC to have a good generalization capability on both kinds of blocks\footnote{\blue As shown in Appendix \ref{appendix:2}, indeed, the performance of MC on an I/O submatrix which considers simultaneously the blocks of intra-country and inter-country exchanges (whose orders of magnitude are highly different) is quite bad.}. Computational reasons (i.e., the need of performing a singular value decomposition step at each iteration of the Soft Impute algorithm) suggest to apply MC to a submatrix associated with a small subset of countries, as this reduces the size of that submatrix. Moreover, as discussed in Subsection \ref{sec:clustering_matrix_completion}, we argue that MC is more effective when it is applied to an I/O subtable made of  ``similar'' blocks (with respect to the case of ``dissimilar'' blocks).} 
For this reason, we have decided to apply MC to submatrices of WIOD tables obtained by excluding systematically the main diagonal blocks\footnote{\blue This is not necessarily a limitation of the proposed method. Indeed, as an alternative, one could exclude systematically the off-diagonal blocks (this is left for future research). The important issue is that the blocks kept have entries with similar orders of magnitude.}, {\blue where 
the selection of the countries associated with the submatrices is made by means of hierarchical cluster analysis. In this context, the choice of the number of clusters is determinant and so, in order to validate hierarchical clustering, we perform some simulation exercises, based on synthetic data\footnote{\blue Methods for generating synthetic I/O matrices can be found in \cite{Wangetal2015}, which used a cubic polynomial with coefficients generated from a standardized normal distribution. The work \cite{Paviaetal2009} added either a normally or a uniformly distributed disturbance term to six heterogeneous origin-destination matrices, whereas \cite{Fernandez-Vazquez} also added random terms to the elements of the I/O tables.}. 
We choose to generate our synthetic I/O matrices by adding a matrix of normally distributed random terms $\epsilon$ to the subset of interest of the WIOD dataset, where each element of the matrix $\epsilon_{i,j} \sim \mathcal{N}(0,sd_{\epsilon}), \forall i, \forall j$, $sd_{\epsilon}$ being a Gamma($\alpha$, $\beta$) such that different generated synthetic matrices display different levels of variability. Specifically, we choose $\alpha=1$ and $\beta= 1$.}

\begin{figure}[h!]
    \centering
    \includegraphics[scale=0.35]{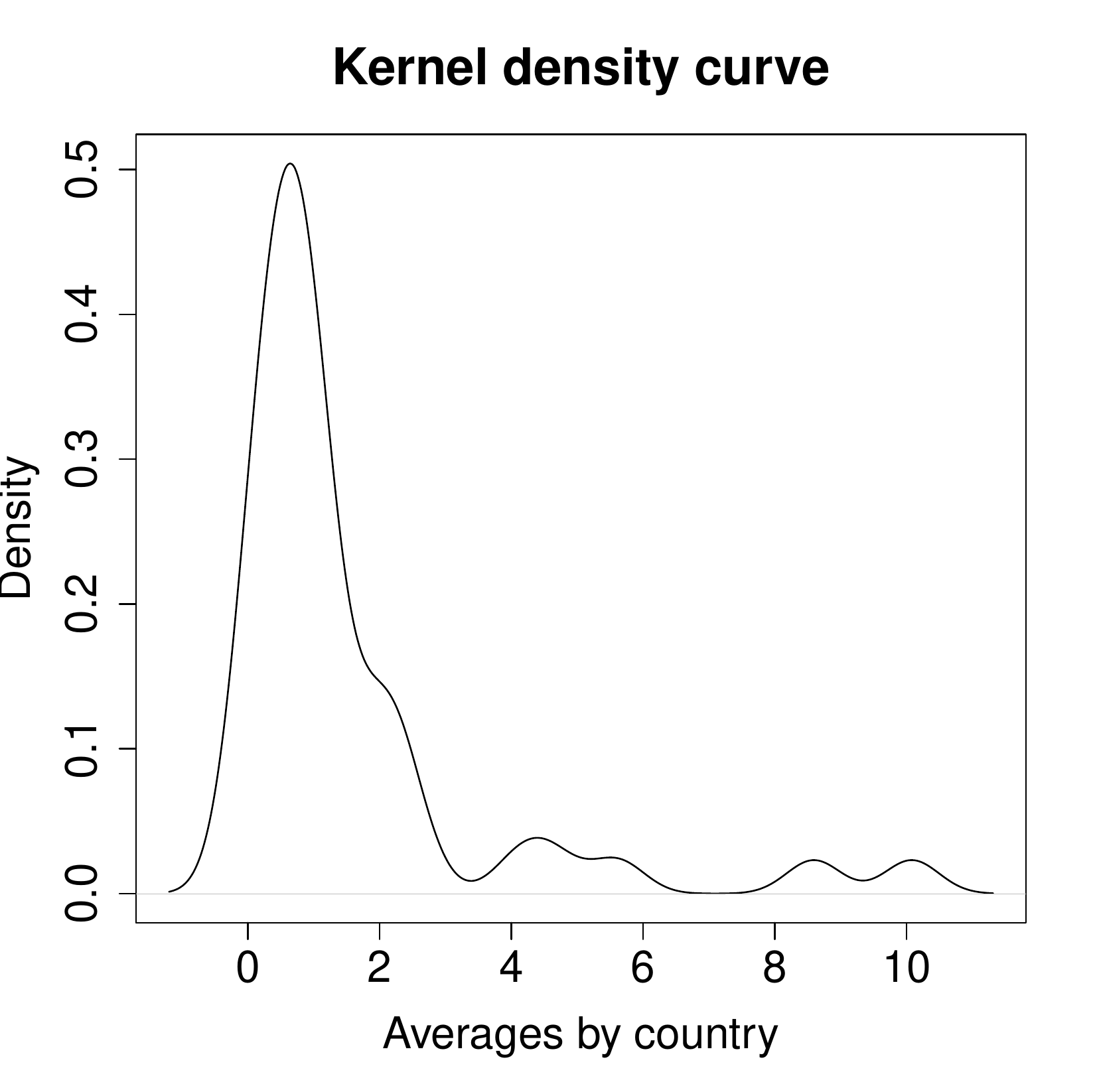}
       \includegraphics[scale=0.35]{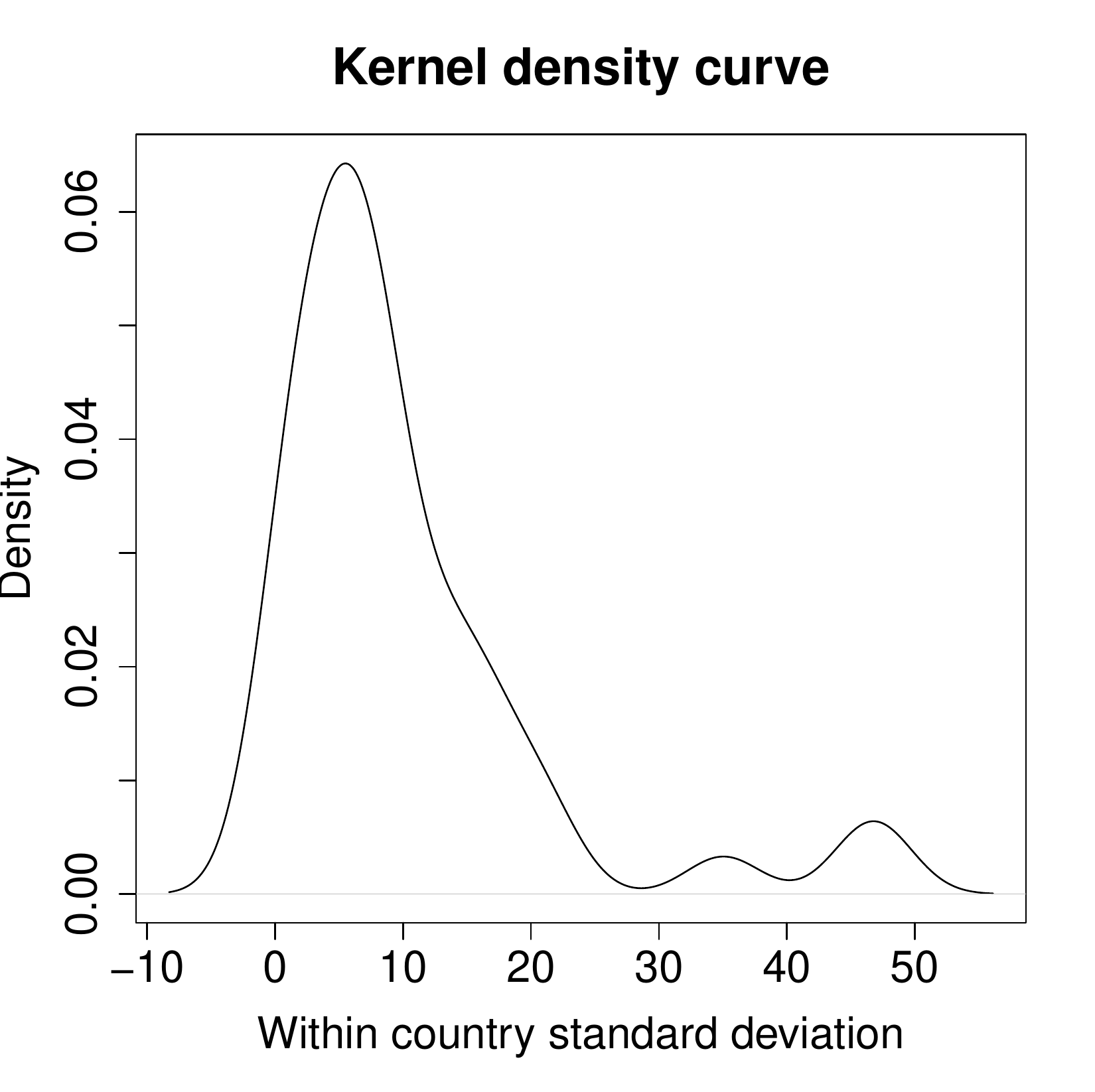}
    \caption{\blue Kernel density curves of the distribution of the average exchanges from Italy (in input), by country of output (\texttt{averages by country}) and of the within country standard deviation (\texttt{within country standard deviation}). All countries excluding Italy. WIOD data, year: 2010.}
    \label{fig:kernell_means}
\end{figure}

{\blue In order to evaluate the correct number of groups in terms of similarity either with respect to input from Italy or with respect to output to Italy, we simulate, respectively\footnote{\blue It is worth clarifying that, while the generalization of the simulation results to other I/O tables holds, the generalization to other reference countries different from Italy does not necessarily hold.}:
i) $N$ = 1000 synthetic I/O matrices of dimension $56 \times 2562$, where 56 is the number of industries in Italy, and 2562 is the product of the 61 industries (final sectors included) and the number of countries except Italy,
ii) $N$ = 1000 synthetic I/O matrices of dimension $2562 \times 56$, where 2562 is the product of the 61 industries and the number of countries except Italy and 56 is the number of industries in Italy.
In order to select the number of clusters\footnote{Alternative criteria for the choice of the optimal number of clusters are the Elbow method \cite{Thorndike1953}, the average silhouette method \cite{Rousseeuw1987} and the ``Gap'' index \cite{Tibshirani2001}} we consider the ratio $\frac{WSS}{TSS}$,  where $WSS$ is the ``Within-cluster sum of squares'' and $TSS$ is the ``Total sum of squares''. 
More in details, the optimal number of clusters is the minimum such number for which $\frac{WSS}{TSS} < K$, where $K$ is a cut-off that we set to $0.5$.

{\blue It is worth observing that the volume of inputs (outputs) taken from (given to) a specific country is not equally distributed all along other considered countries. 
Table \ref{tab:means} and Figure \ref{fig:kernell_means} show, for the case of inputs from Italy in 2010, a strong dispersion both in terms of averages by countries and in terms of within countries standard deviations.
These evidences further motivated us to 
use AACD as a dissimilarity measure for hierarchical clustering.} 
Table \ref{tab:numclus_wss} 
reports the results of the simulation using (stacked) years from 2010 to 2013. 
}



\begin{table}[h!]
\setlength{\tabcolsep}{3pt}
\begin{center}
\begin{tabular}{ l|cccccc } 
\textbf{Statistic} &  \textbf{Min.} &\textbf{ 1st Qu.} &  \textbf{Median} &    \textbf{Mean} &\textbf{3rd Qu.}&    \textbf{Max.}\\ 
\hline
\textbf{Averages by country} & 0.034 & 0.488 & 0.837 & 1.567 & 1.780 & 10.080\\
\textbf{Within country st. dev.} & 0.354 & 4.034 & 6.917 &10.254& 13.097 &47.461\\ 
\end{tabular}
\end{center}
\caption{\blue Distribution of the average exchanges from Italy (in input), by country of output (\texttt{averages by country}) and of the within country standard deviation (\texttt{within country st. dev.}). All countries excluding Italy. WIOD data, year: 2010.}\label{tab:means}
\end{table}

\begin{table}[h!]
\setlength{\tabcolsep}{3pt}
\begin{center}
\begin{tabular}{ |c|c|c|c|c|c|c| } 
\hline
\textbf{Direction }&  \textbf{Min.} &\textbf{ 1st Qu.} &  \textbf{Median} &    \textbf{Mean} &\textbf{3rd Qu.}&    \textbf{Max.} \\ 
\hline
\textbf{Input} & 14  & 21 & 21  & 21.070 &  21 &  28 \\ 
\textbf{Output} & 16&   22&   22&   21.971 &   22 &   28 \\ 
\hline
\end{tabular}
\end{center}
\caption{\blue Simulation results for the choice of the  optimal number of clusters. $N$=1000 replications of hierarchical clustering with complete linkage criterion and AACD as dissimilarity measure, on I/O synthetic matrices derived from WIOD original data. Adopted criterion: $\frac{WSS}{TSS}$. More precisely, we chose the minimum number of clusters such that $\frac{WSS}{TSS} <0.5$. Considering Italy in input and all other countries in output (\texttt{input}), and considering Italy in output and all other countries in input (\texttt{output}). Years: from 2010 to 2013 (4 years).}\label{tab:numclus_wss}
\end{table}

{\blue 
 According to the results in Table \ref{tab:numclus_wss} we obtain a fairly good clustering with around 21/22 groups\footnote{{\blue It is worth noting here that the optimal number of clusters for the matrix where Italy is in input is a bit smaller than the optimal number of clusters for the matrix where Italy is in output. This difference may be due to the asymmetry between inflow and outflow trade data, motivated by the fact that countries, typically, import goods that they do not have, and they export goods that they produce.}}. Following these results, in Subsection \ref{sec:similardissimilar} we are going to consider as similar those groups of countries belonging to the same cluster in a configuration with 21 groups (when Italy is in input) or 22 groups (when Italy is in output). 
}

{\blue Finally, MC has been applied to I/O subtables associated with suitable groups of ``fictitious'' similar and dissimilar countries obtained from synthetic matrices, and the performance has turned out to be similar to the one obtained in the case of real data, which is discussed extensively in Subsection \ref{sec:similardissimilar}. To ease the reading of the work, results for the case of synthetic matrices are reported later in Appendix \ref{appendix:4}.}

\subsection{Application to forecasting}\label{sec:forecasting}

In this subsection, we consider the application of MC to forecasting. {\blue Our focus is not on the absolute RMSE, but on its percentage of reduction (in correspondence of the optimal value of the regularization parameter $\lambda$), with respect to a base case (represented by $\lambda \simeq 0$)}. The following situation is considered. Starting from WIOD tables relative to some consecutive years, the information  associated with a subset of countries is reported in a matrix, keeping only the off-main diagonal blocks, as done at the end of the previous subsection. Then, for one of these ordered pairs of countries, the information about the last year is obscured, and one tries to reconstruct it by MC. {\blue As} an example, Table \ref{tab:2} refers to the case in which the countries considered are France and Italy, and the years analyzed are 2010, 2011, 2012, 2013, and 2014. All the entries related to {\blue France imports in 2014 coming from Italy (i.e., input sectors are from Italy and intermediate/final outputs are from France)} are  obscured\footnote{\blue Here, the term ``obscured'' means ``not observed''. It is important to remark that this is conceptually different from ``set to $0$'' (although they may be set to $0$ at the initialization phase of a MC algorithm, as in the case of Soft Impute). To clarify how the MC optimization problem (\ref{eq:matrix_completion3}) deals with the obscured entries of the partially observed matrix ${\bf M}$, it is enough to look at the form of its objective function, which takes into account only the observed elements $M_{i,j}$ of that matrix. So, any initial assignment of values to the unobserved entries is irrelevant.}, then {\blue such entries} are reconstructed by MC\footnote{{\blue The reader is referred to Appendix \ref{appendix:3} for a technical motivation of the specific displacement of the blocks in Table \ref{tab:2}, which is obtained by comparing it with a possible alternative displacement.}}. The rationale behind this application is that WIOD tables are obtained by combining information coming from different sources, and these are not necessarily synchronized. So, one could combine the complete information available in the past with the partial one currently available, to predict currently missing elements.

\begin{table}[h!]
\setlength{\tabcolsep}{3pt}
\begin{center}
\begin{tabular}{ | c | c | } 
\hline
\multicolumn{2}{ | c | }{\it I/O, year} \\
\hline
FRA/ITA, 2010 & ITA/FRA, 2010\\ 
\hline
FRA/ITA, 2011 & ITA/FRA, 2011\\  
\hline
FRA/ITA, 2012 & ITA/FRA, 2012\\ 
\hline
FRA/ITA, 2013 & ITA/FRA, 2013\\ 
\hline
FRA/ITA, 2014 & \bf{ITA/FRA, 2014}\\ 
\hline
\end{tabular}
\end{center}
\caption{Structure of the WIOD submatrix used for the example reported in Figure \ref{fig:2}. Each block is made of $56$ rows for the inputs (corresponding to the $56$ intermediate sectors considered by WIOD), $56$ columns for the intermediate  outputs (one for each intermediate sector), and other $5$ columns for the final products. All the entries contained in the block highlighted in {\blue bold} are obscured and reconstructed by {\blue MC}.}\label{tab:2}
\end{table}

\begin{figure}
\begin{subfigure}{1\textwidth}
    \begin{center}
    \includegraphics[scale=0.27]{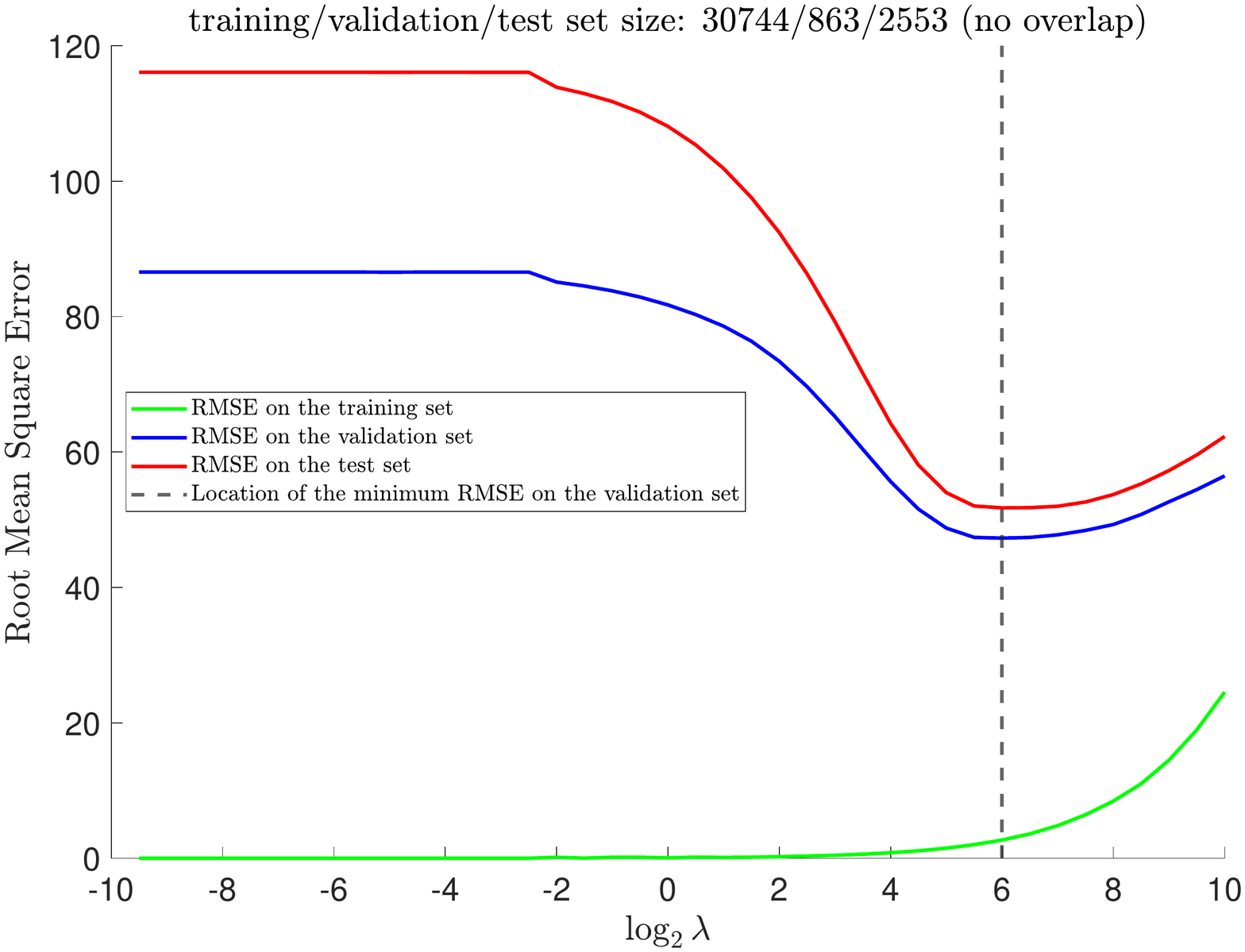}
    \vspace{-0.45cm}
    \caption{Results {\blue (expressed in terms of RMSE)} of the application of Algorithm \ref{alg:1} to the WIOD submatrix reported in Table \ref{tab:2}.}
    \label{fig:2}
    \end{center}
    \end{subfigure}
\begin{subfigure}{1\textwidth}
    \begin{center}
    \includegraphics[scale=0.27]{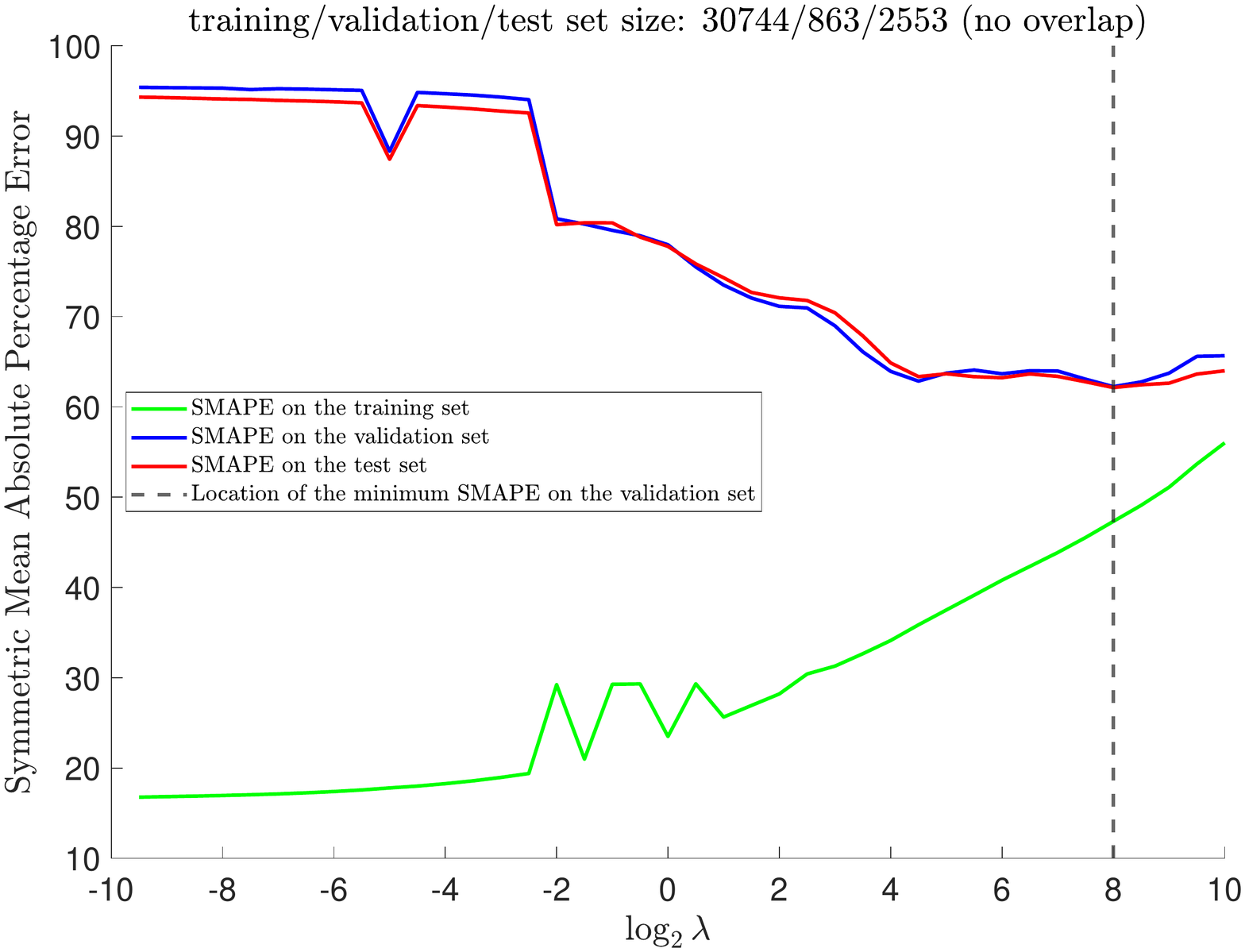}
    \vspace{-0.45cm}
    \caption{\blue Results (expressed in terms of SMAPE) of the application of Algorithm \ref{alg:1} to the WIOD submatrix reported in Table \ref{tab:2}.}
    \label{fig:2var}
    \end{center}
    \end{subfigure}
\begin{subfigure}{1\textwidth}
    \begin{center}
    \includegraphics[scale=0.27]{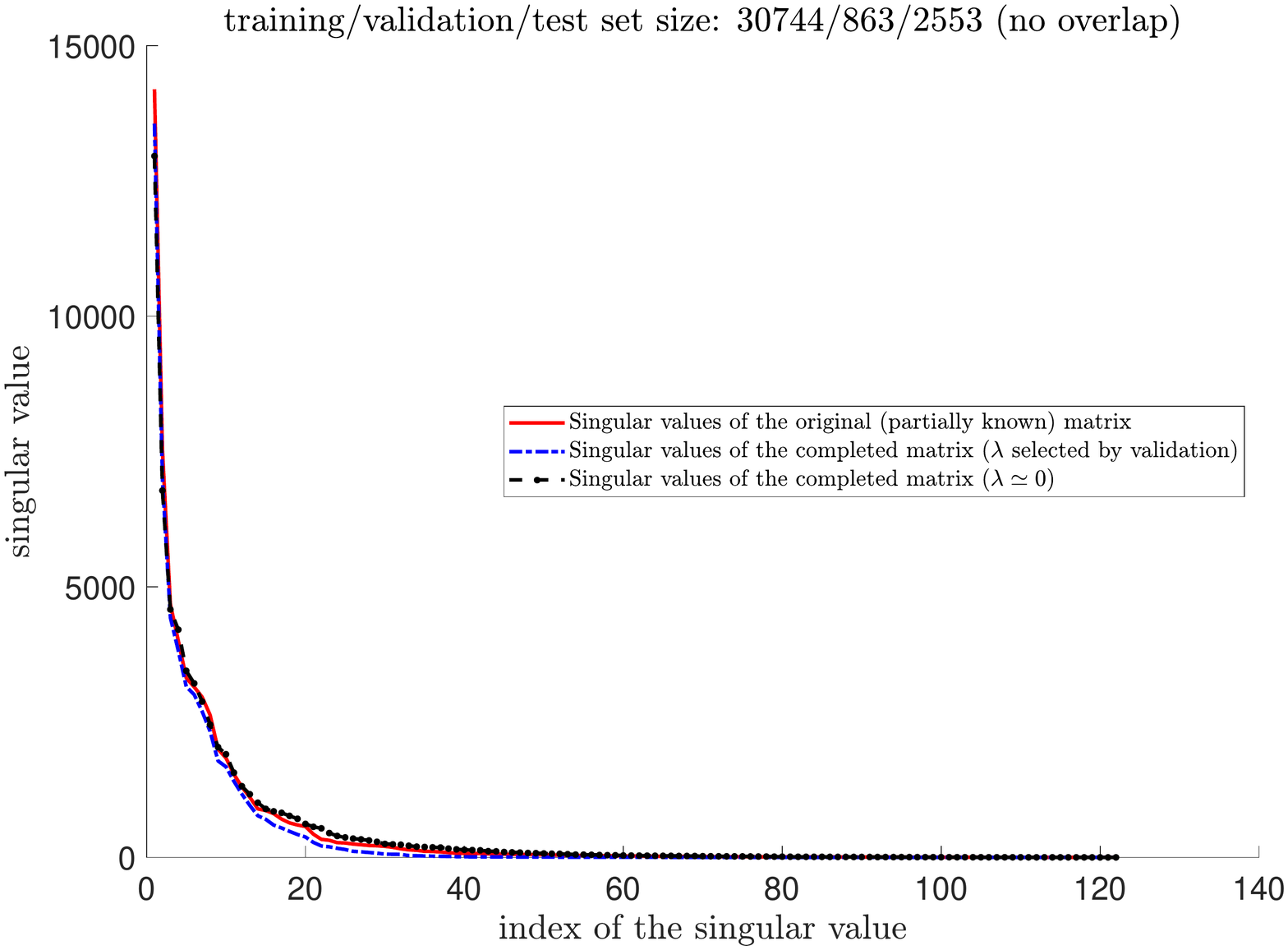}
    \vspace{-0.45cm}
    \caption{Singular values distribution of the WIOD submatrix reported in Table \ref{tab:2}, and the one of the completed submatrix produced 
    by Algorithm \ref{alg:1} for the optimal regularization constant {\blue (RMSE criterion)}.}
    \label{fig:3}
    \end{center}
\end{subfigure}
\begin{subfigure}{1\textwidth}
    \begin{center}
    \includegraphics[scale=0.27]{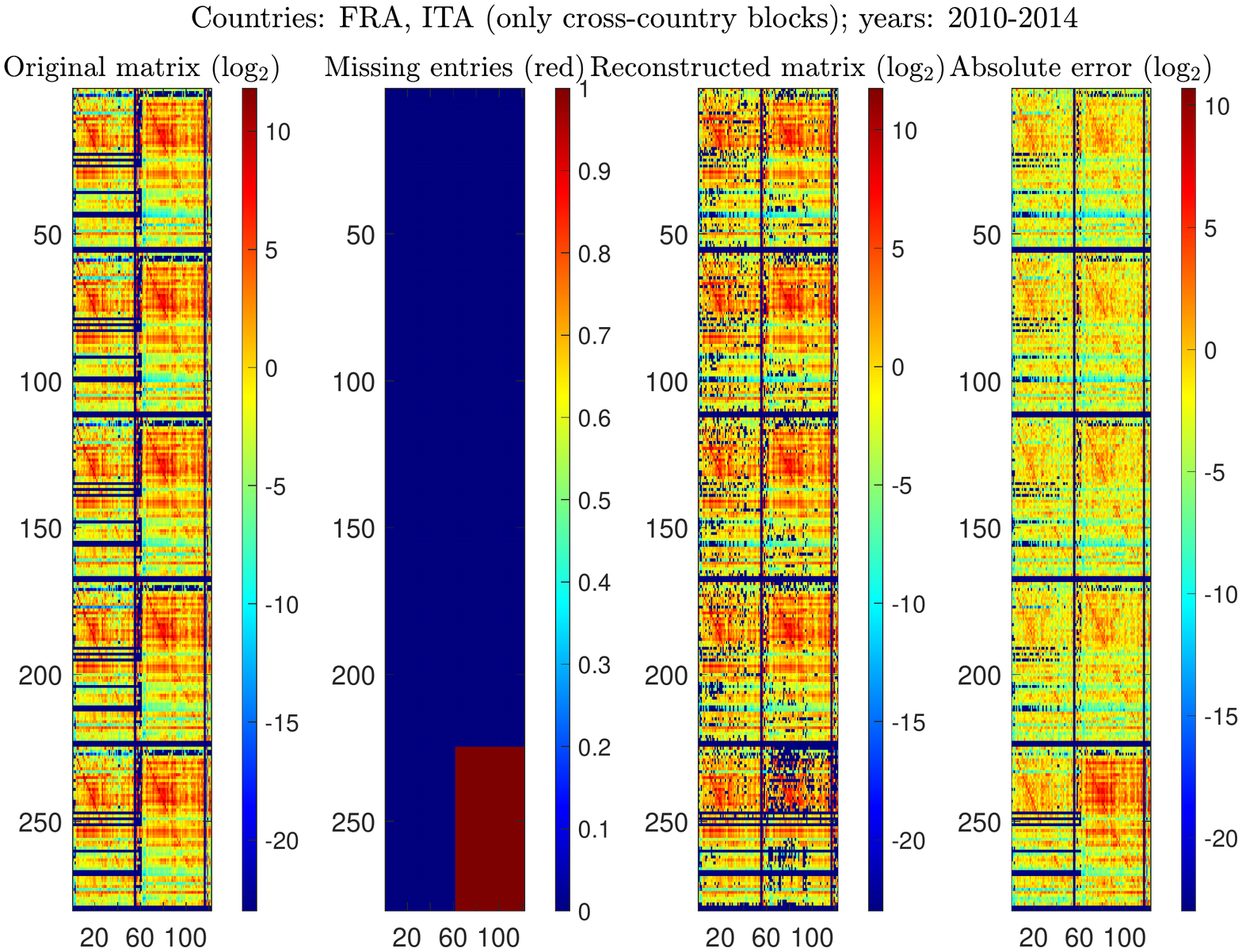}
    \vspace{-0.45cm}
    \caption{Colored visualization of the elements of the WIOD submatrix reported in Table \ref{tab:2}, positions of the missing entries, reconstructed submatrix obtained for the optimal regularization constant {\blue (RMSE criterion)}, and element-wise absolute value of the reconstruction error.}
    \label{fig:4}
    \end{center}
    \end{subfigure}
    \caption{}
\end{figure}

Figure \ref{fig:2} {\blue illustrates} the results of the application of the MC Algorithm \ref{alg:1} to the partially-observed WIOD submatrix reported in Table \ref{tab:2} {\blue  (details about the construction of the training, validation and test sets are provided at the end of Subsection \ref{sec:clustering_matrix_completion})}. 
The figure shows that, in this case, MC is able to reduce significantly the RMSE of reconstruction on the missing elements in 2015, when moving from the case $\lambda \simeq 0$ (for which the predictions of the missing elements are nearly equal to $0$) to the optimal choice of $\lambda$. More precisely, for $\lambda \simeq 0$, one gets
\begin{eqnarray}\nonumber
    RMSE_{\lambda}^{\rm val}&=&  \sqrt{\frac{1}{|\Omega^{\rm val}|}\sum_{(i,j) \in \Omega^{\rm val}} \left(M_{i,j}-{\blue  \hat{M}}_{\lambda,i,j} \right)^2} \nonumber \\
    &\simeq& \sqrt{\frac{1}{|\Omega^{\rm val}|}\sum_{(i,j) \in \Omega^{\rm val}} M_{i,j}^2} \simeq 86.5711
\end{eqnarray}
(since, in this case, one gets ${\bf S}_\lambda({\bf Y}) \simeq {\bf Y}$ from Equation (\ref{eq:operatorS}), hence every time Step 2.a of Algorithm \ref{alg:1} is performed, one gets a matrix ${\bf \blue \hat{M}}^{\rm new}$ whose entries are nearly equal to $0$ in the positions corresponding to unobserved entries of ${\bf M}$). Instead, for the optimal value {\blue  $\lambda^\circ=2^6=64$} of $\lambda$ (whose location is highlighted in the figure), one gets \begin{equation}\nonumber
RMSE_{\lambda^\circ}^{\rm val} \simeq {\blue  47.2791}\,,
\end{equation}
obtaining a reduction of the RMSE of about $45 \%$. A similar behavior is observed for the RMSE of matrix reconstruction on the test set, the reduction of such RMSE in this case being from $116.0874$ {\blue (for $\lambda \simeq 0$)} to {\blue $51.7544$} {\blue (for $\lambda = \lambda^\circ$)}, {\blue which amounts at} about $55 \%$. Hence, a good generalization capability is observed, showing no overfitting occurred in the application of MC\footnote{\blue In more detail, we argue that in this case there is no overfitting due to the two following reasons: the RMSE curves on the validation and test sets have a similar behavior, with approximately the same locations of the respective minimizers; the RMSEs on the three sets (training, validation and test sets) have quite similar orders of magnitude in correspondence of the optimal value of $\lambda$.}. {\blue Results in terms of SMAPE are reported in Figure \ref{fig:2var}. The irregular behavior of the curves associated with the SMAPE is due to the fact that MC does not address directly the SMAPE criterion, whereas the RMSE on the training set is part of the objective function of the MC optimization problem (\ref{eq:matrix_completion2}).} Moreover, Figure \ref{fig:3} compares the singular values distribution of the WIOD submatrix reported in Table \ref{tab:2}, and the one of the completed submatrix produced as output by the algorithm, for both $\lambda \simeq 0$ and the optimal value of $\lambda$. It is evident from the figure that MC was able to reconstruct excellently the singular values distribution of the original WIOD submatrix (part of which was not observed), {\blue due to the large overlap of the curves reported in the figure}. {\blue Moreover, such distribution converges rapidly to $0$, which, as already reported in the Introduction, is a necessary (but not sufficient) condition for a good performance of MC. Indeed,  Eckart-Young theorem (see Appendix \ref{appendix:1}) provides an upper bound on the performance of MC, for a given number of singular values kept.} Finally, Figure \ref{fig:4} shows a colored visualization of the elements of the original WIOD submatrix, the positions of the missing entries (highlighted in red), the reconstructed submatrix obtained for the optimal value of the regularization constant $\lambda$, and the element-wise absolute value of the reconstruction error. {\blue It is worth recalling that the positions of the missing entries form the union of the validation and test sets, whereas the position of the observed entries form the training set\footnote{\blue The explicit partition of the set of missing entries into validation and test sets is not reported in the figure, since such partition is randomly generated.}.} In this case, although the third column in Figure \ref{fig:4} shows that MC looks able to reconstruct some pattern in the missing block of the matrix (with respect to the case $\lambda \simeq 0$, for which the missing block is forecast by a block of all negligible elements), the reconstruction error looks to be still large (fourth column), having a similar pattern as the corresponding original non-obscured block (first column). This is partly due to the fact that a $50\%$ reconstruction error corresponds to a reduction by $1$ in logarithmic scale with base $2$. Improved results are reported in the next subsection (see the next Figures \ref{fig:2temp}
and \ref{fig:2temp_input}), where the choice of the WIOD subtable to which MC is applied is guided by {\blue hierarchical clustering}.

\vspace{-0.45cm}

\subsection{Matrix completion applied to groups of similar/dissimilar countries determined by hierarchical clustering}\label{sec:similardissimilar}

\begin{figure}
    \begin{center}
     \includegraphics[scale=0.61,trim=0.5cm 9.5cm 0 8.7cm]{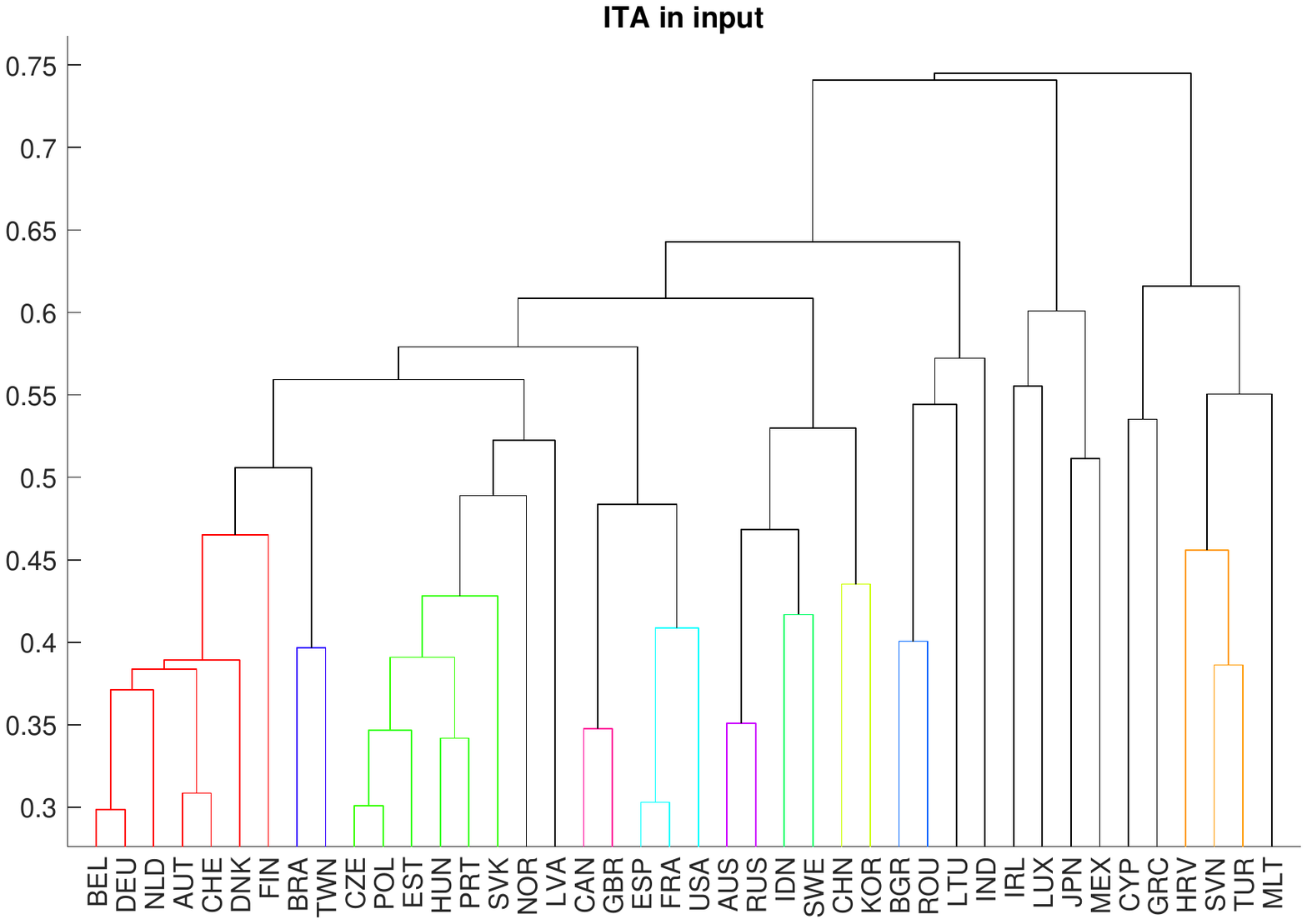}
    \end{center}
    \caption{\blue Dendrograms of output countries with Italian sectors in input, based on 
    WIOD tables (stacked over the years 2010--2013). Hierarchical clustering performed with the AACD dissimilarity measure ($y$-axis) and complete linkage. 21 desired groups (compare with Table \ref{tab:numclus_wss}).
    Countries in the same cluster are depicted with the same color. Countries in singleton clusters are highlighted in black.}\label{fig:dendrograms_in}
\end{figure}

\begin{figure}
    \begin{center}
  \includegraphics[scale=0.61,trim=0.5cm 9.5cm 0 8.7cm]{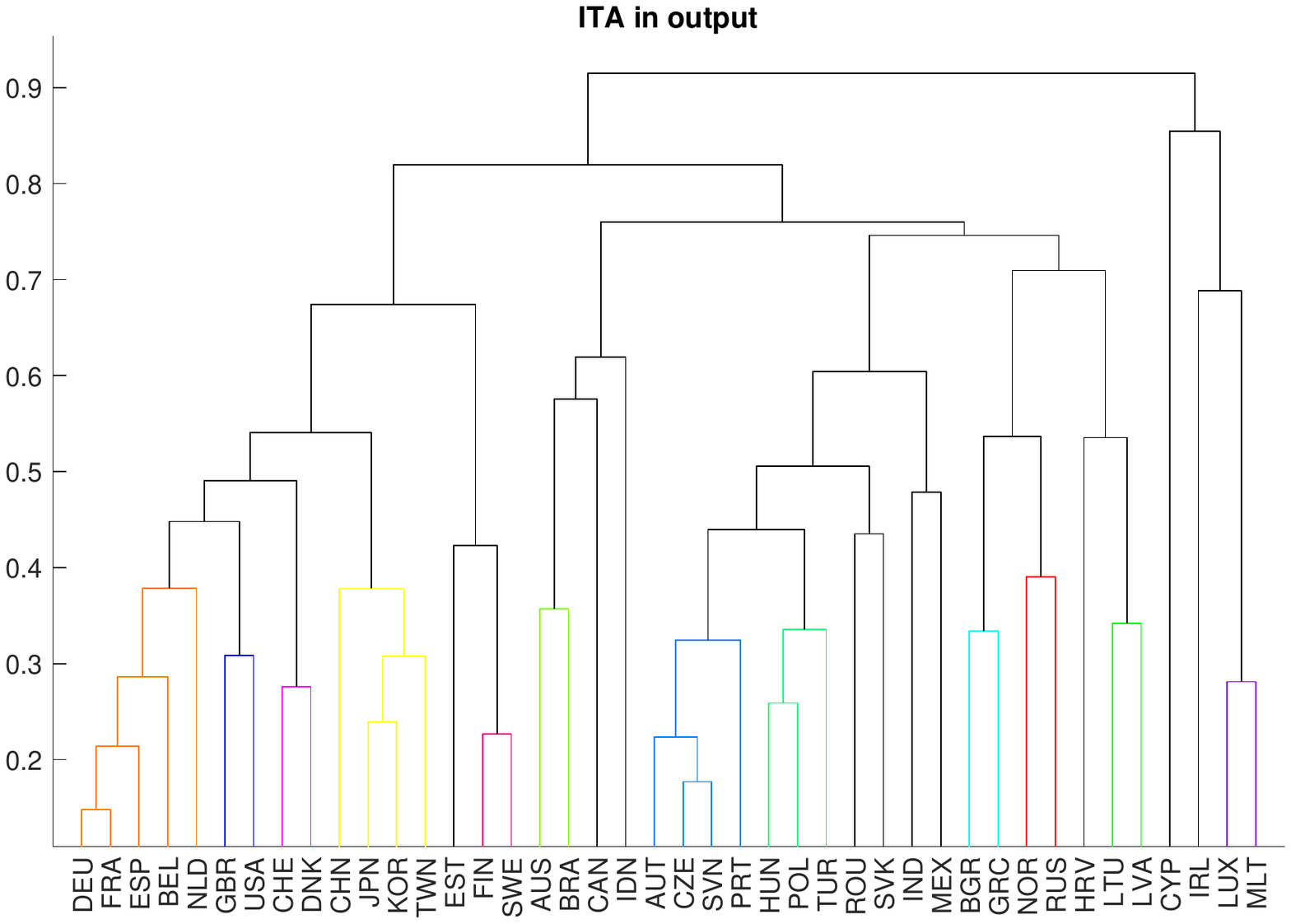}
    \end{center}
    \caption{\blue Dendrograms of input countries with Italian sectors in output, based on 
    WIOD tables (stacked over the years 2010--2013). Hierarchical clustering performed with the AACD dissimilarity measure ($y$-axis) and complete linkage. 22 desired groups (compare with Table \ref{tab:numclus_wss}).
    Countries in the same cluster are depicted with the same color. Countries in singleton clusters are highlighted in black. }\label{fig:dendrograms_out}
\end{figure}

In this subsection, {\blue using data from the WIOD latest release,} we compare the application of MC to WIOD submatrices obtained using a pre-processing step based on hierarchical clustering\footnote{\blue It is well-known from the standard practice of projection of I/O tables that it is typically better to use a table in previous years to project missing pieces of current I/O tables than using current I/O tables of similar countries \cite{Rueda-Cantucheetal2018,Valderas-Jaramilloetal2021}. One reason is that the former gather detailed country-specific information that is not expected to change in the short term. In this subsection, we use a combined approach, because we consider both the information coming from the same block in previous years, and the one coming from similar blocks in the current year.}. The dissimilarity between any two countries {\blue  $c_1$ and $c_2$} is computed as the AACD 
between the corresponding blocks of $\bold{T}$ in the WIOD table {\blue (stacked by considering several consecutive years)}, obtained by either choosing Italian sectors in input and intermediate/final outputs from the two countries {\blue $c_1$} and {\blue $c_2$} ({\blue  $\bold{T}^{Italy,c_1}$} and {\blue  $\bold{T}^{Italy,c_2}$}, recalling the notation introduced in Subsection \ref{sec:inoutmodel}), or choosing Italian intermediate/final outputs and sectors from the two countries {\blue $c_1$} and {\blue $c_2$} in input ({\blue  $\bold{T}^{c_1,Italy}$} and {\blue  $\bold{T}^{c_2,Italy}$}). {\blue In other words, the dissimilarity of the two countries {\blue  $c_1$} and {\blue $c_2$} in their Italian export patterns is evaluated in the first case, whereas their dissimilarity in the respective Italian import patterns is evaluated in the second case.} Both the hierarchical clustering analyses are repeated {\blue taking as inputs stacked I/O blocks associated with} several years (2010, 2011, 2012, and 2013), {\blue and using} complete linkage {\blue to perform} clustering. 
Figures \ref{fig:dendrograms_in} and \ref{fig:dendrograms_out} report the dendrograms  {\blue obtained}, 
where {\blue $c_1$} and {\blue $c_2$} are, respectively, {\blue both} output countries (Figure \ref{fig:dendrograms_in}), and {\blue  both} input countries (Figure \ref{fig:dendrograms_out}).

In this way, it is possible to extract from Figure \ref{fig:dendrograms_in} two groups of 4 output countries (see Tables \ref{tab:2temp} and \ref{tab:2tempbis}) that are, respectively, 
{\blue in the same cluster, 
and 
in 4 different clusters}. 

Table \ref{tab:2temp} refers to a WIOD submatrix whose blocks have Italian sectors in input and intermediate/final outputs associated the first group of extracted countries {\blue  (specifically, Austria, Belgium, Germany, and the Netherlands)}. In contrast, in Table \ref{tab:2tempbis}, the intermediate/final outputs refer to the second group of extracted countries {\blue (specifically, Australia, Belgium, Japan, and Malta)}. For predictive/matrix completion purposes, the tables contain also data related to the year 2014\footnote{\blue It is worth observing that, in Tables \ref{tab:2temp} and \ref{tab:2tempbis}, and in the successive Tables \ref{tab:2temp_input} and \ref{tab:2tempbis_input}, the blocks associated with a specific year are concatenated horizontally according to a given order, and such order does not change when considering the blocks associated with different {\blue years. The choice of a specific order is actually irrelevant for the application of MC, as it follows by combining the two following observations:
\begin{itemize}
\item the singular values of a rectangular matrix are invariant with respect to permutations of rows and/or columns of that matrix;
\item if the Soft Impute algorithm is applied to two different rectangular matrices ${\bf M}_1$ and ${\bf M}_2$ that differ only for permutations of rows and/or columns, and if the same permutations are applied to the matrix that represents the locations of the observed/unobserved elements (i.e., such matrix contains $1$ for every observed entry, and $0$ for every unobserved one), then the two matrices $\hat{\bf M}_1$ and $\hat{\bf M}_2$ obtained as output of MC in the two cases differ only for the same permutations.
\end{itemize}}}. Then, the MC Algorithm \ref{alg:1} is applied to both submatrices, after obscuring all the elements of their last block (highlighted in {\blue bold} in Tables \ref{tab:2temp} and \ref{tab:2tempbis}), which refers to a specific output country in 2014.

\begin{table}[h!]
\setlength{\tabcolsep}{3pt}
\begin{center}
\begin{tabular}{ | c | c | c | c | } 
\hline
\multicolumn{4}{ | c | }{\it I/O, year} \\
\hline
ITA/AUT, 2010 & ITA/BEL, 2010 & ITA/DEU, 2010 & ITA/NDL, 2010 \\
\hline
ITA/AUT, 2011 & ITA/BEL, 2011 & ITA/DEU, 2011 & ITA/NDL, 2011 \\
\hline
ITA/AUT, 2012 & ITA/BEL, 2012 & ITA/DEU, 2012 & ITA/NDL, 2012 \\
\hline
ITA/AUT, 2013 & ITA/BEL, 2013 & ITA/DEU, 2013 & ITA/NDL, 2013 \\
\hline
\bf{ITA/AUT, 2014} & ITA/BEL, 2014 & ITA/DEU, 2014 & ITA/NDL, 2014 \\
\hline
\end{tabular}
\end{center}
\caption{\blue Structure of the WIOD submatrix used for the example reported in Figure \ref{fig:2temp}. Similar comments as in Table \ref{tab:2} apply.}\label{tab:2temp}
\end{table}

\begin{table}[h!]
\setlength{\tabcolsep}{3pt}
\begin{center}
\begin{tabular}{ | c | c | c | c | } 
\hline
\multicolumn{4}{ | c | }{\it I/O, year} \\
\hline
ITA/AUS, 2010 & ITA/BEL, 2010 & ITA/JPN, 2010 & ITA/MLT, 2010 \\
\hline
ITA/AUS, 2011 & ITA/BEL, 2011 & ITA/JPN, 2011 & ITA/MLT, 2011 \\
\hline
ITA/AUS, 2012 & ITA/BEL, 2012 & ITA/JPN, 2012 & ITA/MLT, 2012 \\
\hline
ITA/AUS, 2013 & ITA/BEL, 2013 & ITA/JPN, 2013 & ITA/MLT, 2013 \\
\hline
ITA/AUS, 2014 & ITA/BEL, 2014 & \bf{ITA/JPN, 2014} & ITA/MLT, 2014 \\
\hline
\end{tabular}
\end{center}
\caption{\blue Structure of the WIOD submatrix used for the example reported in Figure \ref{fig:2tempbis}. Similar comments as in Table \ref{tab:2} apply.}\label{tab:2tempbis}
\end{table}

\begin{figure}
\vspace{-0.45cm}
\begin{subfigure}{1\textwidth}
\begin{center}
    \includegraphics[scale=0.27]{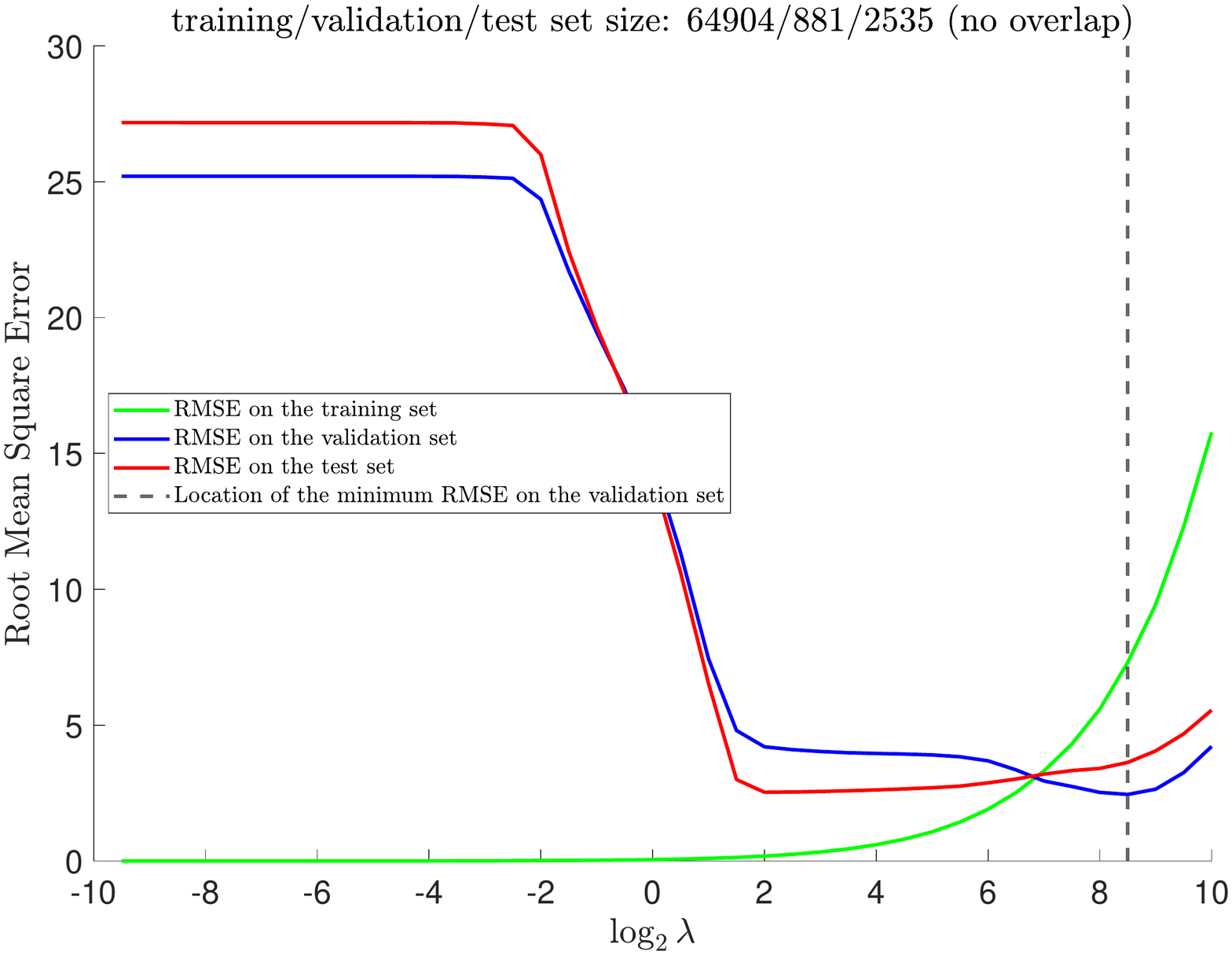}
    \vspace{-0.45cm}
    \caption{Results {\blue (expressed in terms of RMSE)} of the application of Algorithm \ref{alg:1} to the WIOD submatrix reported in Table \ref{tab:2temp}.}
    \label{fig:2temp}
    \end{center}
    \end{subfigure}
\begin{subfigure}{1\textwidth}
\begin{center}
    \includegraphics[scale=0.27]{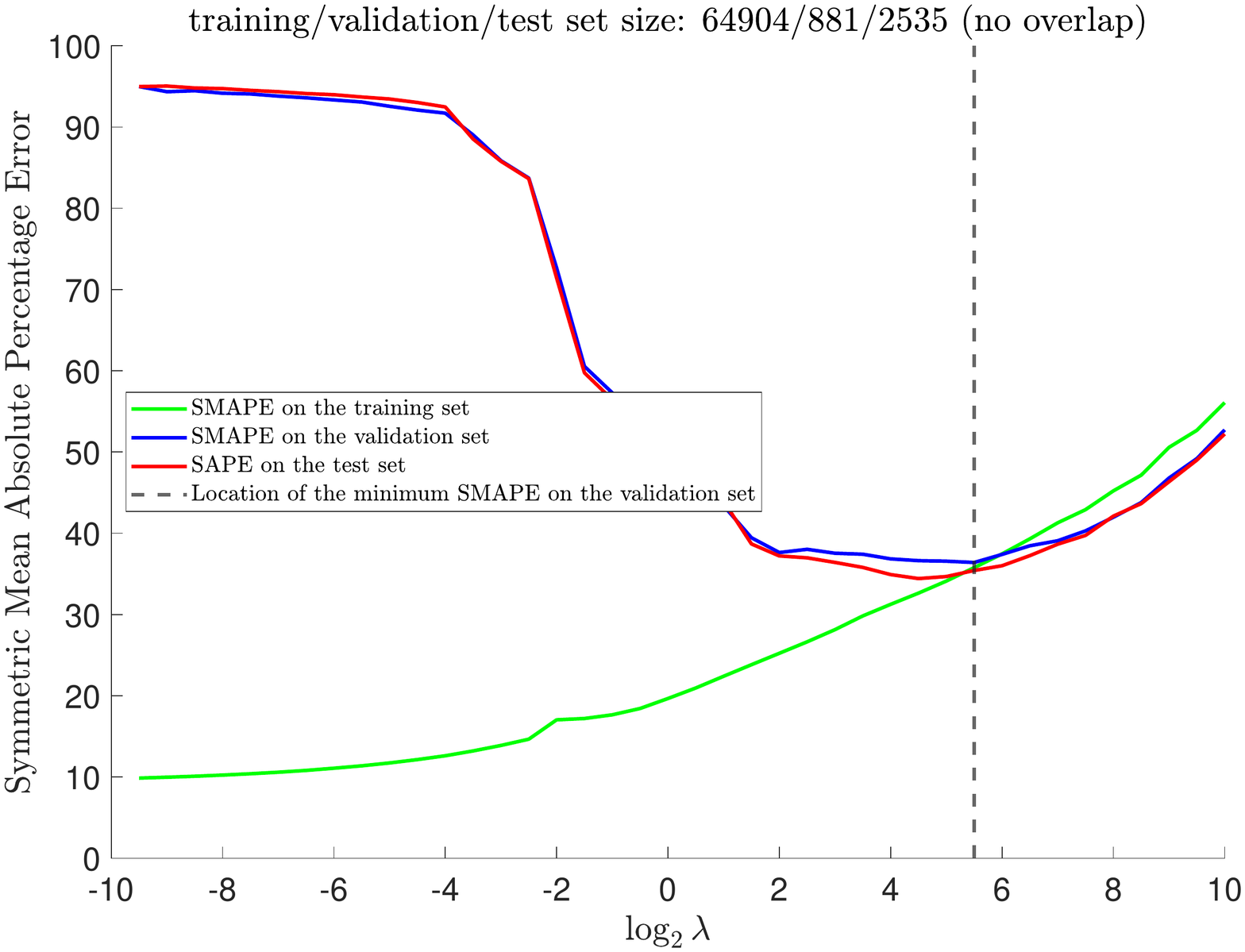}
    \vspace{-0.45cm}
    \caption{{\blue Results (expressed in terms of SMAPE) of the application of Algorithm \ref{alg:1} to the WIOD submatrix reported in Table \ref{tab:2temp}.}}
    \label{fig:2vartemp}
    \end{center}
    \end{subfigure}
\begin{subfigure}{1\textwidth}
\begin{center}
    \includegraphics[scale=0.27]{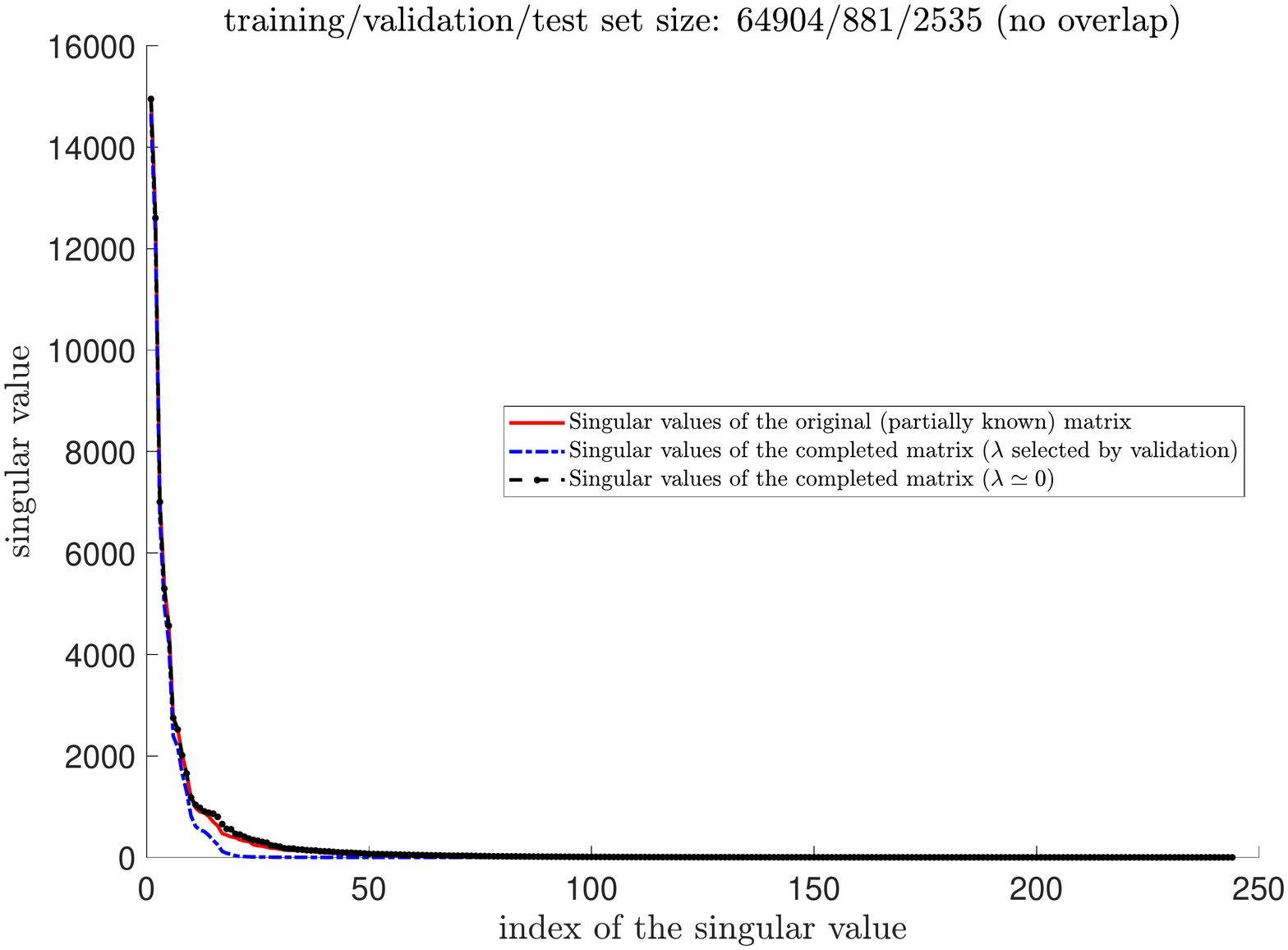}
    \vspace{-0.45cm}
    \caption{Singular values distribution of the WIOD submatrix reported in Table \ref{tab:2temp}, and the one of the completed submatrix produced 
    by Algorithm \ref{alg:1} for the optimal regularization constant {\blue (RMSE criterion)}.}
    \label{fig:3temp}
\end{center}
\end{subfigure}
\begin{subfigure}{1\textwidth}
\begin{center}
    \includegraphics[scale=0.27]{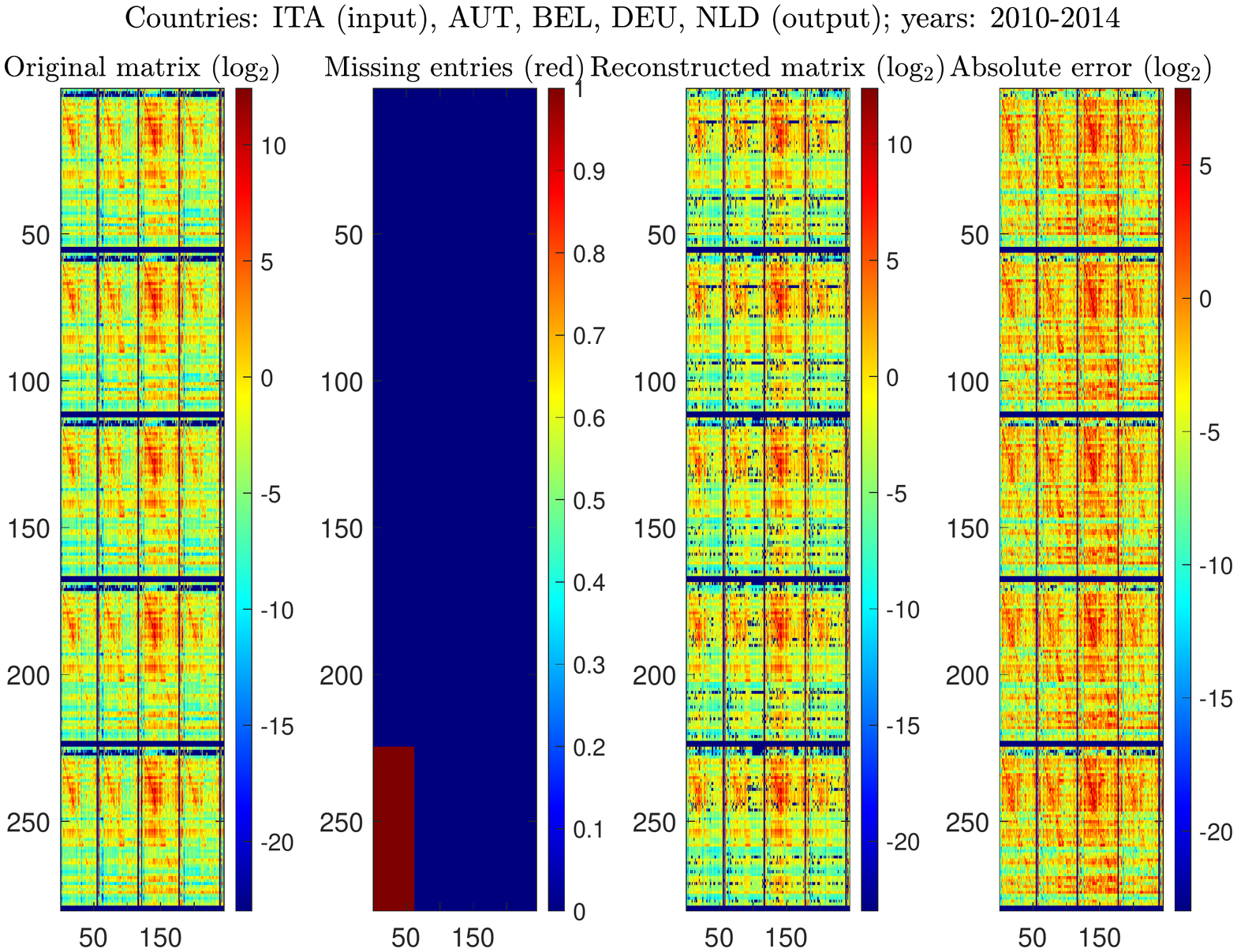}
    \vspace{-0.45cm}
    \caption{Colored visualization of the elements of the WIOD submatrix reported in Table \ref{tab:2temp}, positions of the missing entries, reconstructed submatrix obtained for the optimal regularization constant {\blue (RMSE criterion)}, and element-wise absolute value of the reconstruction error.}
    \label{fig:4temp}
    \end{center}
    \end{subfigure}
    \caption{}
\end{figure}

\begin{figure}
\vspace{-0.45cm}
\begin{subfigure}{1\textwidth}
\begin{center}
    \includegraphics[scale=0.27]{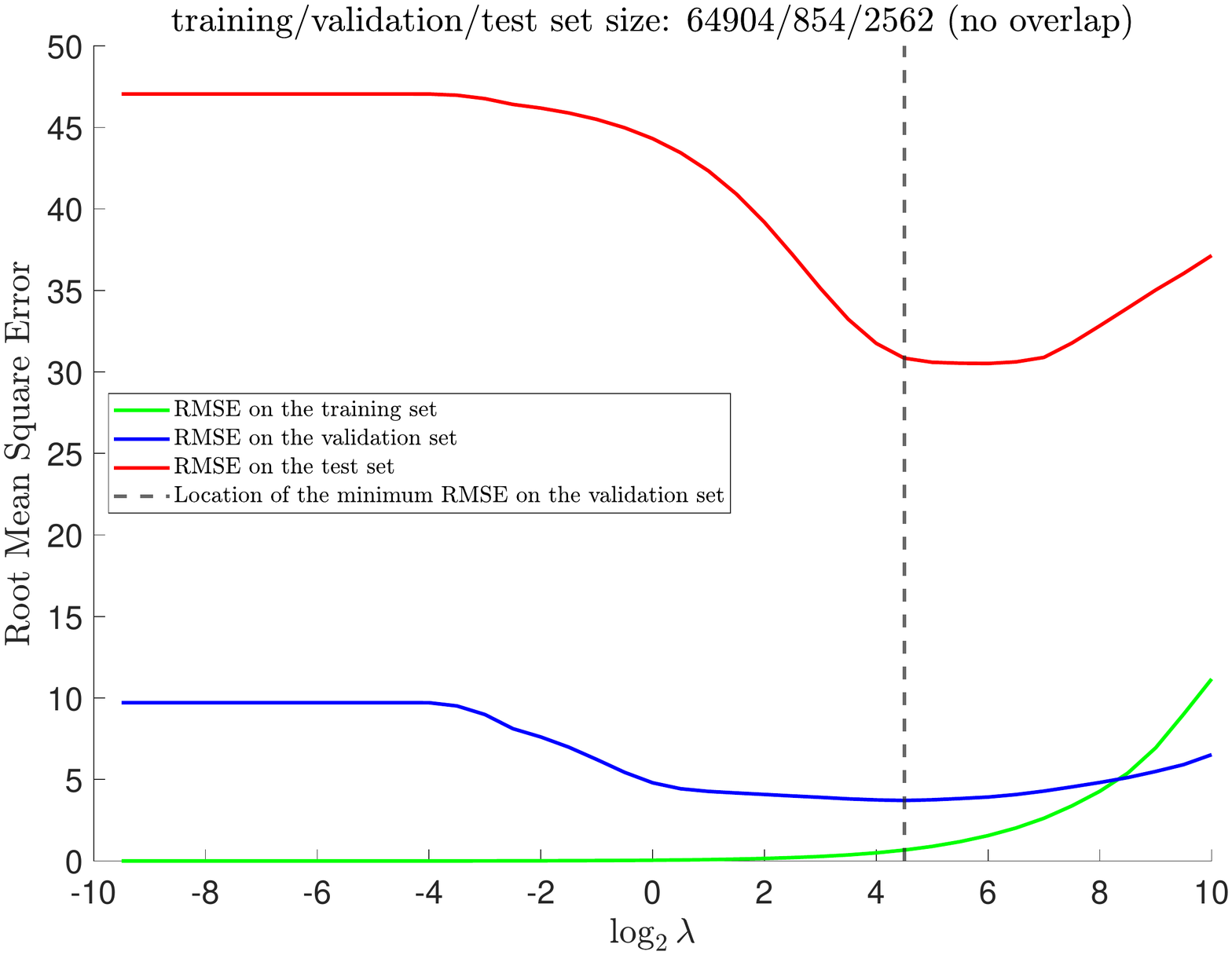}
    \vspace{-0.45cm}
    \caption{Results {\blue (expressed in terms of RMSE)} of the application of Algorithm \ref{alg:1} to the WIOD submatrix reported in Table \ref{tab:2tempbis}.}
    \label{fig:2tempbis}
    \end{center}
    \end{subfigure}
\begin{subfigure}{1\textwidth}
\begin{center}
    \includegraphics[scale=0.27]{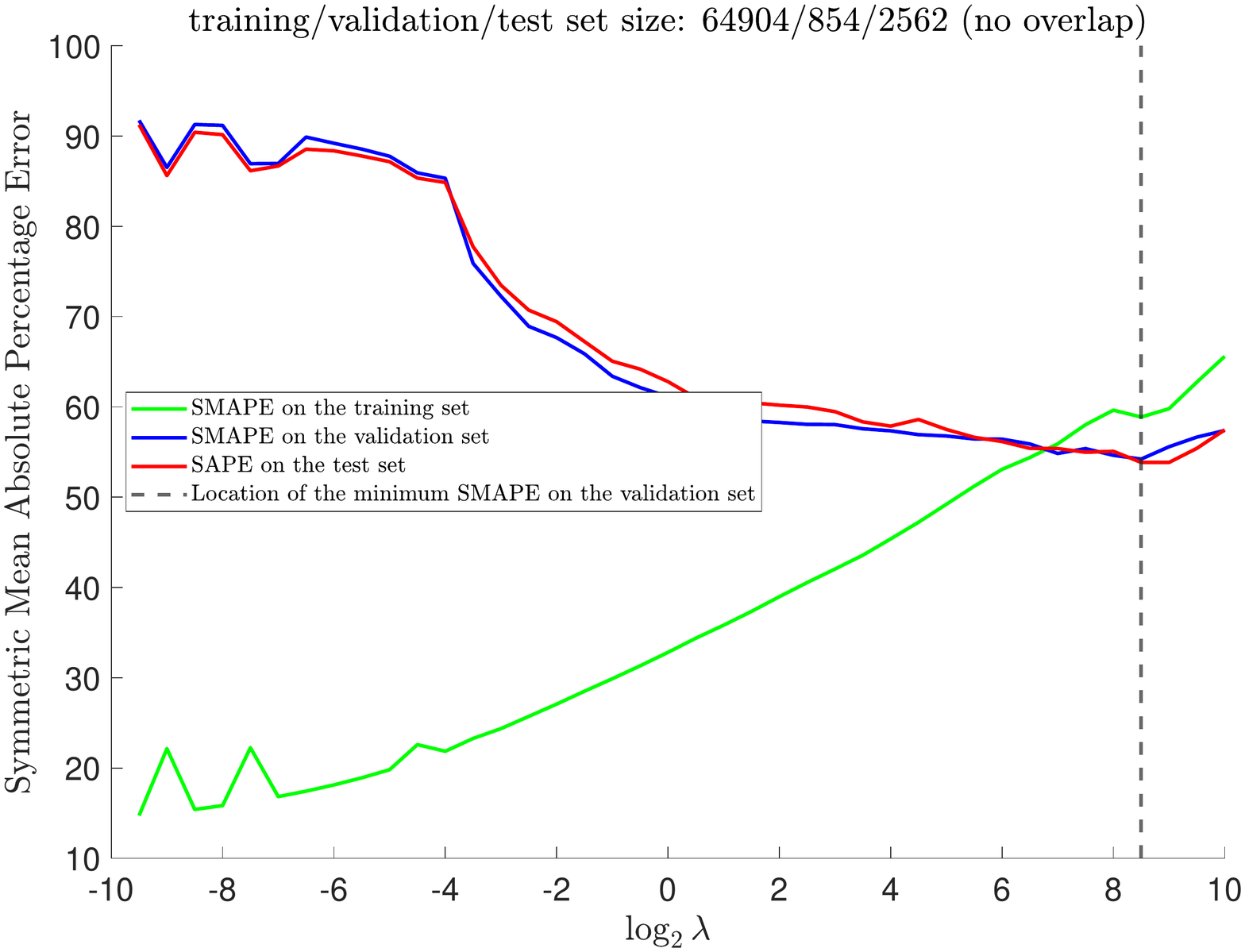}
    \vspace{-0.45cm}
    \caption{{\blue Results (expressed in terms of SMAPE) of the application of Algorithm \ref{alg:1} to the WIOD submatrix reported in Table \ref{tab:2tempbis}.}}
    \label{fig:2vartempbis}
    \end{center}
    \end{subfigure}
\begin{subfigure}{1\textwidth}
\begin{center}
    \includegraphics[scale=0.27]{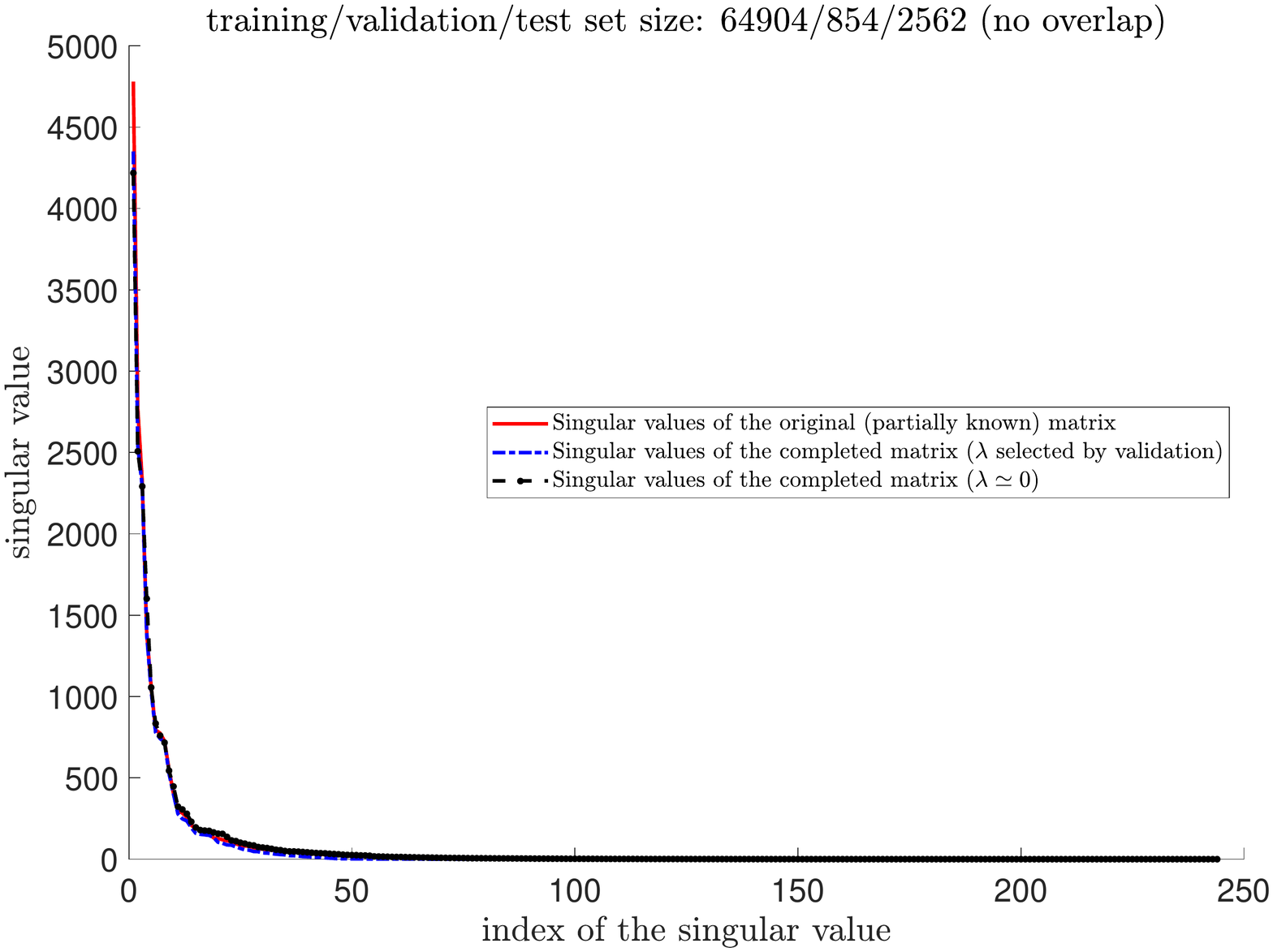}
    \vspace{-0.45cm}
    \caption{Singular values distribution of the WIOD submatrix reported in Table \ref{tab:2tempbis}, and the one of the completed submatrix produced 
    by Algorithm \ref{alg:1} for the optimal regularization constant {\blue (RMSE criterion)}.}
    \label{fig:3tempbis}
\end{center}
\end{subfigure}
\begin{subfigure}{1\textwidth}
\begin{center}
    \includegraphics[scale=0.27]{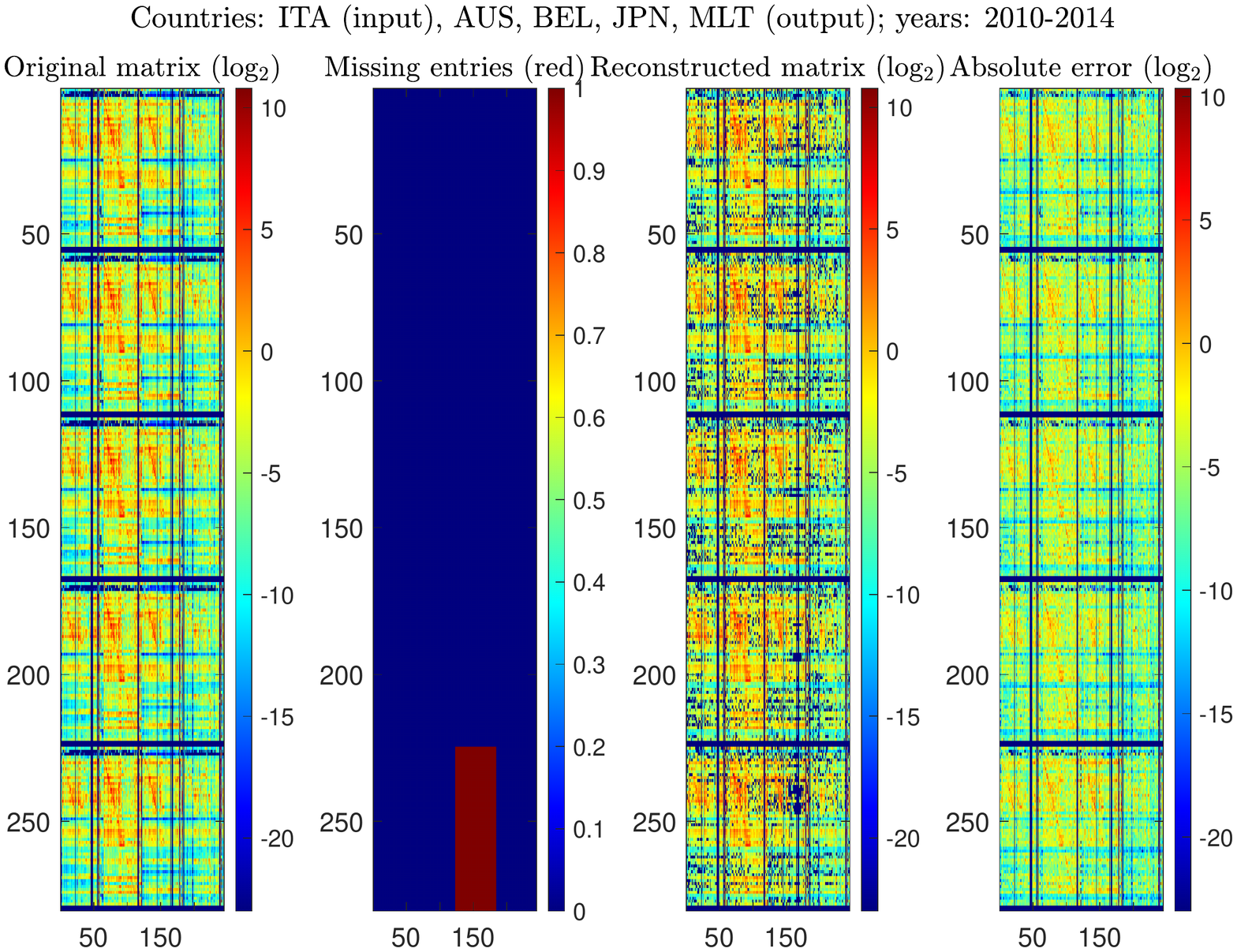}
    \vspace{-0.45cm}
    \caption{Colored visualization of the elements of the WIOD submatrix reported in Table \ref{tab:2temp}, positions of the missing entries, reconstructed submatrix obtained for the optimal regularization constant {\blue (RMSE criterion)}, and element-wise absolute value of the reconstruction error.}
    \label{fig:4tempbis}
    \end{center}
    \end{subfigure}
    \caption{}
\end{figure}

Figures \ref{fig:2temp} and \ref{fig:2tempbis} report the results of the application of the MC Algorithm \ref{alg:1} to the two WIOD submatrices whose structures are described, respectively, in Tables \ref{tab:2temp} and \ref{tab:2tempbis}. 
As expected, the results show a better performance of the MC algorithm, measured in terms of the percentage of reduction of the RMSE on the validation set from $\lambda \simeq 0$ to the optimal choice of $\lambda$, in the case of the first submatrix, whose intermediate/final outputs are associated with more similar countries. 

\begin{figure}
\vspace{-0.45cm}
\begin{subfigure}{1\textwidth}
\begin{center}
    \includegraphics[scale=0.27]{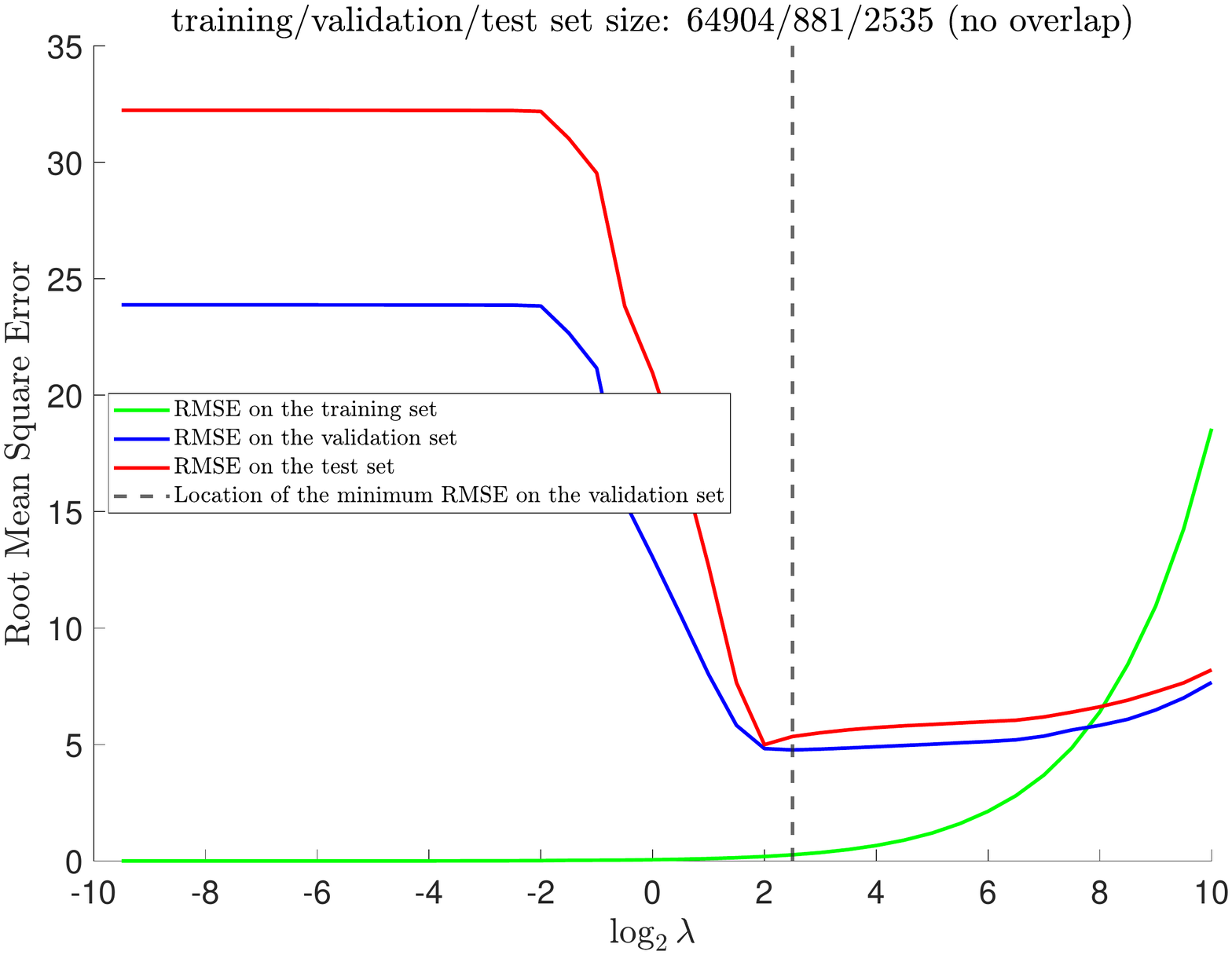}
    \vspace{-0.45cm}
    \caption{Results {\blue (expressed in terms of RMSE)} of the application of Algorithm \ref{alg:1} to the WIOD submatrix reported in Table \ref{tab:2temp_input}.}
    \label{fig:2temp_input}
    \end{center}
    \end{subfigure}
\begin{subfigure}{1\textwidth}
\begin{center}
    \includegraphics[scale=0.27]{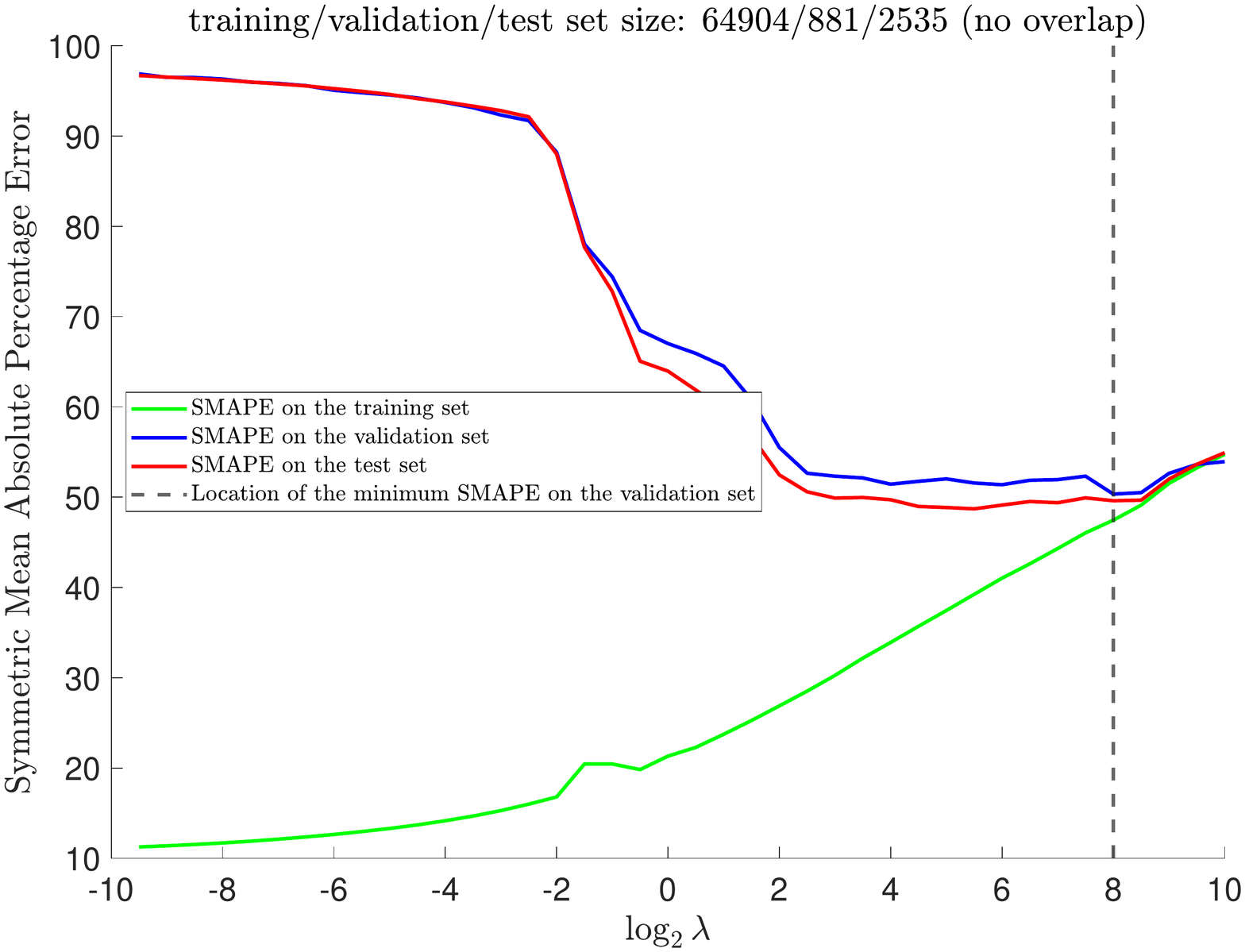}
    \vspace{-0.45cm}
    \caption{{\blue Results (expressed in terms of SMAPE) of the application of Algorithm \ref{alg:1} to the WIOD submatrix reported in Table \ref{tab:2temp_input}.}}
    \label{fig:2vartemp_input}
    \end{center}
    \end{subfigure}
\begin{subfigure}{1\textwidth}
\begin{center}
    \includegraphics[scale=0.27]{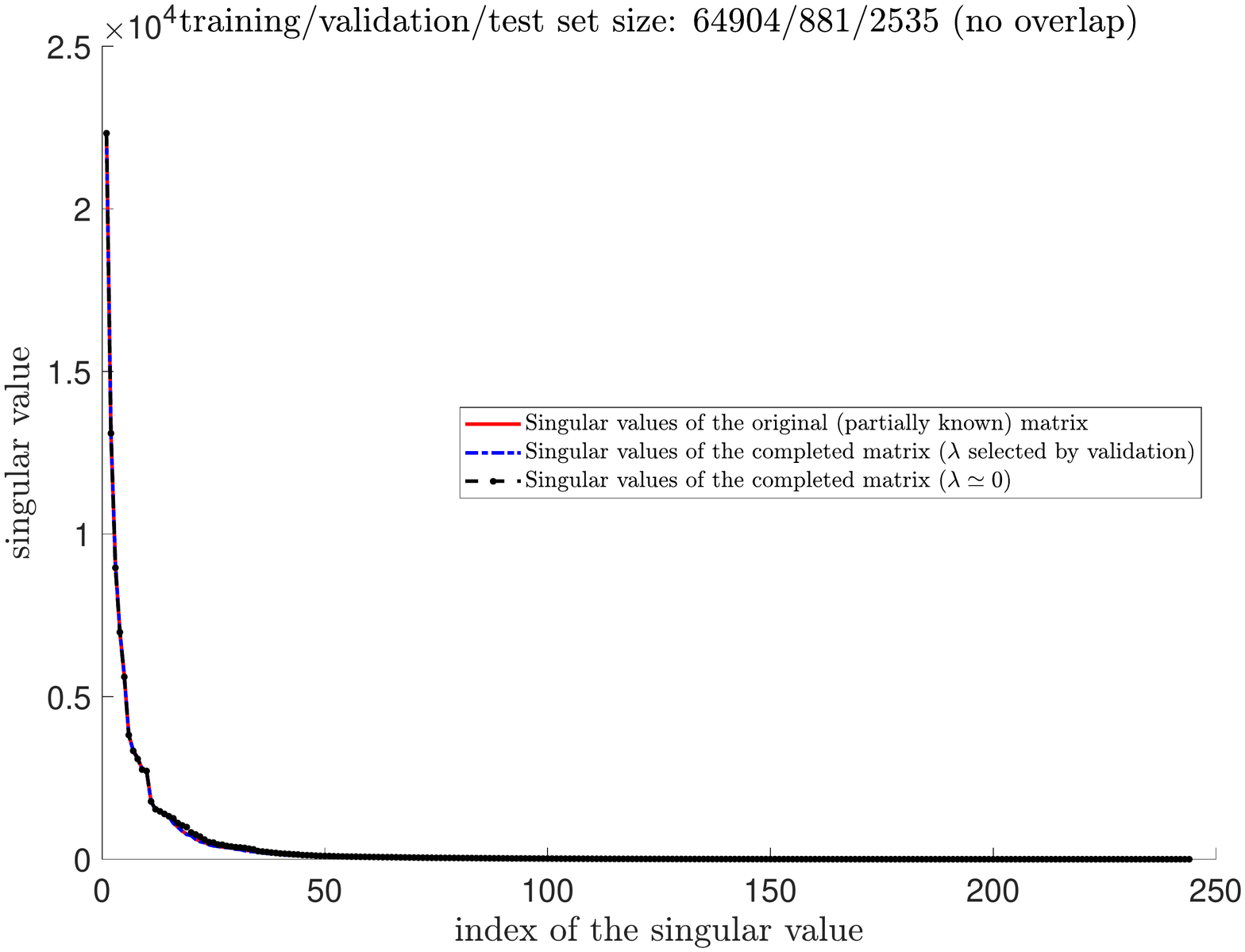}
    \vspace{-0.45cm}
    \caption{Singular values distribution of the WIOD submatrix reported in Table \ref{tab:2temp_input}, and the one of the completed submatrix produced 
    by Algorithm \ref{alg:1} for the optimal regularization constant {\blue (RMSE criterion)}.}
    \label{fig:3temp_input}
\end{center}
\end{subfigure}
\begin{subfigure}{1\textwidth}
\begin{center}
    \includegraphics[scale=0.27]{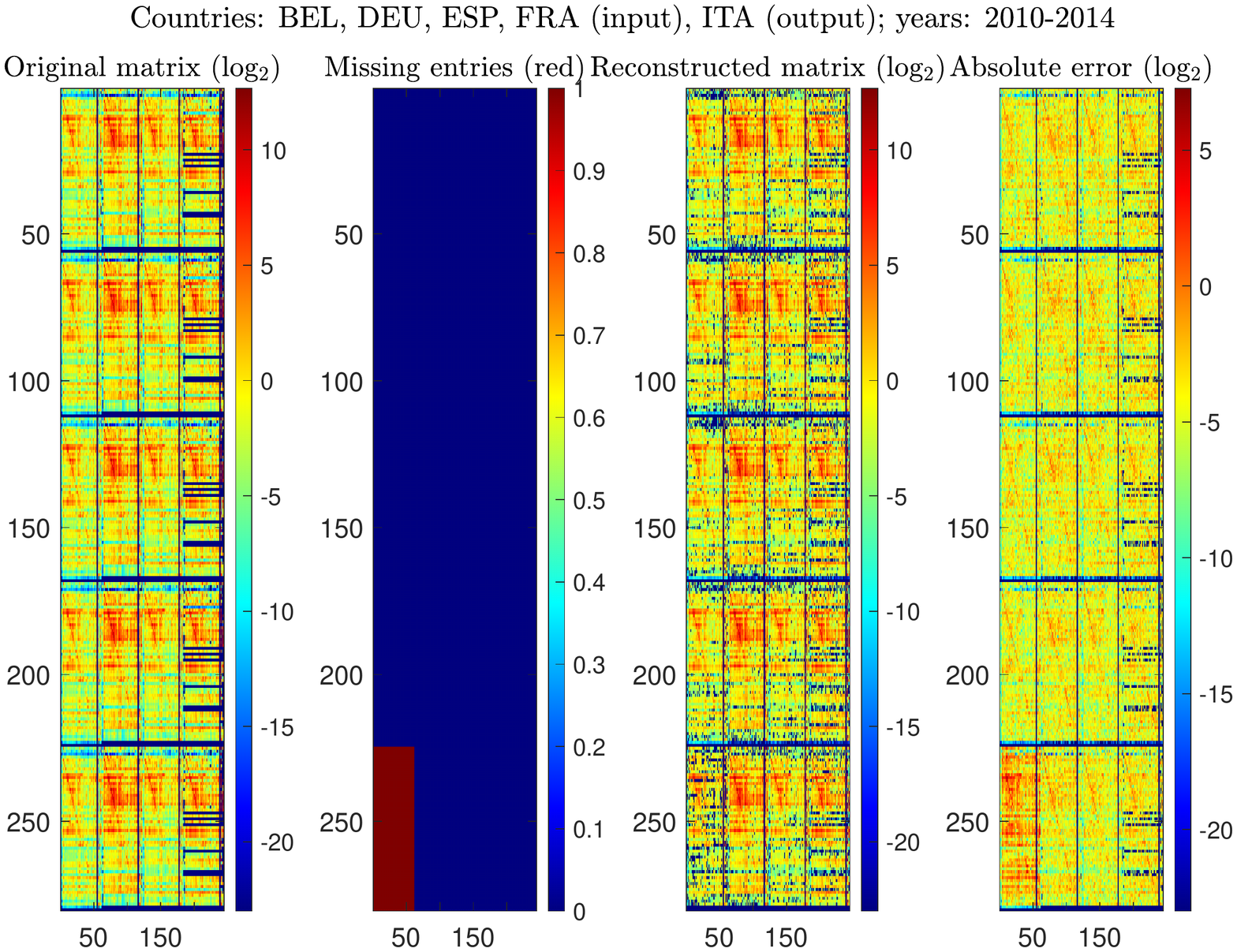}
    \vspace{-0.45cm}
    \caption{Colored visualization of the elements of the WIOD submatrix reported in Table \ref{tab:2temp_input}, positions of the missing entries, reconstructed submatrix obtained for the optimal regularization constant {\blue (RMSE criterion)}, and element-wise absolute value of the reconstruction error.}
    \label{fig:4temp_input}
    \end{center}
    \end{subfigure}
    \caption{}
\end{figure}

\begin{figure}
\vspace{-0.45cm}
\begin{subfigure}{1\textwidth}
\begin{center}
    \includegraphics[scale=0.27]{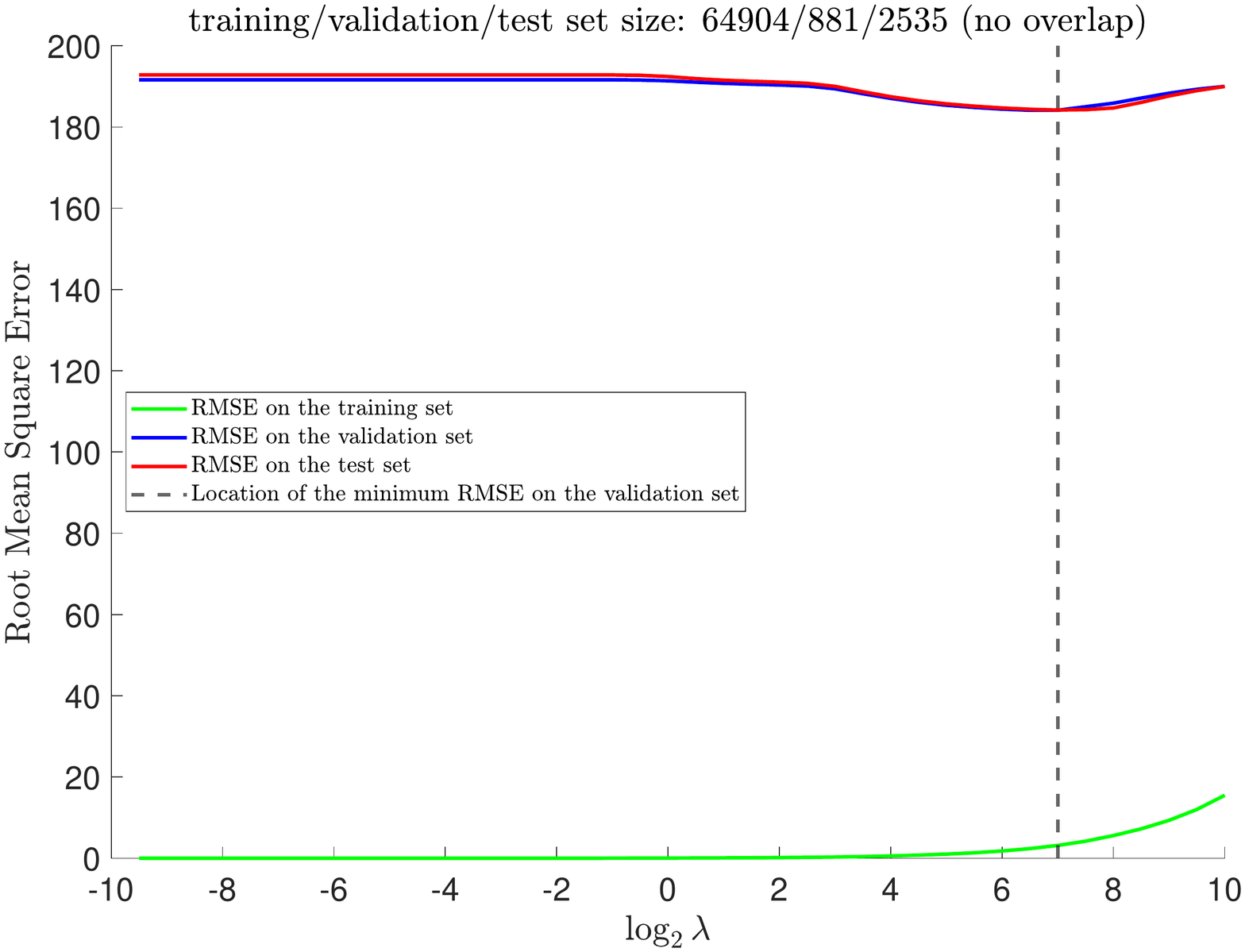}
    \vspace{-0.45cm}
    \caption{Results {\blue (expressed in terms of RMSE)} of the application of Algorithm \ref{alg:1} to the WIOD submatrix reported in Table \ref{tab:2tempbis_input}.}
    \label{fig:2tempbis_input}
    \end{center}
    \end{subfigure}
\begin{subfigure}{1\textwidth}
\begin{center}
    \includegraphics[scale=0.27]{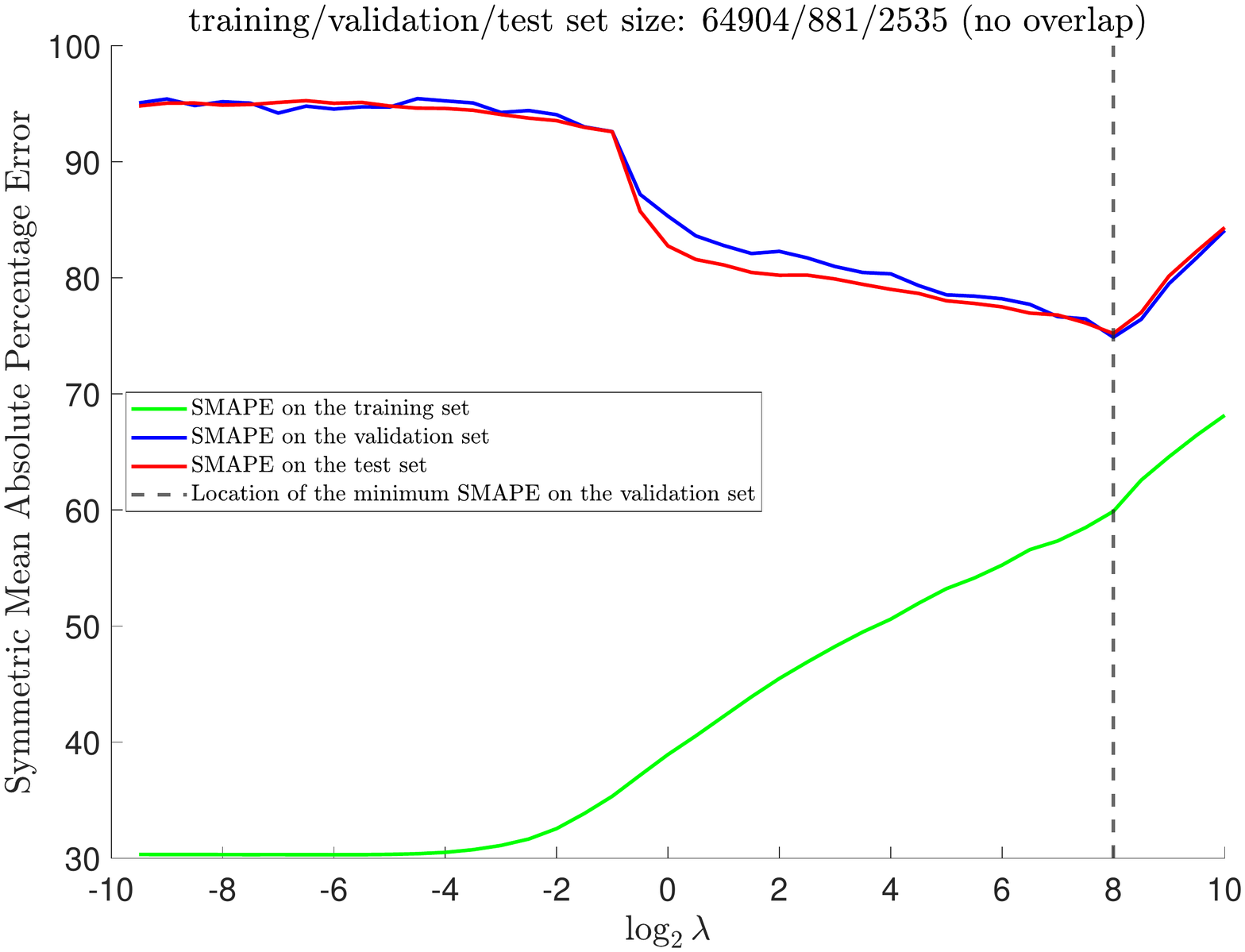}
    \vspace{-0.45cm}
    \caption{{\blue Results (expressed in terms of SMAPE) of the application of Algorithm \ref{alg:1} to the WIOD submatrix reported in Table \ref{tab:2tempbis_input}.}}
    \label{fig:2vartempbis_input}
    \end{center}
    \end{subfigure}
\begin{subfigure}{1\textwidth}
\begin{center}
    \includegraphics[scale=0.27]{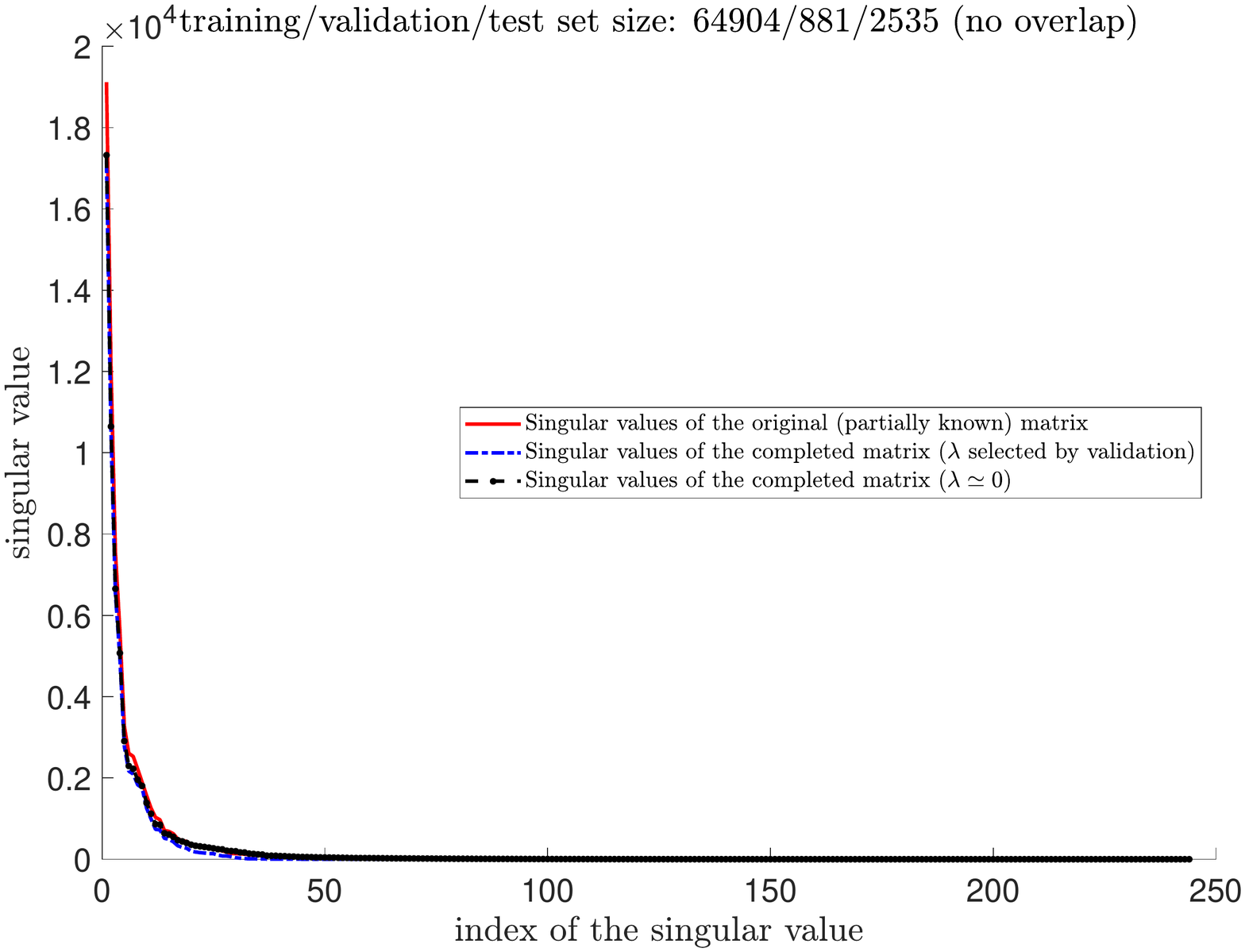}
    \vspace{-0.45cm}
    \caption{Singular values distribution of the WIOD submatrix reported in Table \ref{tab:2tempbis_input}, and the one of the completed submatrix produced 
    by Algorithm \ref{alg:1} for the optimal regularization constant {\blue (RMSE criterion)}.}
    \label{fig:3tempbis_input}
\end{center}
\end{subfigure}
\begin{subfigure}{1\textwidth}
\begin{center}
    \includegraphics[scale=0.27]{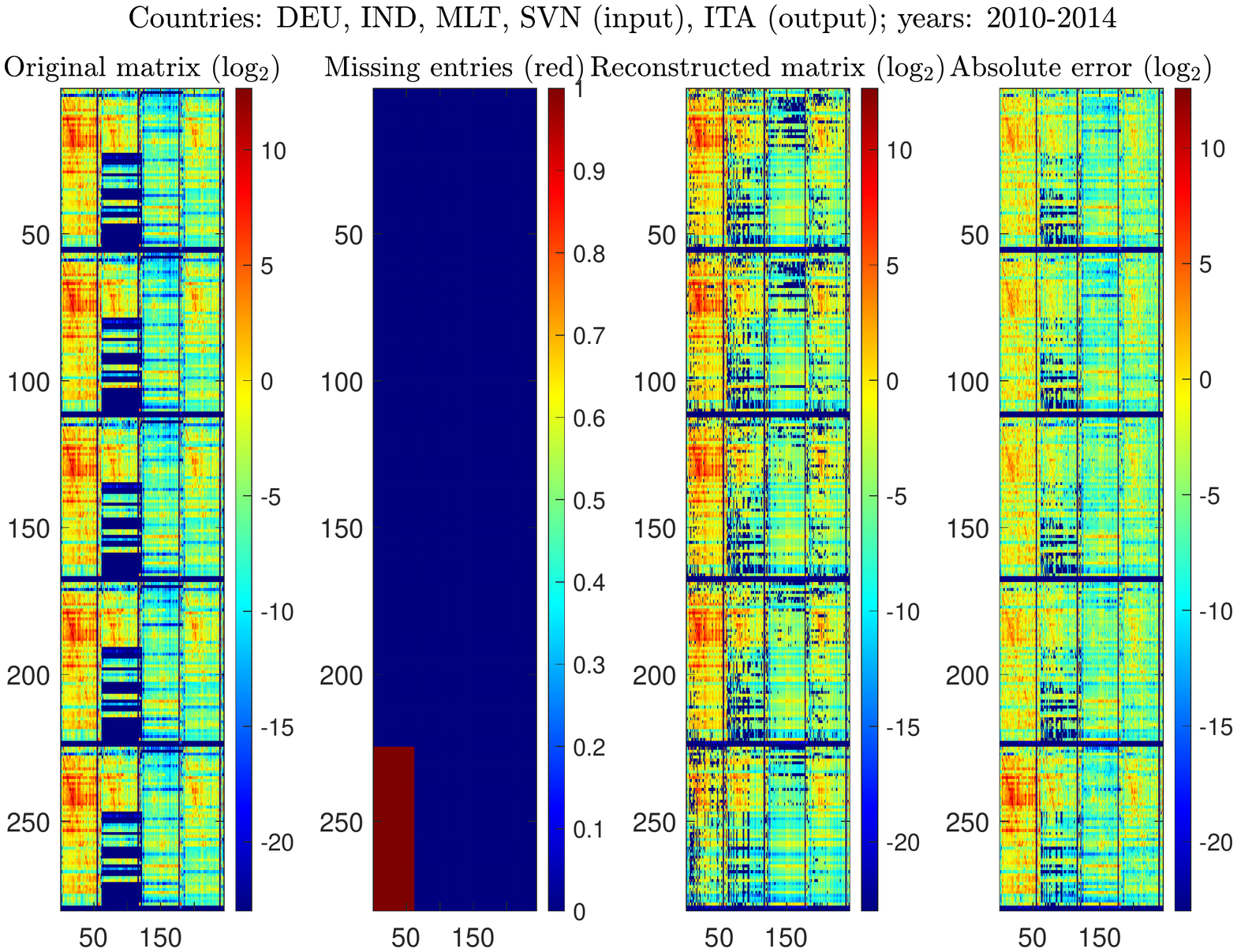}
    \vspace{-0.45cm}
    \caption{Colored visualization of the elements of the WIOD submatrix reported in Table \ref{tab:2tempbis_input}, positions of the missing entries, reconstructed submatrix obtained for the optimal regularization constant {\blue (RMSE criterion)}, and element-wise absolute value of the reconstruction error.}
    \label{fig:4tempbis_input}
    \end{center}
    \end{subfigure}
    \caption{}
\end{figure}

{\blue It} is worth observing that quite similar results have been obtained if a different 2014 block corresponding to another country in the group of 4 countries is obscured in each of the two WIOD submatrices, or when the analysis has been repeated by considering Italy in output and 4 similar/dissimilar countries in input (see the {\blue dendrogram} reported in Figure \ref{fig:dendrograms_out}). 
In this last analysis, the {\blue selected} subset of 4 similar countries {\blue in input} is made by {\blue Belgium, Germany, Spain, and France} (see Table \ref{tab:2temp_input}), whereas the {\blue selected} subset of 4 dissimilar countries {\blue in input} is made by {\blue Germany, India, Malta, and Slovenia} (see Table \ref{tab:2tempbis_input}). Corresponding results of the MC analysis 
{\blue are} reported in Figures \ref{fig:2temp_input} and \ref{fig:2tempbis_input}. Again, similar comments as before apply: when more similar input countries are considered {\blue and the RMSE criterion is considered}, the performance of MC improves.

{\blue Finally, a comparison of Figures \ref{fig:2vartemp}, \ref{fig:2vartempbis}, \ref{fig:2vartemp_input}, and \ref{fig:2vartempbis_input} shows that, also when the SMAPE performance measure is used, MC applied to similar countries produces better results (in terms of relative improvement with the respect to the baseline case) than MC applied to dissimilar countries.}

\begin{table}[h!]
\setlength{\tabcolsep}{3pt}
\begin{center}
\begin{tabular}{ | c | c | c | c | } 
\hline
\multicolumn{4}{ | c | }{\it I/O, year} \\
\hline
BEL/ITA, 2010 & DEU/ITA, 2010 & ESP/ITA, 2010 & FRA/ITA, 2010 \\
\hline
BEL/ITA, 2011 & DEU/ITA, 2011 & ESP/ITA, 2011 & FRA/ITA, 2011 \\
\hline
BEL/ITA, 2012 & DEU/ITA, 2012 & ESP/ITA, 2012 & FRA/ITA, 2012 \\
\hline
BEL/ITA, 2013 & DEU/ITA, 2013 & ESP/ITA, 2013 & FRA/ITA, 2013 \\
\hline
\bf{BEL/ITA, 2014} & DEU/ITA, 2014 & ESP/ITA, 2014 &  FRA/ITA, 2014 \\
\hline
\end{tabular}
\end{center}
\caption{\blue Structure of the WIOD submatrix used for the example reported in Figure \ref{fig:2temp_input}. Similar comments as in Table \ref{tab:2} apply.}\label{tab:2temp_input}
\end{table}

\begin{table}[h!]
\setlength{\tabcolsep}{3pt}
\begin{center}
\begin{tabular}{ | c | c | c | c | } 
\hline
\multicolumn{4}{ | c | }{\it I/O, year} \\
\hline
DEU/ITA, 2010 & IND/ITA, 2010 & MLT/ITA, 2010 & SVN/ITA, 2010 \\
\hline
DEU/ITA, 2011 & IND/ITA, 2011 & MLT/ITA, 2011 & SVN/ITA, 2011 \\
\hline
DEU/ITA, 2012 & IND/ITA, 2012 & MLT/ITA, 2012 & SVN/ITA, 2012 \\
\hline
DEU/ITA, 2013 & IND/ITA, 2013 & MLT/ITA, 2013 & SVN/ITA, 2013 \\
\hline
\bf{DEU/ITA, 2014} & IND/ITA, 2014 & MLT/ITA, 2014 &  SVN/ITA, 2014 \\
\hline
\end{tabular}
\end{center}
\caption{\blue Structure of the WIOD submatrix used for the example reported in Figure \ref{fig:2tempbis_input}. Similar comments as in Table \ref{tab:2} apply.}\label{tab:2tempbis_input}
\end{table}

\section{Future research and concluding remarks} \label{sec:conclusion}


This work represents the first attempt to adopt a Matrix Completion (MC) algorithm, {\blue combined with hierarchical clustering pre-preprocessing step}, to forecast missing entries in {\blue  submatrices of} I/O tables in the context of a panel data analysis.

The particular structure of I/O tables, reported in the article, makes the data reconstruction and forecasting problems not trivial. {\blue Hence}, in the pre-processing phase, we have employed the dissimilarity pattern of countries to define low-rank I/O subtables with few dominant singular values. A panel matrix completion with nuclear norm penalty has been tested on those low-rank subtables. The effectiveness of the proposed method according to historical data available from previous years has been demonstrated when the considered I/O subtables are obtained by selecting similar countries.


A first possible extension of the analysis concerns comparing matrix reconstruction of I/O tables {\blue (in one year, based on current and previous years)} based on the repeated application of matrix completion to several subtables of the original I/O table, instead of a single more computationally expensive and (presumably) less effective application to the whole table {\blue (possibly after removing domestic blocks, likewise in this article)}. 
{\blue The proposed methodology is expected to be applicable, with similar results, also to other I/O tables (either industry-by-industry and product-by-product ones), because their structure is often similar to the one of WIOD tables, as highlighted in this work\footnote{\blue In this regard, it is worth mentioning that WIOD is currently available only till 2014, and it is unlikely that there will be important updates of it in the future, whereas OECD and FIGARO I/O tables are expected to be the main official sources for the future.}.}

{\blue As a second possible extension, algorithms for clustering and matrix completion different from those employed in the present article could be used.} Moreover, matrix completion itself could be compared with other imputation methods for missing entries in panel data models. {\blue A comparison with alternative methods such as the one suggested in \cite{Rueda-Cantucheetal2018} is left for future research.}

As another possible extension, the approach adopted in the paper could be applied to generate counterfactuals of I/O submatrices: e.g., by predicting how the entries of a suitably-specified input-output submatrix related to Japan would have changed, in case the March 2011 earthquake and tsunami and the successive Fukushima Daiichi nuclear disaster \cite{Yonemoto2016} would have not  occurred. 
To do this, one would preliminary need to identify sectors of the economy that were not affected by such events, then obscure (and reconstruct) the entries of that submatrix related to other sectors that, on the contrary, were affected. 

\section*{Declarations}
\section*{Funding}
No external funding
\section*{Conflicts of interest/Competing interests (include appropriate disclosures)}
 No conflicts of interest/competing interest
\section*{Availability of data and material (data transparency)}
Data are freely available online at \url{www.wiod.org}
\section*{Code availability (software application or custom code)}
Codes are available upon request
\section*{Authors' contributions}

(CRediT author statement)

\textbf{RM}: Methodology, Software, Formal Analysis, Investigation, Data Curation, Writing - Original Draft; \textbf{GG}: Validation, Methodology, Software, Formal Analysis, Investigation, Data Curation, Writing - Original Draft; \textbf{FB}: Formal Analysis, Investigation, Writing - Original Draft; \textbf{MR}: Conceptualization, Formal Analysis, Methodology, Resources, Writing - Review \& Editing, Supervision, Project Administration. {\blue All the authors analysed the results and reviewed the manuscript.}

\appendix\section{Appendix}

\setcounter{equation}{0}
\renewcommand{\theequation}{A\arabic{equation}}

\subsection{\blue Relationship between  performance of matrix completion and singular value decomposition of the matrix to be completed}\label{appendix:1}

If one indicates by $k$ any non-negative integer smaller than the rank of a rectangular matrix ${\bf M} \in \mathbb{R}^{m \times n}$ {\blue  (denoted as ${\rm rank}({\bf M})$)}, then, by Eckart-Young theorem (see, e.g., \cite[Theorem 2.1.2]{Moitra2018}), the best rank-$k$ approximation ${\bf M}_k$ of ${\bf M}$ according to the Euclidean norm is provided by the truncated singular value decomposition of ${\bf M}$, in which one keeps only its $k$-largest singular values, and zeroes all the others. Its Root Mean Square Error (RMSE) of reconstruction is equal to \begin{equation}\label{eq:Eckart-Young}
    RMSE_k:=\frac{1}{\sqrt{mn}} \|{\bf M}-{\bf M}_k\|_F=\frac{1}{\sqrt{mn}} \sum_{i=k+1}^{{\rm rank}({\bf M})} \sigma^2_i\,,
\end{equation} being $\sigma_i$ (for $i=1,\ldots,{\rm rank}({\bf M})$) the singular values of ${\bf M}$, ordered nonincreasingly. {\blue Equation (\ref{eq:Eckart-Young}) shows that either a low rank of the matrix to be completed or a fast decay to $0$ of its singular values distribution are important for a successful application of MC to that matrix, as they imply a small right-hand side of the equation. It has to be remarked, however, that they provide only a necessary condition for such a successful application since, in the context of MC, finding exactly the singular value decomposition of the matrix ${\bf M}$ is infeasible, being ${\bf M}$ only partially observed. The reader is referred to \cite{Nguyenetlal2019b} for more details about properties of a matrix and on the distribution of its sampled elements which allow a successful application of some forms of MC to that matrix. It is worth remarking that some of such properties refer to the possibility of an exact (or ``perfect'') reconstruction of the matrix, which is not really needed neither feasible in our specific application of MC to I/O matrices.}

\subsection{\blue Application of matrix completion to a WIOD submatrix containing both intra-country and inter-country blocks}\label{appendix:2}

{\blue In Table \ref{tab:2appendix}, we consider the following variation of Table \ref{tab:2}, in which we take into account also the domestic block associated with Italy, evaluated in different years.}

\begin{table}[h!]
\setlength{\tabcolsep}{3pt}
\begin{center}
\begin{tabular}{ | c | c | c | } 
\hline
\multicolumn{3}{ | c | }{\it I/O, year} \\
\hline
FRA/ITA, 2010 & ITA/FRA, 2010 & ITA/ITA, 2010 \\ 
\hline
FRA/ITA, 2011 & ITA/FRA, 2011 & ITA/ITA, 2011\\  
\hline
FRA/ITA, 2012 & ITA/FRA, 2012 & ITA/ITA, 2012\\ 
\hline
FRA/ITA, 2013 & ITA/FRA, 2013 & ITA/ITA, 2013\\ 
\hline
FRA/ITA, 2014 & ITA/FRA, 2014 & \bf{ITA/ITA, 2014}\\ 
\hline
\end{tabular}
\end{center}
\caption{\blue Structure of the WIOD submatrix used for the example reported in Figure \ref{fig:2appendix}. Similar comments as in Table \ref{tab:2} apply.}\label{tab:2appendix}
\end{table}

\begin{figure}
\vspace{-0.45cm}
\begin{subfigure}{1\textwidth}
\begin{center}
    \includegraphics[scale=0.27]{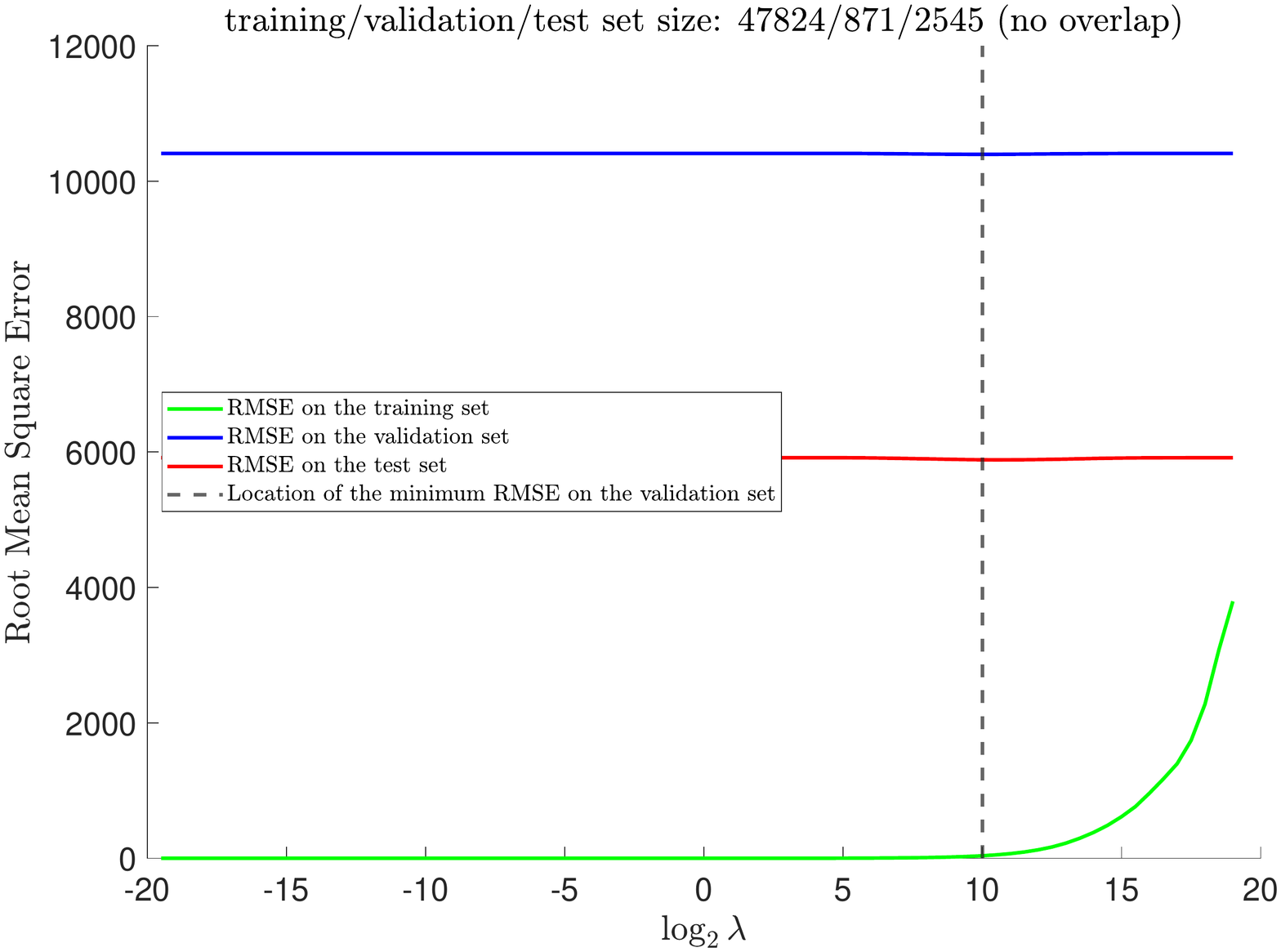}
    \vspace{-0.45cm}
    \caption{\blue Results (expressed in terms of RMSE) of the application of Algorithm \ref{alg:1} to the WIOD submatrix reported in Table \ref{tab:2appendix}.}
    \label{fig:2appendix_RMSE}
    \end{center}
    \end{subfigure}
\begin{subfigure}{1\textwidth}
\begin{center}
    \includegraphics[scale=0.27]{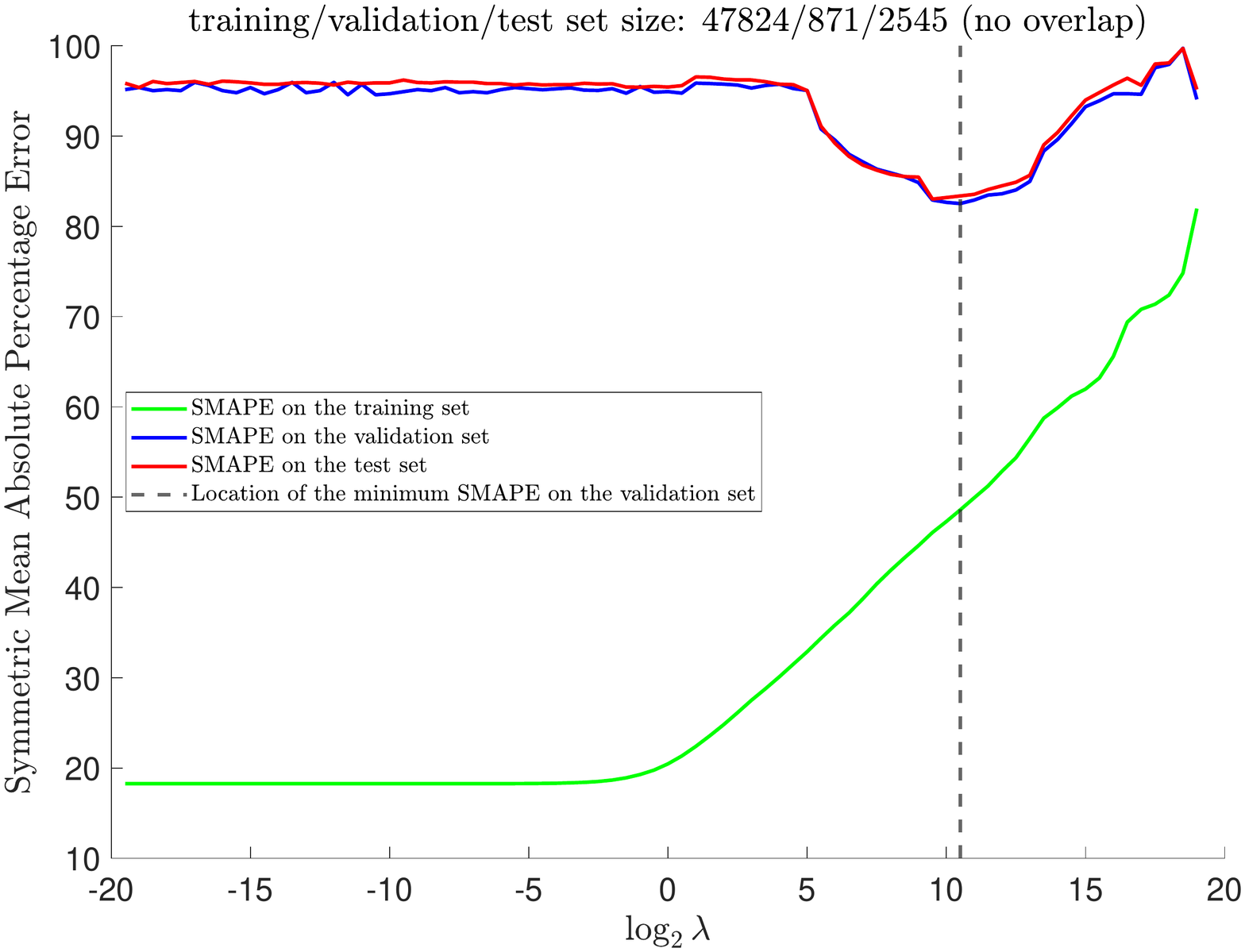}
    \vspace{-0.45cm}
    \caption{\blue Results (expressed in terms of SMAPE) of the application of Algorithm \ref{alg:1} to the WIOD submatrix reported in Table \ref{tab:2appendix}.}
    \label{fig:2varappendix_SMAPE}
    \end{center}
    \end{subfigure}
\begin{subfigure}{1\textwidth}
\begin{center}
    \includegraphics[scale=0.27]{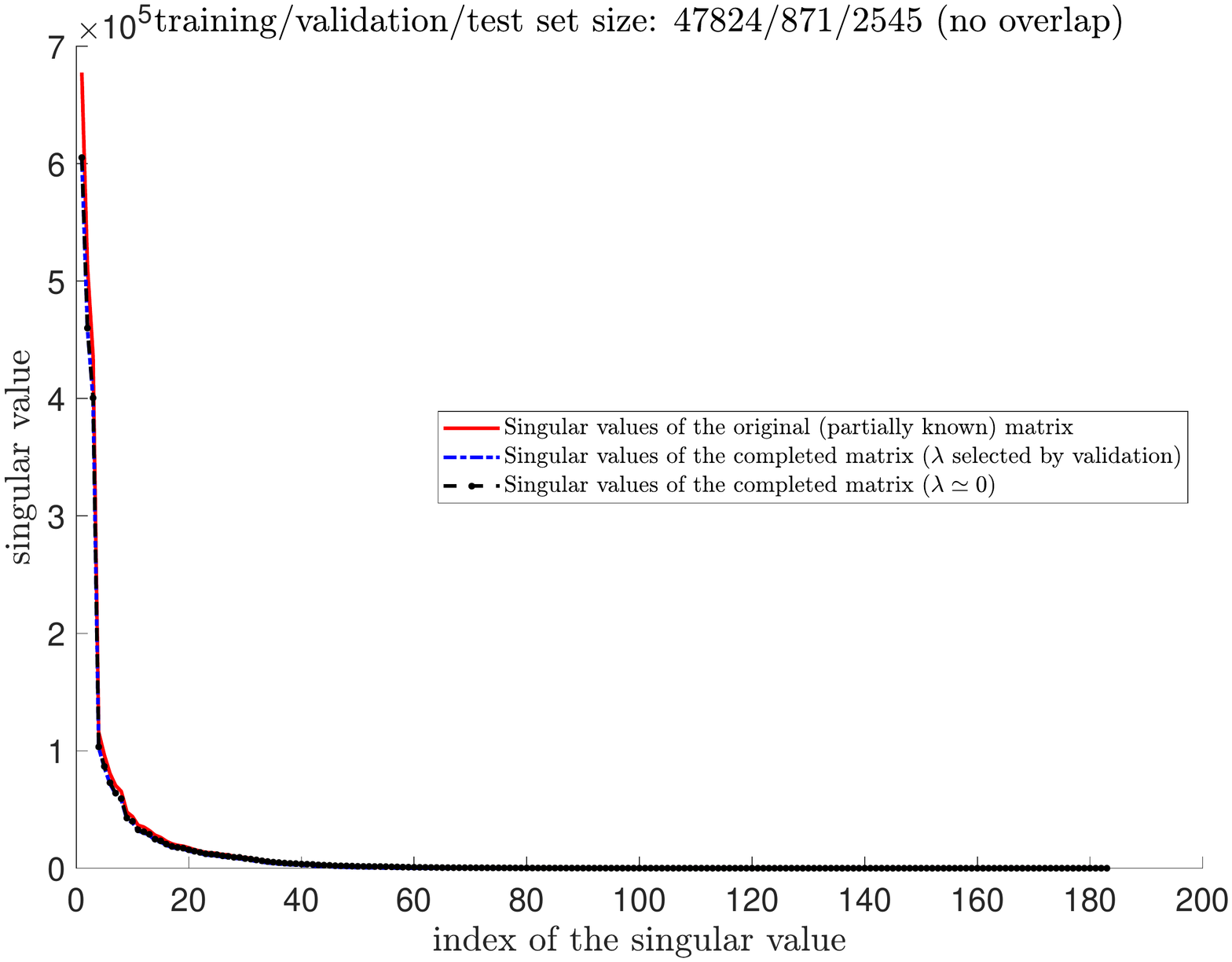}
    \vspace{-0.45cm}
    \caption{\blue Singular values distribution of the WIOD submatrix reported in Table \ref{tab:2appendix}, and the one of the completed submatrix produced 
    by Algorithm \ref{alg:1} for the optimal regularization constant (RMSE criterion).}
    \label{fig:3appendix_spectra}
\end{center}
\end{subfigure}
\begin{subfigure}{1\textwidth}
\begin{center}
    \includegraphics[scale=0.27]{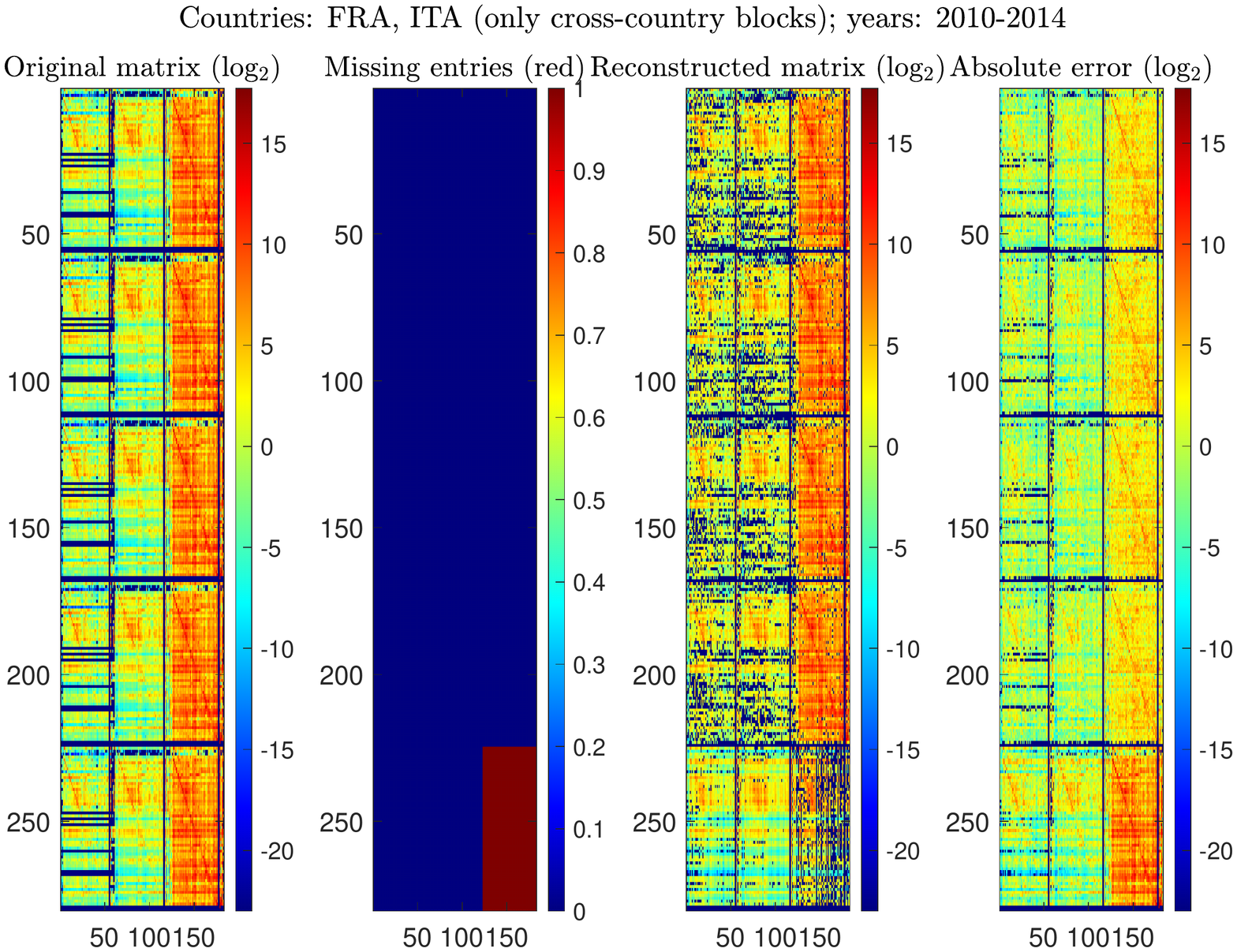}
    \vspace{-0.45cm}
    \caption{\blue Colored visualization of the elements of the WIOD submatrix reported in Table \ref{tab:2appendix}, positions of the missing entries, reconstructed submatrix obtained for the optimal regularization constant (RMSE criterion), and element-wise absolute value of the reconstruction error.}
    \label{fig:4appendix_various}
    \end{center}
    \end{subfigure}
    \caption{}\label{fig:2appendix}
\end{figure}

{\blue For what concerns the application of MC, due to the different orders of magnitude of the elements contained in the domestic blocks compared to the ones belonging to the other blocks, the range of values for the regularization parameter has been increased for this specific example, by setting  $\lambda_k=2^{k/2-20}$, for $k=1,\ldots,80$. As highlighted by Figure \ref{fig:2appendix}, in this case, the performance of MC is quite bad, likely due to the highly different orders of magnitude of the elements in the various blocks.}

\subsection{\blue Alternative displacement of the blocks}\label{appendix:3}

{\blue It is worth observing that the several blocks of the WIOD submatrix reported in Table \ref{tab:2} have been displaced in such a way to make the application of MC possible. To clarify this issue, an alternative displacement is reported in Table \ref{tab:3}.

\begin{table}[h!]
\setlength{\tabcolsep}{3pt}
\begin{center}
\begin{tabular}{ | c | } 
\hline
\multicolumn{1}{ | c | }{\it I/O, year} \\
\hline
FRA/ITA, 2010 \\
\hline
ITA/FRA, 2010 \\ 
\hline
FRA/ITA, 2011 \\
\hline
ITA/FRA, 2011 \\ 
\hline
FRA/ITA, 2012 \\
\hline
ITA/FRA, 2012 \\ 
\hline
FRA/ITA, 2013 \\
\hline
ITA/FRA, 2013 \\ 
\hline
FRA/ITA, 2014 \\
\hline
\bf{ITA/FRA, 2014} \\ 
\hline
\end{tabular}
\end{center}
\caption{Alternative displacement of the blocks for the WIOD submatrix reported in Table \ref{tab:2}. All the entries contained in the block highlighted in {\blue bold} are obscured. In this case, the optimization problem (\ref{eq:matrix_completion2}) is {\blue  trivial}.}\label{tab:3}
\end{table}

In this second case, some rows of the submatrix are obscured as a whole, and the optimal solution to the optimization problem (\ref{eq:matrix_completion2}) is trivial, in the sense that the obscured rows are reconstructed by setting all their elements to $0$. Indeed, given a partially observed matrix ${\bf P}_{\Omega^{\rm tr}}({\bf M})$ and setting $\varepsilon$ to $0$ and $N^{\rm it}$ by $+\infty$ in Algorithm \ref{alg:1}, it is well known from the convergence of that algorithm  (see \cite[Lemma 5]{Mazumder2010}) that the sequence of solutions ${\bf \hat{M}}^{\rm new}$ it generates tends to an optimal solution ${\bf \hat{M}}_{\lambda^\circ}$ of the optimization problem (\ref{eq:matrix_completion2}). 
In case the $i$-th row of ${\bf M}$ is not observed, one can show\footnote{\blue In this case, if at a generic iteration of Algorithm \ref{alg:1} the $i$-th row of the matrix ${\bf \hat{M}}^{\rm old}$ is made of all zeros (and this surely holds at the initialization of the algorithm), then also the $i$-th row of the matrix ${\bf N}^{\rm old}={\bf P}_{\Omega^{\rm tr}}({\bf M})+{\bf P}_{\Omega^{\rm tr}}^{\perp}({\bf \hat{M}}^{\rm old})$ is made of all zeros. Let ${\bf N}^{\rm old}={\bf U}^{\rm old} {\bf \Sigma}^{\rm old} \left({\bf V}^{\rm old}\right)^T$ be the singular value decomposition of such matrix. Then, when {\blue  the index $j$ is associated with a positive singular value of ${\bf N}^{\rm old}$, the element in position $(i,j)$  of ${\bf U}^{\rm old}$ is equal to $0$ (this is shown by finding the singular value decomposition of the matrix obtained from ${\bf N}^{\rm old}$ by removing its $i$-th row, then obtaining the singular value decomposition of ${\bf N}^{\rm old}$ by extending with zero components the resulting left-singular vectors associated with positive singular values). Since the matrix ${\bf \hat{M}}^{\rm new}={\bf S}_\lambda\left({\bf N}^{\rm old}\right)$ has the same left-singular vectors as ${\bf N}^{\rm old}$, also the $i$-th row of ${\bf \hat{M}}^{\rm new}$ is made by all zeros. Finally, since such matrix becomes ${\bf \hat{M}}^{\rm old}$ at the successive iteration of Algorithm \ref{alg:1}, this property holds for all the matrices ${\bf \hat{M}}^{\rm new}$ generated by that algorithm.}} that all the elements {\blue in the same $i$-th row of each matrix ${\bf \hat{M}}^{\rm new}$ generated by the algorithm (which is renamed as ${\bf \hat{M}}^{\rm old}$ at the end of the corresponding iteration) are equal to $0$.  
Then, by induction and convergence, also all the elements in the $i$-th row of ${\bf \hat{M}}_{\lambda^\circ}$ are equal to $0$. More generally, the problem addressed by low-rank MC is {\blue either trivial (if a regularization term is present) or} ill-posed {(if there is no regularization term)} when some rows or columns of the matrix to be completed are obscured as a whole, since the other rows/columns provide no information on the obscured ones \cite{CandesTao2010}. {\blue In both cases, the application of MC becomes meaningless}.}}

{\blue In our specific application of MC to I/O subtables, the arrangement of blocks reported in Table \ref{tab:2} makes this issue of {\blue  triviality/ill-posedness of the problem addressed by MC} disappear (while such issue is present in the case of the arrangement of blocks reported in Table \ref{tab:3}). This can be justified as follows. In the case of Table \ref{tab:2}, the particular arrangement of the blocks makes it possible for MC to ``learn'', if present, a relationship between the observed blocks ``FRA/ITA, $x$'' and ``ITA/FRA, $x$'' in the same year $x$, where $x=2010, 2011, 2012, 2013$. Then, when the block ``FRA/ITA, 2014'' is observed, such a relationship makes it possible to estimate (although with some error) the block ``ITA/FRA, 2014'', which is not observed. In the case of Table \ref{tab:3}, the different arrangement of the same blocks makes impossible for MC to ``learn'' the relationship above\footnote{\blue  Loosely speaking, this conclusion comes from the fact that, if one applies the Soft Impute algorithm to two partially observed matrices that differ only by a permutation of their rows and if also the matrix representing the locations of the observed/unobserved entries is subject to the same permutation, then the two respective completed matrices found by the algorithm differ only by the same permutation. In the case of Table \ref{tab:3}, this means that MC fails to distinguish if the missing block ``ITA/FRA, $2014$'' has indeed the stated form, or if it is instead another block of the form ``FRA/ITA, $x$'' with $x=2010, 2011, 2012, 2013, 2014$, or one of the form ``ITA/FRA, $y$'', with $y \neq 2014$.}, so the MC estimate of the block ``ITA/FRA, 2014'' is now arbitrary.}

{\blue It is worth mentioning that the above-mentioned issue of triviality/ill-posedness of the problem addressed by MC does not actually arise from possibly having either a row or a column made of all zeros in the matrix to be completed, but from the fact of observing no element in such row or column. For simplicity, in the following we illustrate this issue for the case in which one looks for a completed matrix $\hat{\bf M}$ having zero error on the set of observed entries (this is not exactly the case considered in the paper, in which a nonzero error is actually allowed, but the two settings are clearly related, as the first one arises as a limiting case of the optimization problem (\ref{eq:matrix_completion2}) when $\lambda$ tends to $0^+$). So, suppose that one knows that the matrix ${\bf M}$ to be completed has rank $1$, and let the symbol $*$ be used to denote each of its unobserved entries. If ${\bf M}=\begin{bmatrix} 1 & 2 \\
* & 0 \\
3 & *\end{bmatrix}$, then its unique rank-$1$ reconstruction (with no error in the observed entries) is $\hat{\bf M}=\begin{bmatrix} 1 & 2 \\
0 & 0 \\
3 & 6\end{bmatrix}$. Instead, if one has ${\bf M}=\begin{bmatrix} 1 & 2 \\
0 & 0 \\
* & *\end{bmatrix}$, then there is no unique rank-$1$ reconstruction of that matrix. On the other hand, taking an arbitrary such reconstruction would be useless.} 

\subsection{\blue Performance of matrix completion on simulated matrices}\label{appendix:4}

\begin{figure}
\vspace{-0.45cm}
\begin{subfigure}{1\textwidth}
\begin{center}
    \includegraphics[scale=0.27]{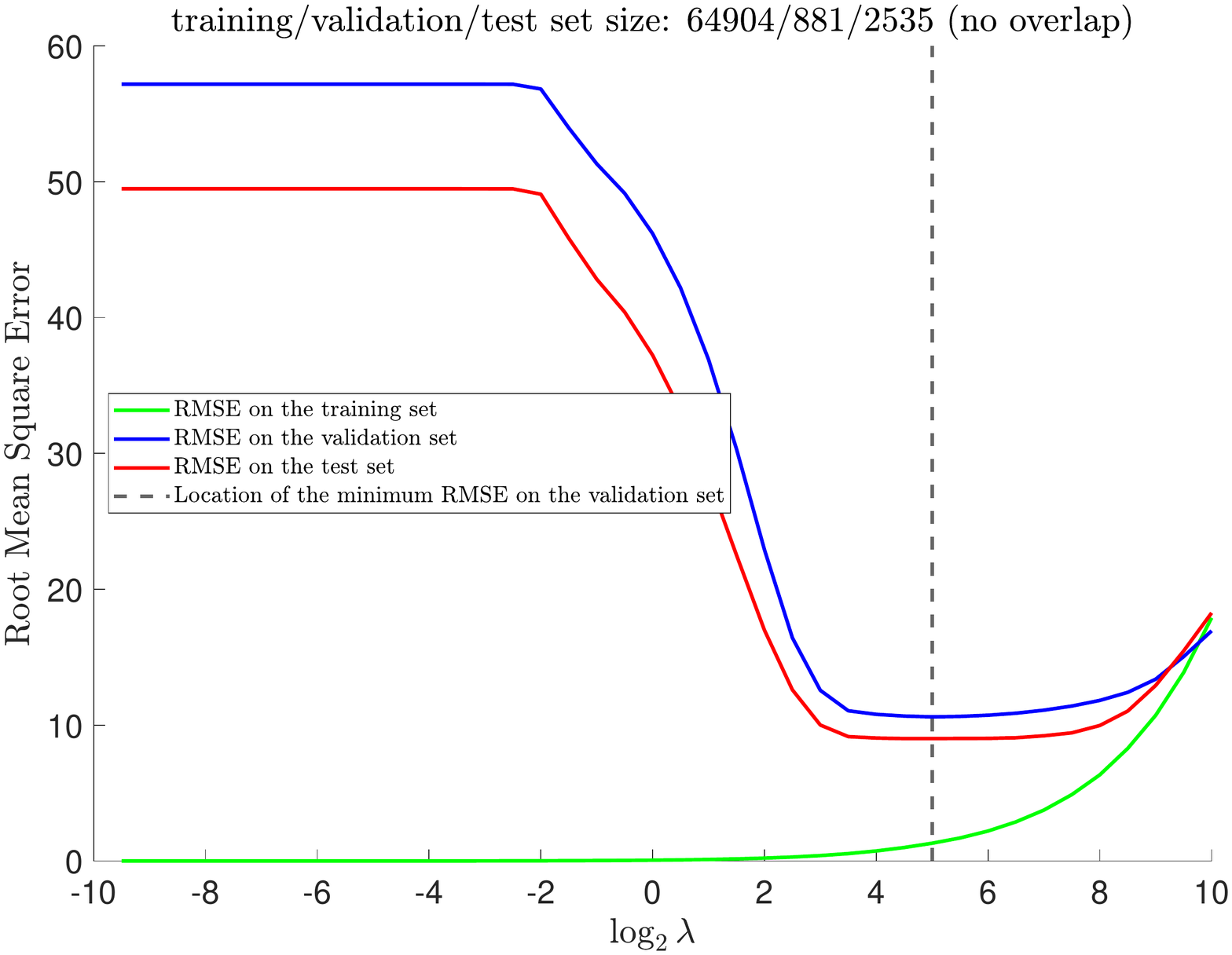}
    \vspace{-0.45cm}
    \caption{\blue Results (expressed in terms of RMSE) of the application of Algorithm \ref{alg:1} to one synthetic WIOD submatrix made of 4 similar synthetic countries, with Italy in input.}
    \label{fig:2appendix_synthetic_similar_input}
    \end{center}
    \end{subfigure}
\begin{subfigure}{1\textwidth}
\begin{center}
    \includegraphics[scale=0.27]{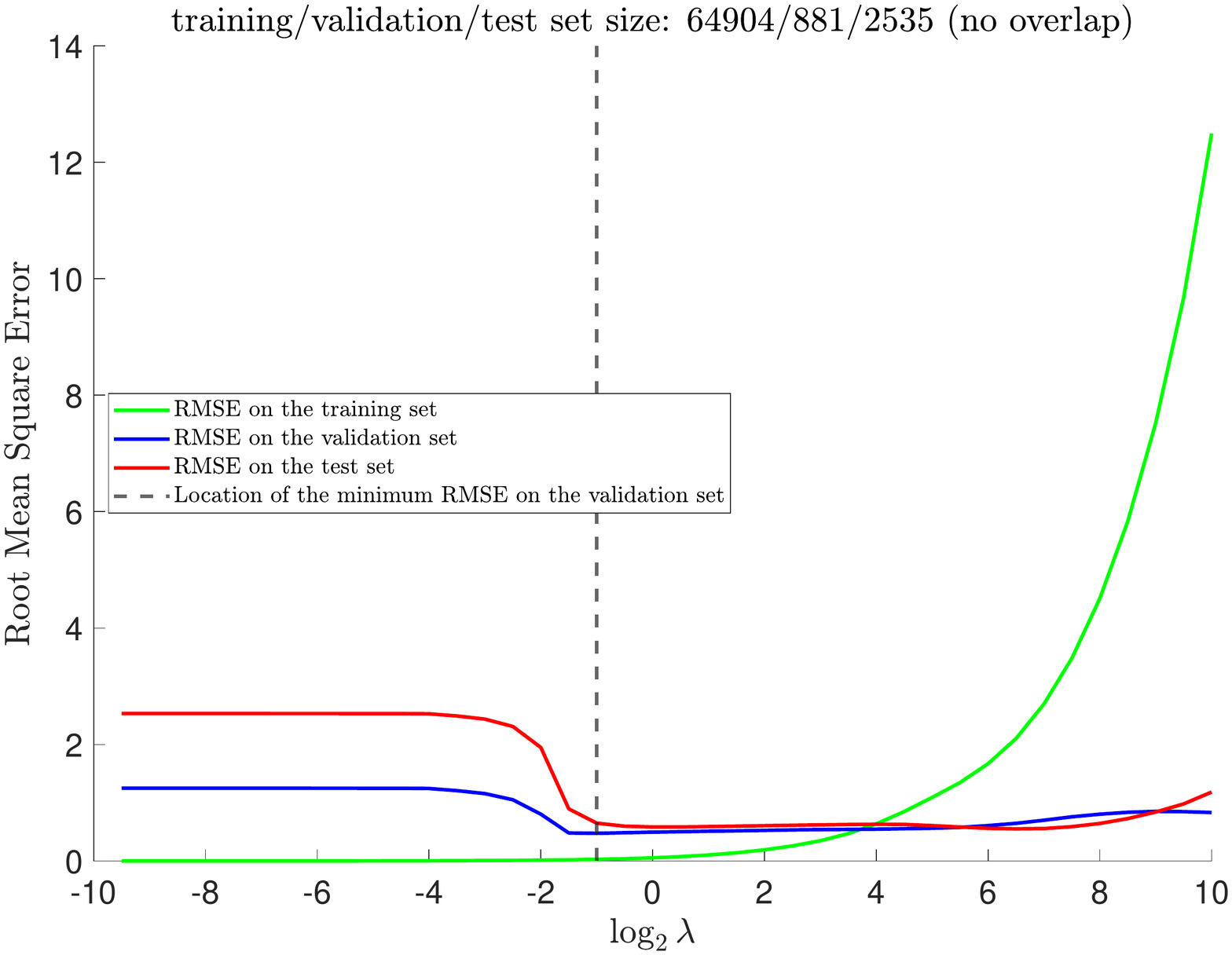}
    \vspace{-0.45cm}
    \caption{\blue Results (expressed in terms of RMSE) of the application of Algorithm \ref{alg:1} to one synthetic WIOD submatrix made of 4 similar synthetic countries, with Italy in input.}
    \label{fig:2appendix_synthetic_similar_input}
    \end{center}
    \end{subfigure}
    \caption{}\label{fig:2appendix_synthetic_input}
\end{figure}

\begin{figure}
\vspace{-0.45cm}
\begin{subfigure}{1\textwidth}
\begin{center}
    \includegraphics[scale=0.27]{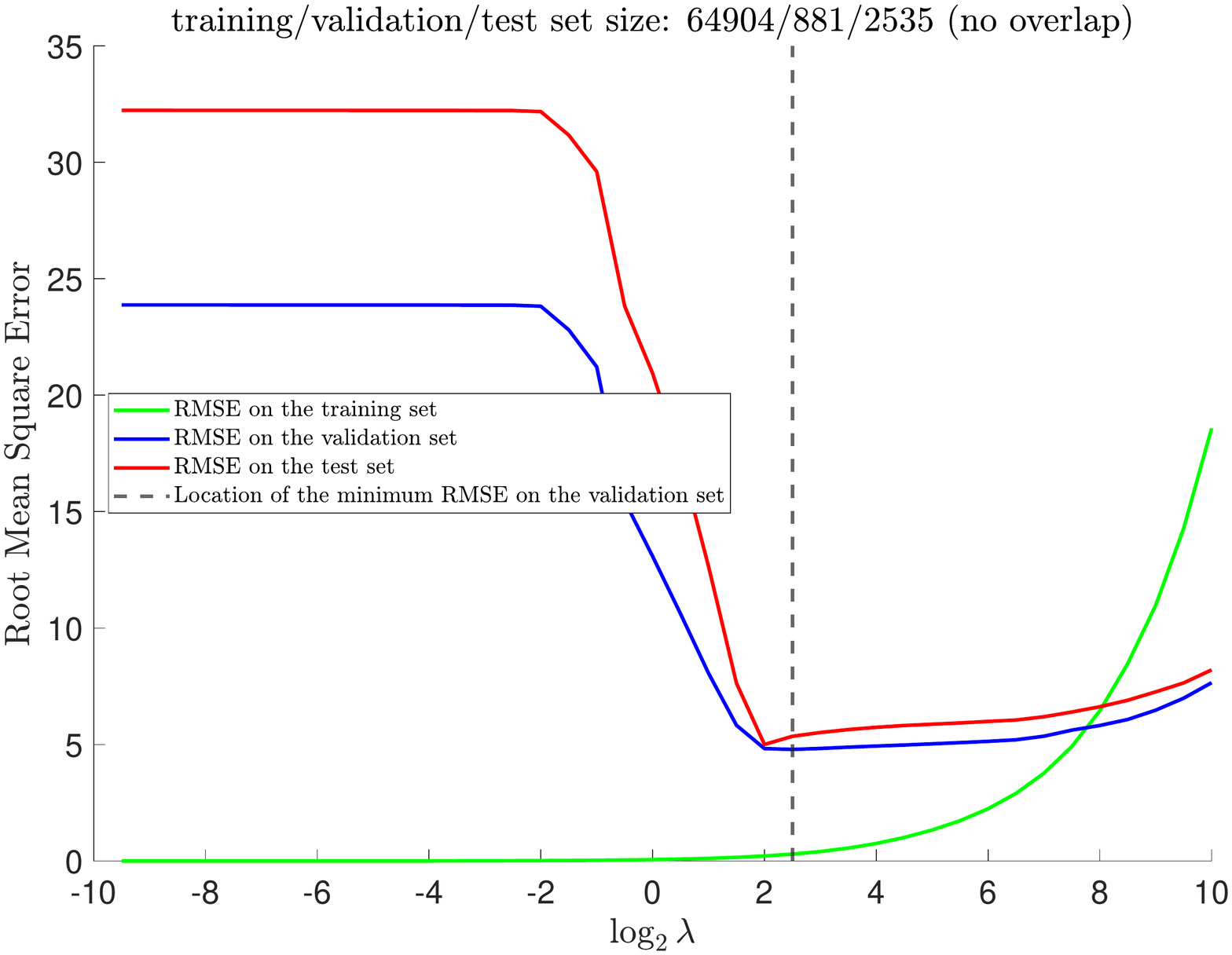}
    \vspace{-0.45cm}
    \caption{\blue Results (expressed in terms of RMSE) of the application of Algorithm \ref{alg:1} to one synthetic WIOD submatrix made of 4 similar synthetic countries, with Italy in output.}
    \label{fig:2appendix_synthetic_similar_output}
    \end{center}
    \end{subfigure}
\begin{subfigure}{1\textwidth}
\begin{center}
    \includegraphics[scale=0.27]{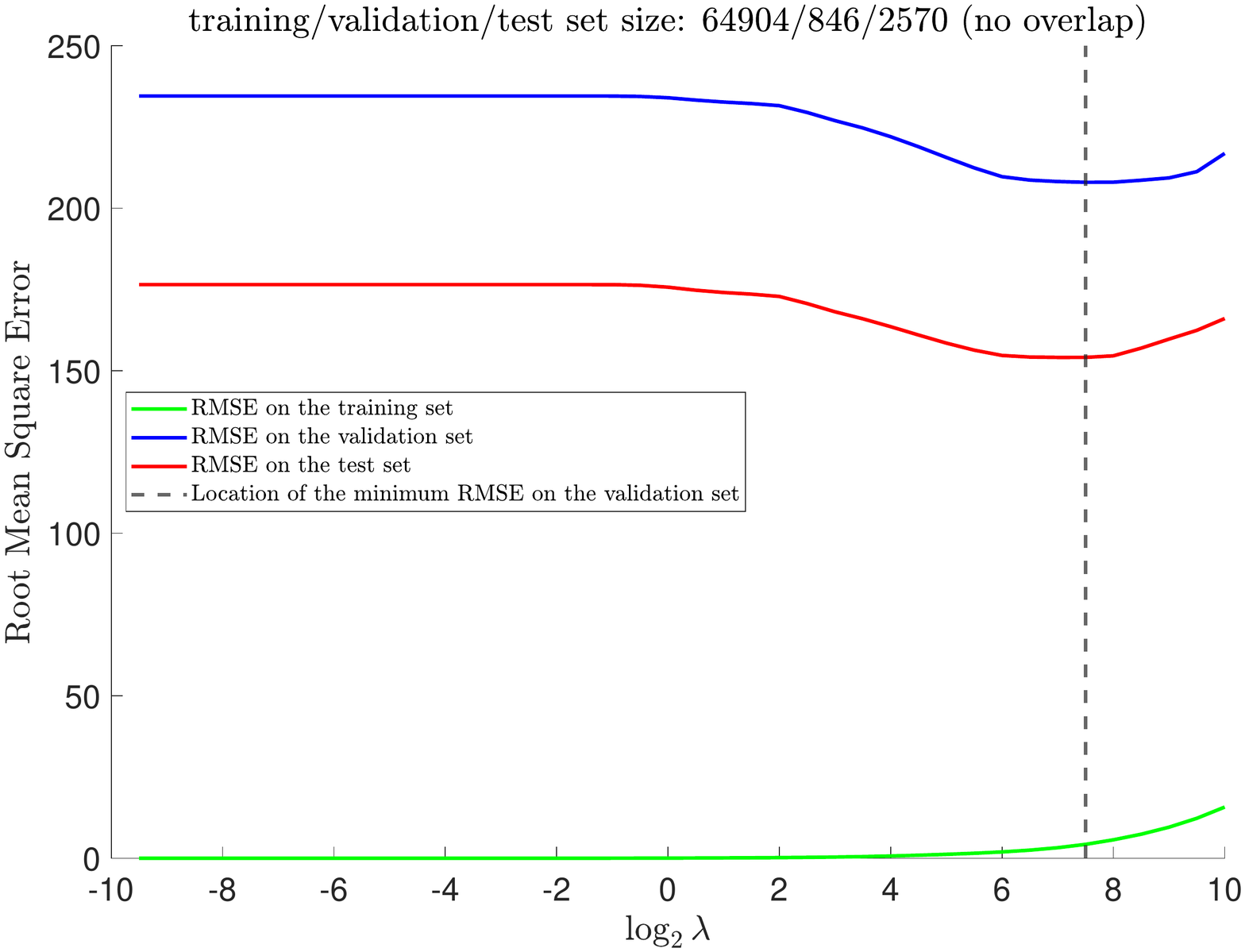}
    \vspace{-0.45cm}
    \caption{\blue Results (expressed in terms of RMSE) of the application of Algorithm \ref{alg:1} to one synthetic WIOD submatrix made of 4 dissimilar synthetic countries, with Italy in output.}
    \label{fig:2appendix_synthetic_dissimilar_output}
    \end{center}
    \end{subfigure}
    \caption{}\label{fig:2appendix_synthetic_output}
\end{figure}

{\blue In this last appendix, we show that the application of MC on the simulated data of Subsection \ref{sec:preprocessing} produces similar results as its application to the original data. In the following, for illustrative purposes, we focus just on one of the simulated matrices considered in that subsection (in the next figures, the synthetic countries are still named as the original countries, since their respective data are obtained by perturbations of the ones of the associated original countries).

Figures \ref{fig:dendrograms_input_synthetic} and \ref{fig:dendrograms_output_synthetic} show the results of the hierarchical clustering, obtained respectively with Italy in input and in output.
\begin{figure}
    \begin{center}
     \includegraphics[scale=0.61,trim=0.5cm 9.5cm 0 8.7cm]{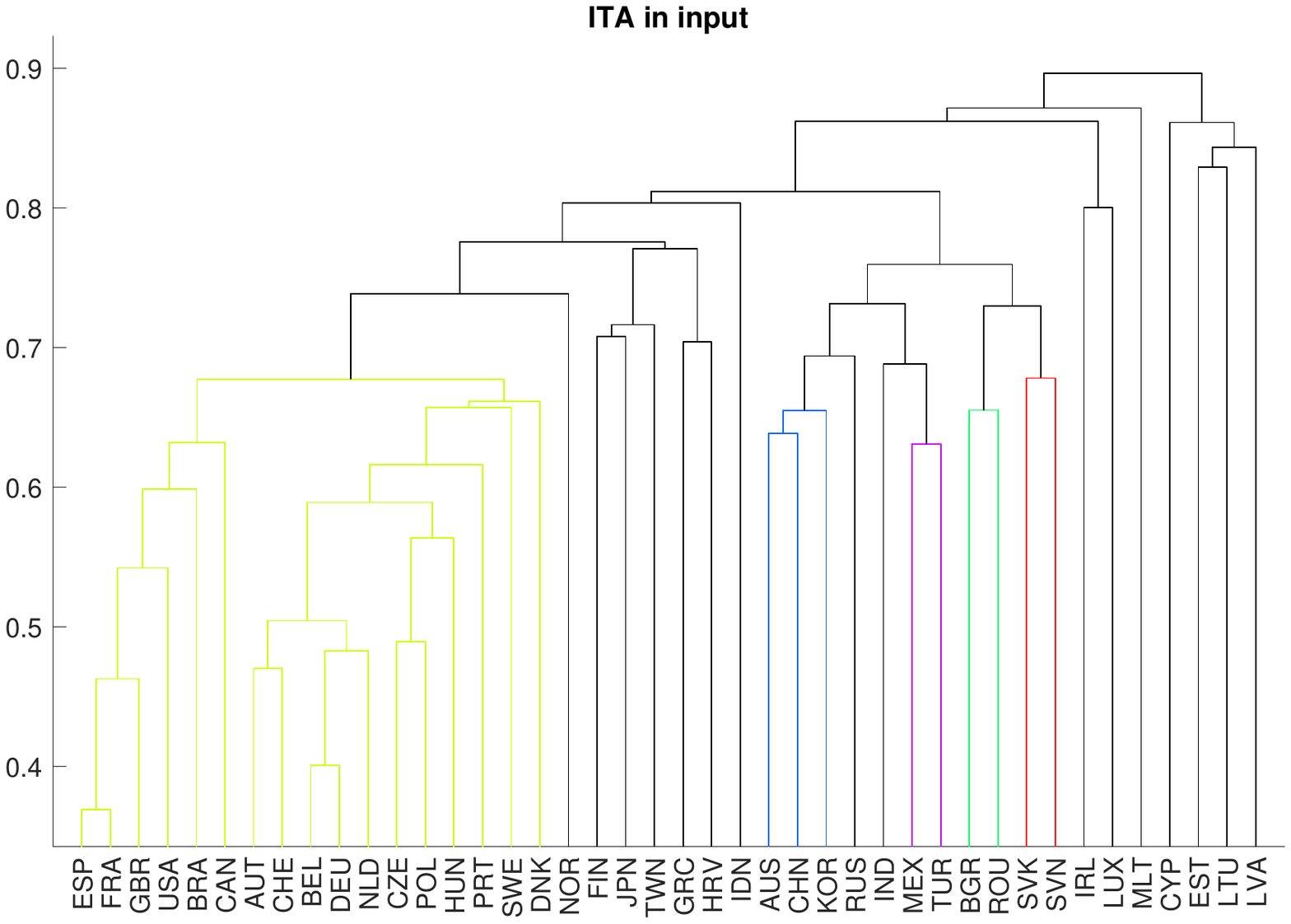}
    \end{center}
    \caption{\blue Dendrograms of synthetic output countries with Italian sectors in input, based on synthetic 
    WIOD tables (stacked over the years 2010--2013). Hierarchical clustering performed with the AACD dissimilarity measure ($y$-axis) and complete linkage. 21 desired groups (compare with Table \ref{tab:numclus_wss}).
    Countries in the same cluster are depicted with the same color. Countries in singleton clusters are highlighted in black.}\label{fig:dendrograms_input_synthetic}
\end{figure}
\begin{figure}
    \begin{center}
     \includegraphics[scale=0.61,trim=0.5cm 9.5cm 0 8.7cm]{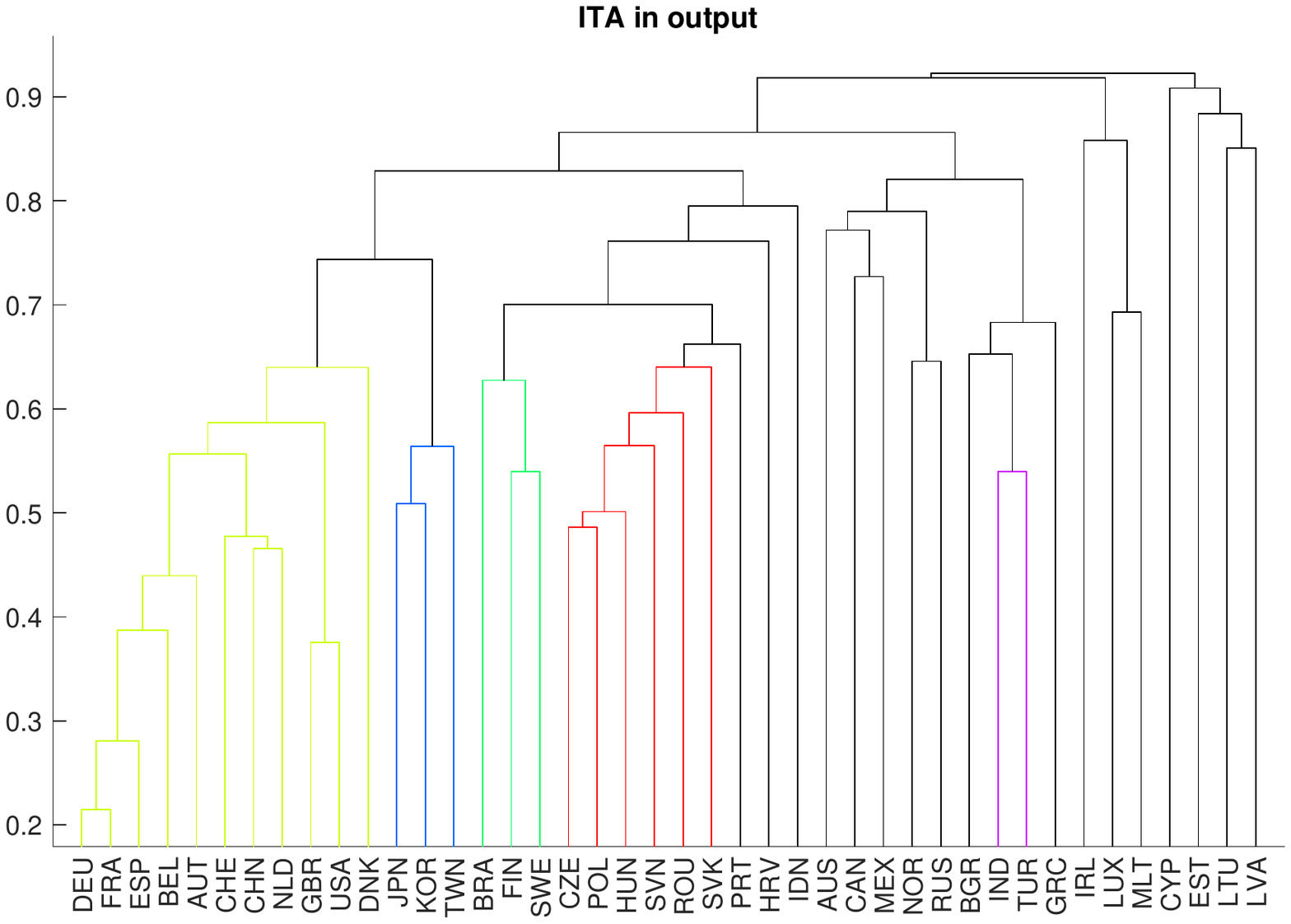}
    \end{center}
    \caption{\blue Dendrograms of synthetic input countries with Italian sectors in output, based on 
    synthetic WIOD tables (stacked over the years 2010--2013). Hierarchical clustering performed with the AACD dissimilarity measure ($y$-axis) and complete linkage. 22 desired groups (compare with Table \ref{tab:numclus_wss}).
    Countries in the same cluster are depicted with the same color. Countries in singleton clusters are highlighted in black.}\label{fig:dendrograms_output_synthetic}
\end{figure}
 Then, based on the dendrogram shown in the Figure \ref{fig:dendrograms_input_synthetic}, Figure \ref{fig:2appendix_synthetic_input} compares the MC performance (for short, limiting to the RMSE criterion) for the cases -- analogous to the ones considered in Subsection \ref{sec:similardissimilar} -- in which the 4 selected synthetic countries belong respectively to the same cluster (similar synthetic countries: ESP, FRA, GBR, USA, obscured one in the last year: ESP) and to different clusters (dissimilar synthetic countries: CYP, ESP, IDN, MEX, obscured one in the last year: CYP). Analogously, based on the dendrogram shown in the Figure \ref{fig:dendrograms_output_synthetic}, Figure \ref{fig:2appendix_synthetic_output} compares the MC performance (again, limiting to the RMSE criterion) for the cases -- analogous to the ones considered in Subsection \ref{sec:similardissimilar} --  in which the 4 selected synthetic countries belong respectively to the same cluster (similar synthetic countries: BEL, DEU, ESP, FRA, obscured one in the last year: BEL) and to different clusters (dissimilar synthetic countries: CZE, DEU, EST, IND, obscured one in the last year: DEU). The results are qualitatively similar to the ones reported in for the original data, and demonstrate the robustness of the proposed approach of analysis, which combines hierarchical clustering and MC. Similar results, not reported here due to space constraints, are obtained when the SMAPE criterion is used to compare the performance of MC for similar and dissimilar countries.} 

\newpage
\bibliographystyle{spmpsci}      

\begin{thebibliography}{}
%
%

\bibitem{AggarwalReddy2014}
Aggarwal, C. C., \& Reddy, C. K., Eds., Data clustering: Algorithms and applications. Chapman \& Hall/CRC Press, London (2014)

\bibitem{Aldstadt2010}
Aldstadt, J., Spatial clustering. In Handbook of applied spatial analysis, pp. 279--300 (Fischer, M. M., \& Getis, A., Eds.). Springer, Berlin, Heidelberg (2010)


\bibitem{BhattacharyaBhandariBairagya2020}
Bhattacharya, T., Bhandari, B., \& Bairagya, I., Where are the jobs? Estimating skill-based employment linkages across sectors for the Indian economy: An input-output analysis, Structural Change and Economic Dynamics, 53, 292--308 (2020) 

\bibitem{CandesTao2010}
Cand\`es, E.J., \& Tao, T., The power of convex relaxation:
Near-optimal matrix completion. IEEE Transactions on Information  Theory, 56(5), 2053--2080 (2010)

\bibitem{Carvalho2009}
Carvalho, V., Aggregate fluctuations and the network structure of intersectoral trade. Working Papers - Universitat Pompeu Fabra. Departamento de Economía y Empresa, 1206(1) (2009)

\bibitem{Cerina2015}
Cerina, F., Zhu, Z., Chessa, A., \&  Riccaboni M., World input-output network. PloS One, 10(7), e0134025 (2015)

\bibitem{ChenWuGuoMengLi2019}
Chen, G. Q., Wu, X. D., Guo, J., Meng, J., \& Li, C., Global overview for energy use of the world economy: Household-consumption-based accounting based on the world input-output database (WIOD). Energy Economics, 81, 835--847 (2019)

\bibitem{FagioloReyes2008}
Fagiolo, G., Reyes, J., \& Schiavo, S., On the topological properties of the world trade web: A weighted network analysis. Physica A: Statistical Mechanics and its Applications, 387(15), 3868--3873 (2008)

\bibitem{Fernandez-Vazquez}
Fernandez-Vazquez, E., A generalized cross entropy formulation for matrix balancing with both positive and negative entries (2016, unpublished)

\bibitem{Hastie2015}
Hastie, T., Tibshirani, R., \& Wainwright, M., Statistical learning with sparsity: the Lasso and generalizations. 
CRC Press, New York (2015)

\bibitem{IoriDeMasi2008}
Iori, G., De Masi, G., Precup, O. V., Gabbi, G., \& Caldarelli, G., A network analysis of the Italian overnight money market. Journal of Economic Dynamics and Control, 32(1), 259--278 (2008)


\bibitem{LanceWilliams1967}
Lance, G.N. \& Williams, W.T., A general theory of classificatory sorting strategies. 1. Hierarchical systems. The Computer Journal, 9 (4), 373--380 (1967)

\bibitem{Lenzen2012}
Lenzen, M., Kanemoto, K., Moran, D., \& Geschke, A., Mapping the structure of the world economy. Environmental Science \& Technology, 46(15), 8374-8381 (2012)

\bibitem{Leontief1986}
Leontief, W. (Ed.), Input-output economics. Oxford University Press. New York (1986)

\bibitem{LiZhou2017}
Li, C., \& Zhou, H., Svt: Singular value thresholding in MATLAB. Journal of Statistical Software, 81(2), DOI: 10.18637/jss.v081.c02 (2017)

\bibitem{LiangQi2016}
Liang, S., Qi, Z., Qu, S., Zhu, J., Chiu, A. S., Jia, X., \& Xu, M., Scaling of global input–output networks. Physica A: Statistical Mechanics and its Applications, 452, 311--319 (2016)



\bibitem{MacQueen1967}
MacQueen, J., Some methods for classification and analysis of multivariate observations. In Proceedings of the fifth Berkeley Symposium on Mathematical Statistics and Probability, 1 (14), 281--297 (1967)

\bibitem{Mazumder2010}
Mazumder, R., Hastie, T., \& Tibshirani, R., Spectral regularization algorithms for learning large incomplete matrices. Journal of Machine Learning Research, 11, 2287--2322 (2010)

\bibitem{McNerneyFath2013}
McNerney, J., Fath, B. D., \& Silverberg, G., Network structure of inter-industry flows. Physica A: Statistical Mechanics and its Applications, 392(24), 6427--6441 (2013)

\bibitem{Metulini2017}
Metulini, R., Riccaboni, M., Sgrignoli, P., \& Zhu, Z., The indirect effects of foreign direct investment on trade: A network perspective. The World Economy, 40(10), 2193--2225 (2017)


\bibitem{Moitra2018}
Moitra, A., Algorithmic aspects of machine learning. Cambridge University Press, Cambridge (2018)

\bibitem{MurtaghLegendre2014}
Murtagh, F., \& Legendre, P., Ward's hierarchical agglomerative clustering method: which algorithms implement Ward's criterion? Journal of Classification, 31, 274--295 (2014)

\bibitem{Negahban2012}
Negahban, S., \& Wainwright, M. J., Restricted strong convexity and weighted matrix completion: Optimal bounds with noise. Journal of Machine Learning Research, 13(1), 1665--1697 (2012)

\bibitem{Nguyenetlal2019}
Nguyen, L.T., Kim, J., Kim, S., \& Shim, B., Localization of IoT networks via low-rank matrix completion. IEEE Transactions on Communications, 67(8): 5833--5847 (2019)

\bibitem{Nguyenetlal2019b}
Nguyen, L.T., Kim, J., \& Shim, B., Low-rank matrix completion: A contemporary survey. IEEE Access, 7: 94215--94237 (2019)

\bibitem{OECD2018}
OECD, Input-Output Tables. For download at http://oe.cd/i-o. Organisation for Economic Co-operation and Development, Paris (2018)

\bibitem{OlivaSetolaPanzieri2016}
Oliva, G., Setola, R., \& Panzieri, S., Critical clusters in interdependent
economic sectors: A data-driven spectral clustering analysis. The European Physical Journal Special Topics, 225: 1929–-1944 (2016)

\bibitem{Paviaetal2009}
Pavia, J. M., Cabrer, B., \& Sala, R., Updating input–output matrices: assessing alternatives through simulation. Journal of Statistical Computation and Simulation, 79(12), 1467--1482 (2009)

\bibitem{Percoco2006}
Percoco, M., Hewings, G., \& Senn, L., Structural change decomposition through a global sensitivity analysis of input-output models. Economic Systems Research, 18(2): 115--131 (2006)

\bibitem{Remond2019}
Rémond-Tiedrez, I., \& Rueda-Cantuche, J. M., Full international and global accounts for research in input-output analysis (FIGARO). Eurostat (2019)

\bibitem{Revelle1979}
Revelle, W., Hierarchical cluster analysis and the internal structure of tests. Multivariate Behavioral Research, 14(1), 57--74 (1979)

\bibitem{Riccaboni2019}
Riccaboni, M., Wang, X., \& Zhu, Z., Firm performance in networks: The interplay between firm centrality and corporate group size. Journal of Business Research, 129(C), 641--653 (2021)

\bibitem{Rousseeuw1987}
Rousseeuw, P. J., Silhouettes: a graphical aid to the interpretation and validation of cluster analysis. Journal of Computational and Applied Mathematics, 20, 53--65 (1987)

\bibitem{Rueda-Cantucheetal2018}
Rueda-Cantuche, J. M., Amores, A. F., Beutel, J., \& Remond-Tiedrez, I., Assessment of European use tables at basic prices and valuation matrices in the absence of official data. Economic Systems Research, 30(2), 252--270 (2018)

\bibitem{Sgrignoli2015}
Sgrignoli, P., Metulini, R., Schiavo, S., \& Riccaboni, M., The relation between global migration and trade networks. Physica A: Statistical Mechanics and its Applications, 417, 245--260 (2015)

\bibitem{Thorndike1953}
Thorndike, R. L., Who belongs in the family?. Psychometrika, 18(4), 267--276 (1953)

\bibitem{Tibshirani1996}
Tibshirani, R., Regression shrinkage and selection via the Lasso. Journal of the Royal Statistical Society. Series B (Methodological), 58(1), 267--288 (1996)

\bibitem{Tibshirani2001}
Tibshirani, R., Walther, G., \& Hastie, T., Estimating the number of clusters in a data set via the gap statistic. Journal of the Royal Statistical Society: Series B (Statistical Methodology), 63(2), 411--423 (2001)

\bibitem{Timmer2015}
Timmer, M. P., Dietzenbacher, E., Los, B., Stehrer, R., \& De Vries, G. J., An illustrated user guide to the world input–output database: the case of global automotive production. Review of International Economics, 23(3), 575--605 (2015)

\bibitem{Timmer2016}
Timmer, M. P., Los, B., Stehrer, R., \& De Vries, G. J., An anatomy of the global trade slowdown based on the WIOD 2016 release. GGDC research memorandum number 162, University of Groningen, available online at the following hyperlink: \url{https://www.rug.nl/ggdc/html_publications/memorandum/gd162.pdf} (2016)

\bibitem{Tukker2013}
Tukker, A., De Koning, A., Wood, R., Hawkins, T., Lutter, S., Acosta, J., Rueda Cantuche, M., Bouwmeester, M., Oosterhaven, J., Drosdowski, T., \& Kuenen, J., EXIOPOL–development and illustrative analyses of a detailed global MR EE SUT/IOT. Economic Systems Research, 25(1), 50--70 (2013)

\bibitem{UN2018}
United Nations, Handbook on supply and use tables and
input-output tables with extensions and
applications (2018)

\bibitem{Valderas-Jaramilloetal2019}
Valderas-Jaramillo, J. M., Rueda-Cantuche, J. M., Olmedo, E.,  \& Beutel, J., Projecting supply and use tables: new variants and fair comparisons. Economic Systems Research, 31(3), 423--444 (2019)

\bibitem{Valderas-Jaramilloetal2021}
Valderas-Jaramillo, J. M., Rueda-Cantuche, J. M., \& Beutel, J., The Euro and SUT-RAS methods: some further considerations. Economic Systems Research, 33(2), 276--286 (2021)

\bibitem{Wangetal2015}
Wang, H., Wang, C., Zheng, H., Feng, H., Guan, R., \& Long, W., Updating input–output tables with benchmark table series. Economic Systems Research, 27(3), 287--305 (2015)

\bibitem{WangDingGuanZia2020}
Wang, H., Ding, L., Guan, R., \& Xian, Y., Effects of advancing internet technology on Chinese employment: a spatial study of inter-industry spillovers. Technological Forecasting and Social Change, 161, 120259 (2020)

\bibitem{WenXuWen2014}
Wen, S., Xu, F., \& Wen, Z., Robust linear optimization under matrix completion. Science China Mathematics, 57(4), 699--710 (2014)

\bibitem{XuLiang2019}
Xu, M., \& Liang, S., Input–output networks offer new insights of economic structure. Physica A: Statistical Mechanics and its Applications, 527, 121178 (2019)

\bibitem{XuLinHeWen2014}
Xu, F., Lin, C., He, G., \& Wen, Z., Nonnegative matrix completion for life-cycle assessment and input-output analysis. CER working paper, available online at the following hyperlink:  \url{http://www.cer.sdu.edu.cn/info/1093/3013.htm} (2014)

\bibitem{Yonemoto2016}
Yonemoto, K., Changes in the input–output structures of the six regions of Fukushima, Japan: 3 years after the disaster. Journal of Economic Structures, 5(1), 2 (2016)

\bibitem{Zhu2015}
Zhu, Z., Puliga, M., Cerina, F., Chessa, A., \& Riccaboni, M., Global value trees. PloS One, 10(5), e0126699 (2015)

\bibitem{ZhuMorrisonPuligaChessaRiccaboni2018}
Zhu, Z., Morrison, G., Puliga, M., Chessa, A., \& Riccaboni, M., The similarity of global value chains: A network-based measure. Network Science, 6(4), 607--632 (2018)

\end{thebibliography}


\end{document}